\documentclass{article} % For LaTeX2e
\usepackage{iclr2027_conference,times}

\usepackage{amsmath,amsfonts,bm}

\def\eqref#1{equation~\ref{#1}}
\def\1{\bm{1}}

\DeclareMathAlphabet{\mathsfit}{\encodingdefault}{\sfdefault}{m}{sl}
\SetMathAlphabet{\mathsfit}{bold}{\encodingdefault}{\sfdefault}{bx}{n}

\usepackage[utf8]{inputenc} % allow utf-8 input
\usepackage[T1]{fontenc}    % use 8-bit T1 fonts
\usepackage{url}            % simple URL typesetting
\usepackage{booktabs}       % professional-quality tables
\usepackage{amsfonts}       % blackboard math symbols
\usepackage{nicefrac}       % compact symbols for 1/2, etc.
\usepackage{microtype}      % microtypography
\usepackage{graphicx}
\usepackage{colortbl}
\usepackage{array}
\usepackage{geometry}
\usepackage{caption}
\usepackage{makecell}
\usepackage{wrapfig}
\usepackage{subcaption} 
\usepackage{graphicx}
\usepackage{natbib}
\usepackage{pifont}
\usepackage[dvipsnames]{xcolor}
\usepackage{pifont} % for cmark / xmark
\newcommand{\cmark}{\ding{51}}
\newcommand{\xmark}{\ding{55}}
\usepackage{tcolorbox}
\tcbuselibrary{skins,breakable,listings}
\usepackage{xcolor}
\usepackage{wrapfig}
\usepackage{multirow}
\usepackage{multicol}
\usepackage{listings} % Add code blocks
\usepackage{titletoc} % for table of contents of appendix only, should be BEFORE HYPERREF
\usepackage{fancyvrb}
\usepackage{tcolorbox} % Prompts in appendix

\usepackage{longtable}

\usepackage{xltabular}
\usepackage{booktabs}
\usepackage{needspace}
\definecolor{promptgray}{RGB}{247,247,247}

\newtcolorbox{promptbox}[1][]{
    enhanced,
    breakable,
    colback=promptgray,
    colframe=black!35,
    boxrule=0.6pt,
    arc=2pt,
    left=6pt,
    right=6pt,
    top=6pt,
    bottom=6pt,
    fonttitle=\bfseries,
    title=Prompt,
    #1
}

\definecolor{easycolor}{HTML}{F2B134}
\definecolor{midcolor}{HTML}{E2828F}
\definecolor{hardcolor}{HTML}{74B9D3}

\newcommand{\name}{OmniSmartHome}
\newcommand{\ours}{PROME}

\definecolor{uptri}{RGB}{38,114,168}
\definecolor{dntri}{RGB}{178,34,52}

\definecolor{lightgray}{rgb}{0.83, 0.83, 0.83}
\definecolor{Gray}{gray}{0.6}
\definecolor{aliceblue}{rgb}{0.94, 0.97, 1.0}
\definecolor{mistyrose}{rgb}{1.0, 0.89, 0.88}
\definecolor{backcolour}{rgb}{0.95,0.95,0.92}

\newcommand{\newpara}[1]{\vspace{-0.8pt}\noindent\textbf{#1}}
\newcommand{\excell}[3]{\begin{minipage}[t]{\linewidth}%
  \includegraphics[width=\linewidth]{asset/examples/#1}\par\vspace{1pt}
  \scriptsize\raggedright\textit{``#2''}\par\vspace{1pt}
  \textbf{Target:} #3\end{minipage}}
\newcommand{\excellreal}[4]{\begin{minipage}[t]{\linewidth}%
  \includegraphics[width=\linewidth]{asset/examples/#1}\par\vspace{1pt}
  \scriptsize\raggedright\textbf{#2}\par\textit{``#3''}\par\vspace{1pt}
  \textbf{Target:} #4\end{minipage}}
\newcommand{\exrow}[2]{\rotatebox{90}{\scriptsize\hspace{0.4em}\textbf{#1}~#2}}

\usepackage[accsupp]{axessibility}
\usepackage{hyperref}
\usepackage{cleveref}

\crefname{equation}{Eq.}{Eqs.}
\Crefname{equation}{Equation}{Equations}

\crefname{figure}{Fig.}{Figs.}
\Crefname{figure}{Figure}{Figures}

\crefname{table}{Tab.}{Tabs.}
\Crefname{table}{Table}{Tables}

\crefname{section}{Sec.}{Secs.}
\Crefname{section}{Section}{Sections}

\crefname{algorithm}{Alg.}{Algs.}
\Crefname{algorithm}{Algorithm}{Algorithms}

\definecolor{modelhighlight}{RGB}{225,240,250}

\usepackage{hyperref}
\usepackage{url}

\title{OmniSmartHome: A Multimodal Reasoning \\ Benchmark for Smart-Home Agents}

\author{%
  \begin{tabular}{@{}l@{}}
    Jihoo Jung$^{1\ast}$ \quad Suho Yoo$^{1}$\thanks{Equal contribution.} \quad
    Jeongsoo Choi$^{1}$ \quad Hyebin Cho$^{1}$ \\ Tae Wook Haam$^{1}$ \quad
    Hyeonggon Ryu$^{2}$ \quad Sumin Park$^{1}$ \quad Joon Son Chung$^{1}$
  \end{tabular} \\
  $^{1}$Korea Advanced Institute of Science and Technology (KAIST) \\
  $^{2}$Hankuk University of Foreign Studies \\
  \texttt{\{jihoojung, suho.yoo, joonson\}@kaist.ac.kr}
}
\iclrfinalcopy
\begin{document}

\maketitle
\lhead{Preprint}
\begin{figure*}[htbp]
    \vspace{-7mm}
    \centering
    \includegraphics[width=0.99\textwidth]{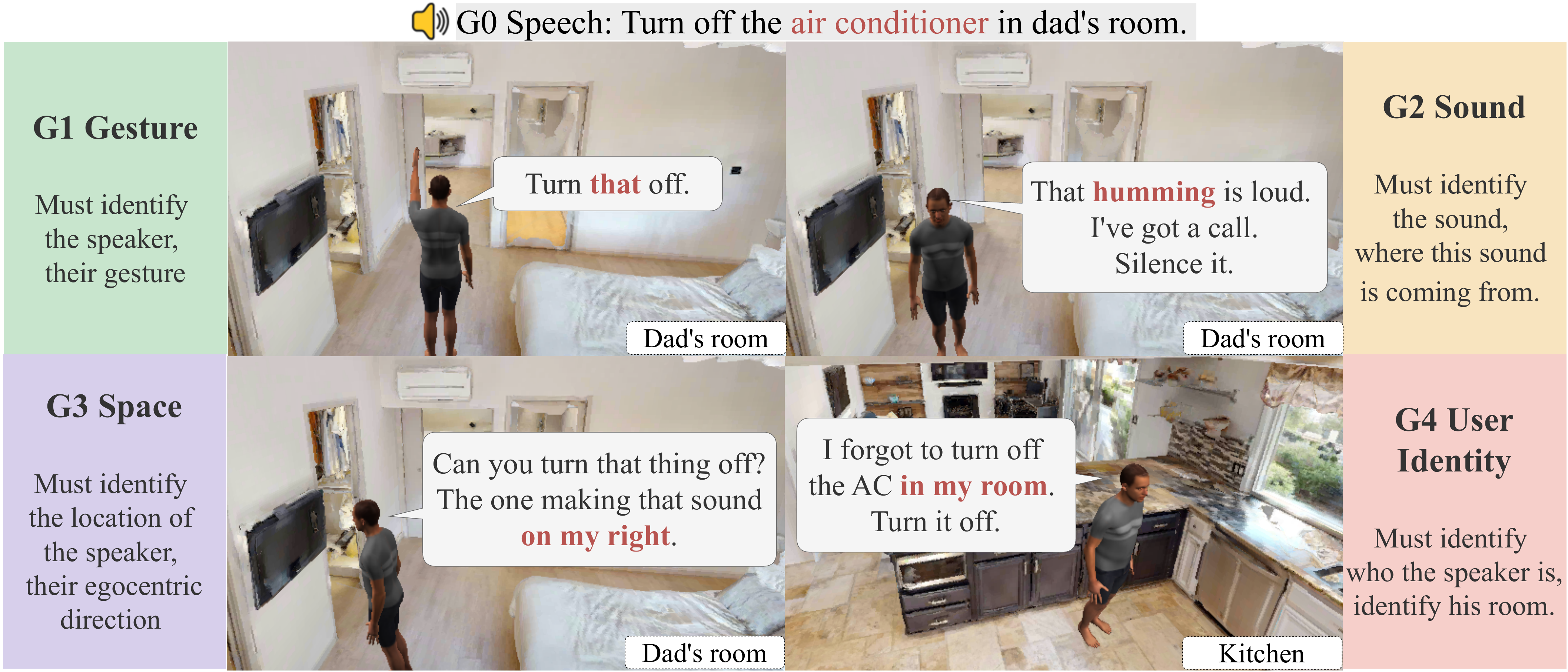} 
    \vspace{-2mm}
    \caption{\textbf{Overview of \name{}.} In real-world interactions, a user's spoken request is often underspecified in language alone, with the intended target device resolved from multimodal cues distributed across the surrounding audio-visual context (G1-G4).}
    \label{fig:main}
    \vspace{-2mm}
\end{figure*}
% while speaking—pointing at objects or referring to what they see or hea
\begin{abstract} 
Smart-home assistants are expected to handle diverse, realistic requests that arise in daily life. In such interactions, users often rely on the surrounding multimodal context—pointing at objects or referring to what they see or hear, leaving their requests underspecified in language alone. Existing smart-home benchmarks, however, express user requests solely through language, leaving context-dependent real-world requests underexplored. To bridge this gap, we introduce OmniSmartHome, a multimodal smart-home benchmark where each spoken request is paired with the surrounding visual and spatial-audio context, providing complementary cues to disambiguate underspecified requests. OmniSmartHome comprises 1,360 synthetic and 272 real-world episodes. We evaluate 16 omnimodal large language models (Omni-LLMs) and reveal that, while they perform strongly when speech alone sufficiently conveys the user's intent, performance drops substantially when resolving it requires reasoning over multimodal contextual cues. As a simple agent baseline, we provide PROME (PROcedural Memory for multimodal Evidence gathering), which equips agents with specialized audio-visual perception tools and procedural memory for orchestrating their use. PROME generally improves performance across six Omni-LLMs. Demos and examples are available at \url{https://omni-smart-home.github.io}.
% we introduce \textbf{\name{}}, a multimodal smart-home benchmark in which each spoken request is accompanied by the surrounding visual and spatial-audio context, which provides complementary cues for disambiguating the ine-grained perception with procedural memory for orchestrating them
\end{abstract}

% \begin{figure*}[htbp]
%     \vspace{-5mm}
%     \centering
%     \includegraphics[width=0.99\textwidth]{asset/main.pdf} 
%     \vspace{-2mm}
%     \caption{\textbf{Overview of \name{}.} In real-world interactions, a user's spoken request is often underspecified in language alone, with the intended target device resolved from multimodal cues distributed across the surrounding audio-visual context (G1-G4).}
%     \label{fig:main}
%     \vspace{-6mm}
% \end{figure*}

\section{Introduction}

Smart-home assistants control home devices on a user's behalf, translating each request into a device operation. Commercial systems such as Amazon Alexa have made voice-controlled home device operation widely accessible, yet they remain largely centered on explicit verbal commands. In everyday interaction, however, people often make requests that are ambiguous from language alone, as they rely on the surrounding multimodal context while speaking—pointing at objects or referring to what they see or hear~\citep{tomasello2008origins, clark1996using}.
Interpreting such requests therefore depends on that context: resolving ``\emph{Turn that off}'' requires the accompanying gesture, while resolving ``\emph{Silence that humming noise}'' requires identifying the source of the sound. Supporting such context-grounded requests is particularly beneficial for users with impaired memory or limited expressive ability, such as young children and individuals with cognitive or communication impairments~\citep{so2010speech, hendriks2014referential,jiang2026reibench}. Next-generation smart-home agents powered by large language models (LLMs) are thus expected to handle such everyday requests with their multimodal capabilities.

Benchmarks for LLM-based smart-home agents have increasingly incorporated the complexities of real-world interaction, covering invalid user requests~\citep{li-etal-2025-homebench}, dynamically changing environments~\citep{seo2026simuhome}, and personalized dialogue~\citep{li2026smh, bharadwaj2026personalhomebench}. These benchmarks, while capturing the diverse and dynamic nature of everyday interactions, are composed of requests conveyed through language alone. Despite the importance of surrounding multimodal context in everyday communication, no existing smart-home benchmark evaluates agents on requests grounded in such context.

To address this gap, we introduce \textbf{\name{}}, a multimodal smart-home benchmark in which each spoken request is accompanied by the surrounding visual and spatial-audio context, which provides complementary cues for disambiguating the underspecified request. \name{} categorizes requests into five types by how the target device is indicated in user's request: \textit{speech}, where it is named explicitly, and \textit{gesture}, \textit{sound}, \textit{space}, and \textit{user identity}, where speech alone is ambiguous and the target is inferred from the corresponding multimodal contextual cues. As illustrated in \cref{fig:main}, a target explicitly named in \textit{speech} as ``\emph{air conditioner in dad's room}'' may instead be referred to as ``\emph{that}'' with a pointing gesture (\textit{gesture}), by its sound (``\emph{humming}''; \textit{sound}), by its relative location (``\emph{on my right}''; \textit{space}), or through a possessive reference (``\emph{my room}''; \textit{user identity}). Fulfilling such requests involves two steps: identifying the target device and executing the requested operation on it. We evaluate these steps with \emph{grounding} and \emph{goal} accuracy, respectively.
% \name{} contains 1,360 synthetic and 272 real-world episodes. 

The grounding step is the new challenge introduced by \name{}: it requires \emph{perceiving} fine-grained cues from audio-visual context, and \emph{aggregating} them across modalities. Recent omnimodal large language models (Omni-LLMs), despite accepting audio-visual input, struggle to perceive such cues as pointing direction, sound source location, spatial relations, and speaker identity~\citep{Choi_2026_CVPR, li-etal-2026-mllms, chen2026savvy, bai2026humanomni}, and even when the cues are accessible, often fail to integrate them into a coherent decision~\citep{luong2026mcbench,li2026omnibench,Zhang_2026_CVPR}. Evaluating 16 Omni-LLMs on \name{} exposes these limitations: models perform well when the target device is explicitly named in \textit{speech}, but struggle to infer it from multimodal contextual cues, leaving a clear gap to human performance. As a simple agent baseline, we provide \ours{} (PROcedural Memory for multimodal Evidence gathering), which combines multimodal tools for fine-grained perception with procedural memory for orchestrating them. Across six Omni-LLMs, \ours{} improves both grounding and goal accuracy.

Our contributions are threefold: (1) We introduce \name{}, the first multimodal smart-home benchmark in which spoken requests are accompanied by surrounding visual and spatial-audio context that provides complementary cues---gesture, sound, space, user identity---for resolving ambiguous requests.
 (2) We evaluate 16 Omni-LLMs, showing that current models struggle to interpret user requests that depend on multimodal context, while performing strongly when the request is fully specified in speech alone. (3) To narrow this gap, we propose \ours{}, a simple agent baseline that pairs specialized audio-visual perception tools with a procedural memory, improving both grounding and goal accuracy across six Omni-LLMs.
% when resolving the target requires reasoning over implicit multimodal evidence. improves both device grounding and task execution acros.
\section{Related Works}

\begin{table}[t]
    \centering
    \vspace{-2mm}
    \caption{\textbf{Comparison with related smart-home benchmarks.} \name{} covers diverse modalities spanning text, visual, and audio, and provides both synthetic and real-world data.}
    \vspace{-3mm} 
    \label{tab:benchmark_comparison}
    \setlength{\tabcolsep}{6pt}
    \resizebox{\columnwidth}{!}{
    \begin{tabular}{lcccc|cc|c}
        \toprule
        \multirow{2}{*}{\textbf{Benchmark}}
        & \multicolumn{4}{c|}{\textbf{Modality}}
        & \multicolumn{2}{c|}{\textbf{Data Type}}
        & \multirow{2}{*}{\textbf{Data Size}} \\
        \cmidrule(lr){2-5}
        \cmidrule(lr){6-7}
        & \textbf{Text}
        & \textbf{Visual}
        & \makecell{\textbf{Speech}\\\textbf{Instruction}}
        & \makecell{\textbf{General}\\\textbf{Audio}}
        & \textbf{Synthetic}
        & \makecell{\textbf{Real-}\\\textbf{world}}
        & \\
        \midrule
        HomeBench~\citep{li-etal-2025-homebench} & \cmark & \xmark & \xmark & \xmark & \cmark & \xmark & 170k \\
        SimuHome~\citep{seo2026simuhome}     & \cmark & \xmark & \xmark & \xmark & \cmark & \xmark & 0.6k \\
        PersonalHomeBench~\citep{bharadwaj2026personalhomebench} & \cmark & \cmark & \xmark & \xmark & \cmark & \cmark & 9k \\
        SMH-Bench~\citep{li2026smh}          & \cmark & \xmark & \xmark & \xmark & \cmark & \xmark & 1.1k \\
        MIST~\citep{chen2026mist}            & \cmark & \xmark & \cmark & \xmark & \cmark & \xmark & 10k \\
        \midrule
        \textbf{\name{}}                       & \cmark & \cmark & \cmark & \cmark & \cmark & \cmark & \textbf{1.6k} \\
        \bottomrule
    \end{tabular}
    }
    % \vspace{-7mm}
    \vspace{-1mm}
\end{table}
\newpara{Smart-home agents.} Smart-home agents fulfill user requests by operating home devices through tool calls. A growing number of benchmarks have been proposed to evaluate such agents under realistic interactions (Table~\ref{tab:benchmark_comparison}). HomeBench~\citep{li-etal-2025-homebench} includes both valid and invalid requests. SimuHome~\citep{seo2026simuhome} introduces an interactive simulator in which the agent's device-control calls update device states and, in turn, environmental variables. It further introduces a simulator-based evaluation protocol that evaluates agents by the resulting simulator state. SMH-Bench~\citep{li2026smh} incorporates ambiguous instructions, multi-turn interaction, and personalization. More recent benchmarks have begun to incorporate non-textual inputs, but only partially. PersonalHomeBench~\citep{bharadwaj2026personalhomebench} includes silent video, but primarily evaluates video understanding rather than device control, while MIST~\citep{chen2026mist} replaces text with speech but does not require broader multimodal context, such as general audio or visual cues, to resolve the request.  In contrast, \name{} incorporates spatial audio containing both speech and general environmental sounds, alongside visual observations, requiring agents to reason over such multimodal context.

\newpara{Omni-LLMs.}
Extending the capabilities of LLMs, Omni-LLMs have recently emerged,
broadening text-centric understanding to encompass audio and visual perception. Omni-LLMs range from smaller-scale models~\citep{tang2025video, xu2025qwen25omnitechnicalreport, cui2026minicpm} to larger-scale models with strong reasoning capabilities~\citep{xu2025qwen3, deshmukh2026nemotron, team2026gemma}. Despite advances in holistic multimodal understanding, these models remain weak at perceiving fine-grained cues~\citep{Choi_2026_CVPR, li-etal-2026-mllms, chen2026savvy, zheng2025multimodal, bai2026humanomni}, and, even when such cues are available, often fail to aggregate them into a coherent conclusion~\citep{luong2026mcbench,li2026omnibench,xu2026agentic,Zhang_2026_CVPR}. A growing line of work therefore augments multimodal LLMs with specialized perception tools~\citep{yang2023mmreact, chen2026savvy} and learns to orchestrate them, typically with reinforcement learning~\citep{tong2026autagent,zheng2026deepeyes} or procedural memory~\citep{wang2025mirix,si2026vla}. Similarly, our agent baseline, \ours{}, defines a set of perception tools tailored to smart-home and builds a procedural memory to guide tool usage and evidence aggregation.

% \newpara{Agentic memory system.} 
% Procedural memory provides reusable guidance from past experience to inform these decisions.
% Agent Workflow Memory~\citep{wang2025awm} extracts reusable procedures
% from past trajectories and places the procedure collection in the context of the agent.
% Memp~\citep{fang2026memp} instead retrieves trajectories relevant to the current task and combines them with generalized procedures.
% These approaches have further evolved to use reinforcement learning
% for memory management and usage.
% Memory-R1~\citep{yan2025memoryr1} uses one LLM to add, update, or delete memory entries and another to reason over retrieved memory, training them separately with GRPO~\citep{shao2024deepseekmath}. We propose a lightweight agentic system equipped with multimodal perception tools to complement the ability of Omni-LLMs and a procedural memory that guides their use.

\section{\name}
We present the benchmark taxonomy (\cref{sec:benchmark_taxonomy}), dataset generation pipeline (\cref{sec:episode_gen}), and evaluation protocol (\cref{sec:eval_methods}), followed by a simple agent baseline (\cref{sec:agent}).
\subsection{Benchmark Taxonomy} 
\label{sec:benchmark_taxonomy}
\name{} defines an \emph{episode profile} along three dimensions: a \emph{grounding type}, indicating which multimodal cue the user relies on to specify the target device; a \emph{query type}, indicating what the user asks for; and a \emph{feasibility type}, indicating whether the request can be fulfilled or not.

\newpara{Grounding types.}
We define five \emph{grounding types}: one explicit baseline (G0) and four implicit types (G1--G4) that rely on multimodal contextual cues.
\textit{Speech-grounded (G0).} The user explicitly names the target device in the spoken utterance, so grounding reduces to speech recognition alone.
\textit{Gesture-grounded (G1).} The user refers to the target device deictically (e.g., ``Turn \emph{that} off'') while pointing toward it, requiring the agent to interpret the gesture and ground it in the visual scene.
\textit{Sound-grounded (G2).} The user refers to the target device by the sound it emits (e.g., ``Silence that \emph{humming noise}''), requiring the agent to detect and localize the sound source and associate it with the corresponding device.
\textit{Space-grounded (G3).} The user refers to the target device using an egocentric direction (e.g., ``Turn off the light \emph{on my right}'' or ``Silence that sound coming \emph{from my right}''), requiring the agent to reason about the referenced direction from the user's perspective and identify the target located there.
\textit{User-Identity-grounded (G4).} The user refers to the target device through a possessive reference (e.g., ``Turn off the AC in \emph{my room}''), requiring the agent to identify the speaker by face or voice and resolve the reference against that person's household profile.

\begin{wrapfigure}{r}{0.38\linewidth}
  \centering
  \vspace{-6mm}
  \includegraphics[width=\linewidth]{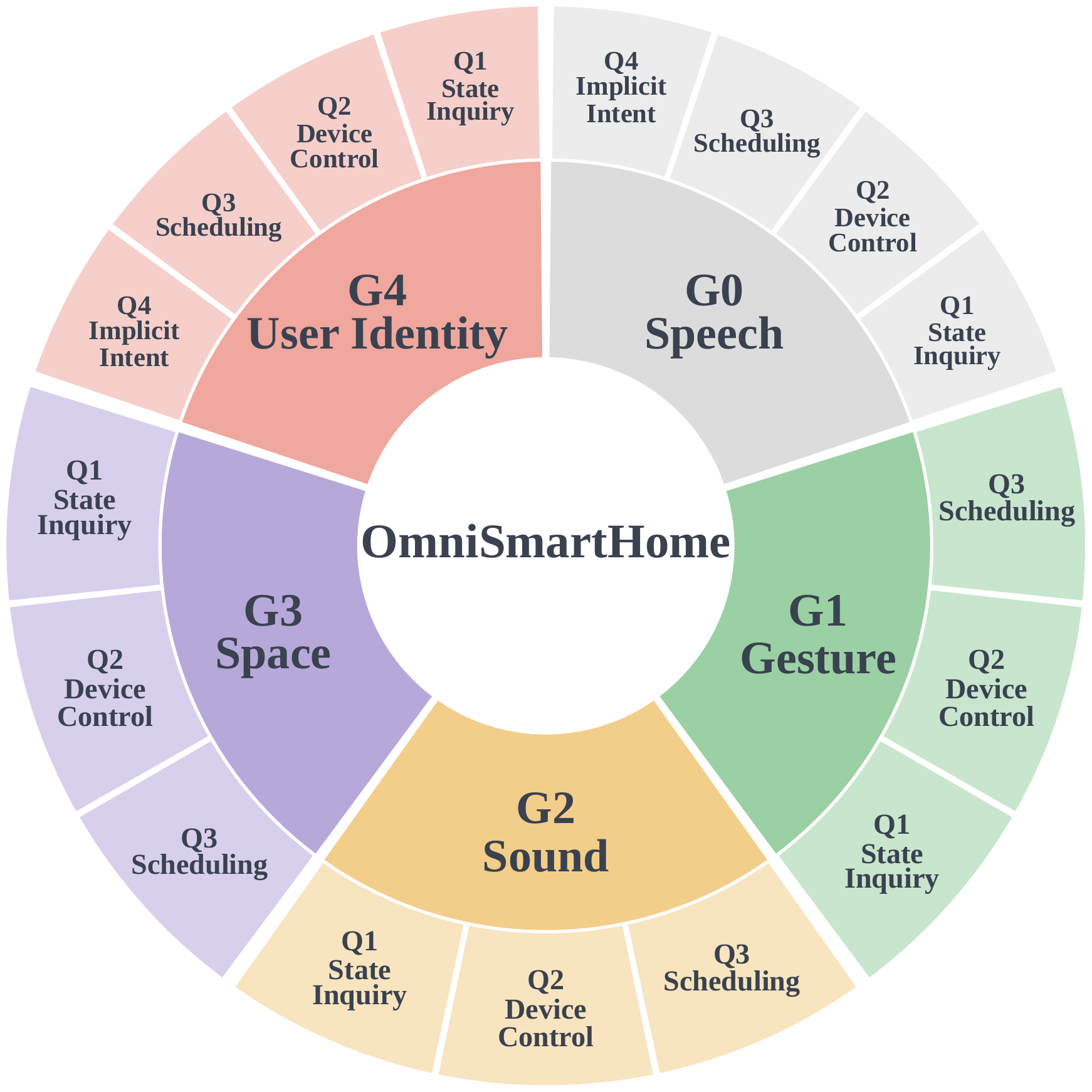}
  \vspace{-7mm}
    \caption{\textbf{Taxonomy of OmniSmartHome.} Each episode has a grounding, query, and feasibility type.}
  \vspace{-5mm}
\end{wrapfigure}

\newpara{Query types.} We define four query types, following SimuHome~\citep{seo2026simuhome}.
\textit{State inquiry (Q1).} The user asks about the current state or attributes
of a device (e.g., ``What temperature is the AC set to?''),
requiring the agent to retrieve the relevant information and report it in a textual response.
\textit{Explicit device control (Q2).} The user asks the agent to set a specific
device to a desired state (e.g., ``Turn that off''), requiring the
agent to execute the corresponding device operation.
\textit{Workflow scheduling (Q3).} The user requests one or more device actions
to be executed at a future time or in coordination with another event
(e.g., ``In ten minutes, turn that off''), requiring the agent to schedule the appropriate workflow.
\textit{Implicit intent (Q4).} The user expresses a desired outcome without
explicitly specifying the action to take (e.g., ``it's too cold in my room''),
requiring the agent to infer the underlying intent and determine the appropriate device operation. Since implicit-intent requests are difficult to combine with G1-G3 grounding types, Q4 is instantiated only for G0 and G4 episodes.

\newpara{Feasibility types.}
We construct both \emph{feasible} and \emph{infeasible} episodes.
For the synthetic dataset, feasible episodes are further divided into three
difficulty levels---\emph{easy}, \emph{medium}, and \emph{hard}---based on
the distractors in the scene.
For example, in gesture-grounded (G1) episodes, easy scenes feature a single candidate device; medium scenes contain multiple devices requiring precise gesture grounding; and hard scenes introduce non-speakers, necessitating active speaker detection prior to gesture recognition. For infeasible episodes, we consider three causes of infeasibility: \emph{non-existent resources}, where the referenced device or perceptual cue is absent; \emph{physical limits}, where the relevant device exists but cannot realize the requested change due to its operational limits; and \emph{temporal contradictions}, where the requested timing conflicts with other time constraints or device operating conditions. (Details in Appendix~\ref{sec:benchmark_detail}.)

\subsection{Dataset Generation}
\label{sec:episode_gen}
\newpara{Definition and components of an episode.}
 Each episode comprises four components. (1) The \textit{visual and acoustic observations} capture the room where the user is currently located: the visual observation shows the room and any people present, and the acoustic observation is a first-order Ambisonics (FOA) recording of the user's spoken request along with sounds from active devices. (2) The \textit{initial home state} defines the starting configuration of the home: the devices in each room, their positions and initial states, and household-member profiles. (3) The \textit{target device} is the ground-truth device that the request intends to control or query. (4) The \textit{goal} specifies the correct outcome: the ground-truth state or attribute of the target device for Q1, and the desired state of the target device for Q2--Q4.
 
\begin{figure*}[t]
\vspace{-2mm}
    \centering
    \includegraphics[width=\textwidth]{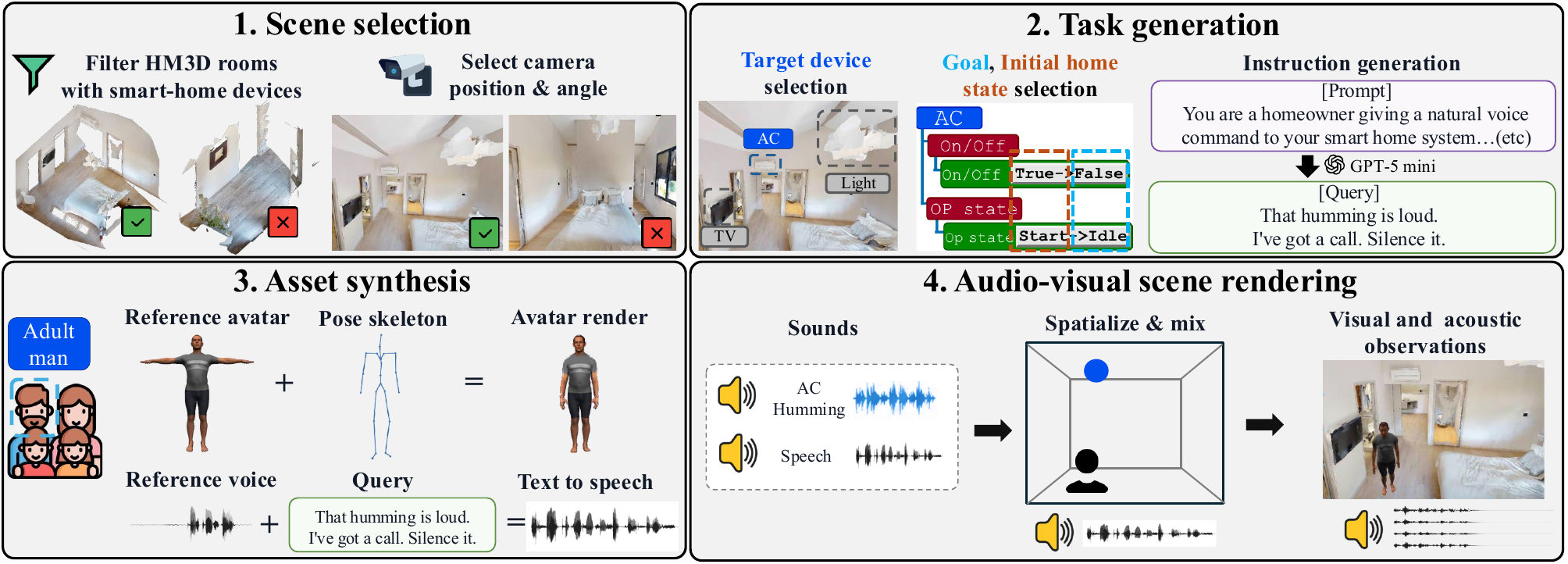}
    \vspace{-7mm}
% \caption{\textbf{Synthetic episode generation pipeline.} The initial home state, target device, and goal are determined in Stage~2, and the visual and acoustic observations are rendered in Stage~4.}
\caption{\textbf{Synthetic episode generation pipeline.}
(1) Select an room containing suitable smart-home devices and a camera viewpoint.
(2) Determine the target device, goal and initial home state and generate a corresponding user request in text.
(3) Synthesize the avatar and speech, prepare device sounds.
(4) Place the assets in the scene and spatialize the audio sources to produce the final visual and acoustic observations.}
    \label{fig:episode_gen}
    \vspace{-6mm}
\end{figure*}

\newpara{Synthetic episodes.}
We construct synthetic episodes in four stages, as illustrated in
Fig.~\ref{fig:episode_gen}.
(1) \textit{Scene selection.}
Scenes are drawn from the test split of HM3D~\citep{yadav2023habitat}, a collection of photorealistic 3D meshes of indoor home environments. We retain only rooms containing suitable home devices and manually select camera viewpoints that provide clear visibility for each room.
(2) \textit{Task generation.}
For each room, we assign episode profiles (grounding, query, and feasibility types). Conditioned on this profile, we select an appropriate \textit{target device} and \textit{goal}, and construct the \textit{initial home state} such that the goal is initially unsatisfied. An LLM~\citep{openai2025gpt5mini} then generates a natural user request in text consistent with the profile and goal.
(3) \textit{Asset synthesis.}
We sample one of four demographic categories (girl, boy, adult woman, adult man), render a matching humanoid avatar~\citep{puig2024habitat} in a natural pose, and synthesize the user request as speech with a matching voice using recent speech synthesis model~\citep{zhang2026moss}. For episodes requiring device sounds (e.g., humming, media playback, beeping, or chimes), we use a device-specific audio clip collected from Freesound~\citep{font2013freesound}.
(4) \textit{Audio-visual scene rendering.}
Using the 3D audio-visual simulator SoundSpaces~2.0~\citep{chen22soundspaces2}, we place the selected avatar in the scene and emit the synthesized speech from the avatar's position, along with device-specific sounds from the corresponding device locations. This yields the \textit{visual and acoustic observation} (a rendered image and an FOA audio).

\newpara{Real-world episodes.} We construct real-world episodes in five indoor environments. Five participants act out LLM-generated queries~\citep{openai2025gpt5mini}, which we record using a camera and a spatial-audio microphone. Unlike the synthetic set with static scene images, the real-world set consists of video recordings. 

All episodes are manually verified by human (Appendix~\ref{sec:human_valid}). We retain, for each grounding--query combination, 20 synthetic episodes per feasible difficulty level and 20 infeasible episodes, and eight feasible and eight infeasible real-world episodes, yielding 1,360 synthetic and 272 real-world episodes.

\subsection{Episode Execution and Evaluation}
\label{sec:eval_methods}

\begin{figure*}[t]
    \vspace{-1mm}
    \centering
    \includegraphics[width=0.99\linewidth]{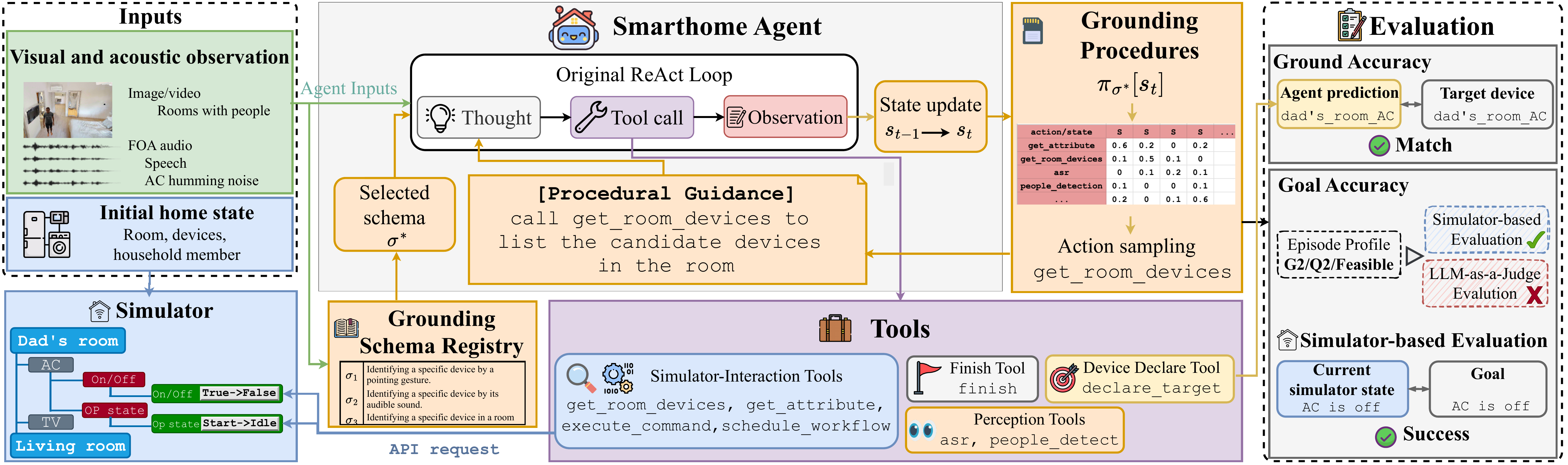}
    \vspace{-3mm}
% \caption{\textbf{Illustration of the agent framework.} The base agent excludes the components highlighted in orange, while \ours{} incorporates them. The figure illustrates the roles and interactions of the four episode components, simulator, tools, and evaluation.}
\caption{\textbf{Illustration of our agent framework.}
Components highlighted in \textcolor[HTML]{D79B00}{orange} are used only by \ours{}.
The agent receives the visual and acoustic observations and iteratively gathers evidence to identify the target device using simulator-interaction or \textcolor[HTML]{D79B00}{perception tools}.
\ours{} additionally provides \textcolor[HTML]{D79B00}{procedural guidance}.
Once the target device is identified, the agent declares it and either queries its information (Q1) or executes the requested operation (Q2-Q4) via simulator-interaction tools.
Grounding accuracy evaluates the declared target device, while goal accuracy is evaluated from the simulator state or by an LLM judge.}
    \label{fig:agentic}
    \vspace{-5mm}
\end{figure*}

\newpara{Episode execution.}
\cref{fig:agentic}, excluding the \textcolor[HTML]{D79B00}{orange components}, illustrates the full architecture of the base agent. Each episode begins by initializing the simulator~\citep{seo2026simuhome} with the \textit{initial home state}, while the Omni-LLM agent receives the \textit{visual and acoustic observations}. The agent operates in a ReAct loop~\citep{yao2023react}, alternating between reasoning, tool invocation, and observation. At each step, the agent can invoke simulator-interaction tools as well as two dedicated tools, \textit{Device Declare} and \textit{Finish}. Simulator-interaction tools access the simulator either to retrieve information about the device state or to modify it by executing device operations. The \textit{Device Declare} tool records the agent's predicted target device and is required before the \textit{Finish} tool can be called, which terminates the episode and returns the final textual response.

\newpara{Evaluation metrics.} We evaluate agents using two metrics: \textit{grounding accuracy} and \textit{goal accuracy}. \textit{Grounding accuracy} measures whether the device declared via the \textit{Device Declare} tool matches the ground-truth target device. For infeasible requests due to \emph{non-existent resources}, where no valid target exists, the agent is expected to declare no target.
\textit{Goal accuracy} measures whether the agent ultimately satisfies the requested goal, evaluated using either a simulator or an LLM-as-a-judge~\citep{zheng2023judging, liu2023geval}. For feasible Q2--Q4 queries, we inspect the final simulator state and check whether it matches the state specified in the ground-truth goal. For Q1 queries and infeasible requests, where the agent is expected to return a textual response rather than execute a device operation, we use an LLM-as-a-judge. For Q1, the judge assesses whether the response is consistent with the states and attributes of the target device; for infeasible requests, it assesses whether the agent correctly recognizes the infeasibility and explains its cause.

\subsection{A Simple Agent Baseline}
\label{sec:agent}

Before operating a device, an agent first resolves which device the user refers to from the audio-visual context. This involves actively gathering fine-grained multimodal cues and integrating them differently depending on how the request is grounded in the environment---recent Omni-LLMs often struggle with~\citep{Choi_2026_CVPR, li-etal-2026-mllms, chen2026savvy, bai2026humanomni, luong2026mcbench, li2026omnibench}. To support this process, we introduce \textbf{\ours}
(\textbf{PRO}cedural \textbf{M}emory for multimodal \textbf{E}vidence
gathering) as a simple agent baseline.
\ours{} equips the agent with a set of audio-visual perception tools and
augments its reasoning with a procedural memory that guides when such tools should be invoked and in what order.

\newpara{Multimodal perception tools.}
We provide the agent with perceptual tools including speech recognition, pointing estimation, sound localization, head pose recognition, face recognition and speaker recognition~\citep{radford2023robust,ovseld,jiang2024rtmw,zhou2023directmhp,he2017mask,oquab2023dinov2,chen2022wavlm}. Further details are provided in Appendix~\ref{app:tool}.

\newpara{Procedural memory.}
Procedural memory guides the agent by suggesting the next tool-calling action, based on which contextual cue the request relies on to indicate the target device and the evidence gathered so far. It consists of a bank of reusable \emph{grounding procedures}, $\mathcal{M}=\{\pi_\sigma\}_{\sigma\in\Sigma}$, where each $\pi_\sigma$ encodes the tool-calling strategy for a \emph{grounding schema} $\sigma\in\Sigma$, characterized by the type of contextual cue that identifies the target device (e.g., a pointing gesture resolving ``this'' or ``that''). Formally, $\pi_\sigma$ defines a probability distribution over tool actions $a\in\mathcal{A}$ conditioned on the current agent state $s_t\in\mathcal{S}$, which summarizes the evidence acquired and the remaining uncertainty at ReAct step $t$. We parameterize $\pi_\sigma$ as a tabular softmax policy,
$\pi_\sigma(a\mid s)\propto\exp(\theta_{\sigma,s,a})$,
with learnable logits
$\theta\in\mathbb{R}^{|\Sigma|\times|\mathcal{S}|\times|\mathcal{A}|}$.

At inference time, the agent first examines the multimodal input and retrieves the grounding schema $\sigma^\ast \in \Sigma$ that best matches the episode. After each ReAct step, the agent updates the current state $s_{t}$ based on the evidence accumulated so far, samples $a_{t}\sim\pi_{\sigma^\ast}[s_{t}]$, and injects the sampled action as a brief \textsc{[Procedural Guidance]} before the next reasoning step. 
 % In this way, procedural memory provides step-wise guidance for evidence acquisition.

\newpara{Procedural memory construction.}
We build procedural memory from synthetic training
episode, generated using the same pipeline as in \cref{sec:episode_gen} with scenes drawn from the HM3D training split. To construct $\Sigma$, we prompt an Omni-LLM~\citep{xu2025qwen3} to describe, for each training episode, which cue is needed to identify the target device, and to group episodes with similar cue descriptions into schemas. For $\pi_\sigma$, we initialize the logits $\theta^{0}$ as a cost prior plus a utility bonus for each perception tool, where the utility is estimated from two memory-free reference rollouts per episode:
$\tau_{\text{ref}}^{+}$ with perception tools and $\tau_{\text{ref}}^{-}$
without. We then refine
$\theta^{0}$ with a policy-gradient update~\citep{williams1992simple}:
\begin{equation}
\nabla_{\theta}\mathcal{J}
=\mathbb{E}_{e}\,\mathbb{E}_{\tau\sim\mathcal{G}_e}\Big[A(\tau)\sum_{t\in\mathcal{T}(\tau)}
\nabla_{\theta}\log\pi_{\sigma^*}[s_t](a_t)\Big],
\qquad
A(\tau)=G(\tau)-\tfrac{1}{|\mathcal{G}_e|}\textstyle\sum_{\tau'\in\mathcal{G}_e}G(\tau'),
\label{eq:reinforce}
\end{equation}
where 
$\mathcal{G}_e=\{\tau,\tau_{\text{ref}}^{+},\tau_{\text{ref}}^{-}\}$ adds a
memory-guided rollout $\tau$ sampled under $\theta^{0}$ to the two reference
rollouts, $G(\tau)$ is the grounding reward, and
$\mathcal{T}(\tau)$ is the set of decision steps. The training set is rolled
out once and updated in a single batch (details in Appendix~\ref{app:details_of_method}).

\section{Experiments}
% =========================================================
% Required packages / macros
% =========================================================

% Required:
% \usepackage{booktabs}
% \usepackage{array}
% \usepackage{graphicx}
% \usepackage[table]{xcolor}   % \rowcolor 쓰려면 table 옵션 필요
% \usepackage{amssymb}

% Metric column
\newcolumntype{C}{>{\centering\arraybackslash}p{0.9cm}}
\newcommand{\psize}[1]{\,\textcolor{neutralgray}{\scriptsize #1}}
% Colors for gains / losses
\definecolor{gainblue}{RGB}{90,120,160}
\definecolor{lossred}{RGB}{160,115,120}
\definecolor{neutralgray}{RGB}{110,110,110}
\definecolor{modelhighlight}{RGB}{235,240,250}  % 이미 정의돼 있으면 이 줄 제거
\definecolor{allshade}{RGB}{245,245,245}
% =========================================================
% Score macros
% =========================================================

\newcommand{\scoreup}[2]{%
  \ensuremath{#1_{\textcolor{gainblue}{\scriptscriptstyle #2}}}%
}

\newcommand{\scoredown}[2]{%
  \ensuremath{#1_{\textcolor{lossred}{\scriptscriptstyle #2}}}%
}

\newcommand{\scoreflat}[2]{%
  \ensuremath{#1_{\textcolor{neutralgray}{\scriptscriptstyle #2}}}%
}

\newcommand{\ms}[2]{%
  \ensuremath{#1_{\scriptscriptstyle\pm#2}}%
}

\newcommand{\gain}[1]{\textcolor{gainblue}{\textbf{\scriptsize #1}}}
\newcommand{\loss}[1]{\textcolor{lossred}{\scriptsize #1}}

\begin{table*}[t]
\centering
\vspace{-2mm}
\caption{
\textbf{Results on OmniSmartHome across grounding types (G0--G4).}
For each grounding type, we report grounding accuracy (Gr.) and goal accuracy (Goal), averaged over query types.
}

\label{tab:main_results}
\vspace{-3mm}
\renewcommand{\arraystretch}{1.10}
% \vspace{-1mm}
\resizebox{\textwidth}{!}{%
\footnotesize
\begin{tabular}{
l
>{\columncolor{allshade}}C@{\hspace{2pt}}>{\columncolor{allshade}}C@{\hspace{8pt}}
C@{\hspace{2pt}}C@{\hspace{8pt}}
C@{\hspace{2pt}}C@{\hspace{8pt}}
C@{\hspace{2pt}}C@{\hspace{8pt}}
C@{\hspace{2pt}}C@{\hspace{8pt}}
C@{\hspace{2pt}}C
}

\toprule

&
\multicolumn{2}{c}{\textbf{All}}
&
\multicolumn{2}{c}{\textbf{G0} Speech}
&
\multicolumn{2}{c}{\textbf{G1} Gesture}
&
\multicolumn{2}{c}{\textbf{G2} Sound}
&
\multicolumn{2}{c}{\textbf{G3} Space}
&
\multicolumn{2}{c}{\textbf{G4} Identity}
\\

\cmidrule(lr){2-3}
\cmidrule(lr){4-5}
\cmidrule(lr){6-7}
\cmidrule(lr){8-9}
\cmidrule(lr){10-11}
\cmidrule(lr){12-13}

\textbf{Model}
&
Gr. & Goal
&
Gr. & Goal
&
Gr. & Goal
&
Gr. & Goal
&
Gr. & Goal
&
Gr. & Goal
\\

\midrule

Human Level
& 94.8 & 82.3
& 97.8 & 82.1
& 97.3 & 89.4
& 93.1 & 91.7
& 93.4 & 84.1
& 91.7 & 80.4
\\

\midrule
\multicolumn{13}{c}{\textit{Open-source Omni-LLMs (<10B)}} \\
\midrule

Video-LLaMA2\psize{7B}
& 5.8 & 4.9
& 6.8 & 3.4
& 6.2 & 6.9
& 6.2 & 6.2
& 7.6 & 6.2
& 2.9 & 2.9
\\

Qwen2.5-Omni\psize{7B}
& 33.0 & 17.6
& 55.5 & 25.5
& 30.9 & 19.4
& 28.1 & 14.2
& 27.2 & 17.4
& 20.3 & 10.9
\\

Spatial-Omni\psize{7B}
& 20.0 & 11.5
& 39.1 & 14.1
& 14.9 & 11.5
& 16.3 & 10.4
& 16.0 & 11.8
& 10.4 & 9.4
\\

video-SALMONN2+\psize{7B}
& 27.0 & 13.7
& 48.2 & 20.6
& 20.1 & 11.8
& 23.6 & 13.5
& 24.7 & 14.9
& 15.4 & 7.6
\\

video-SALMONN-o1\psize{7B}
& 24.5 & 10.2
& 44.5 & 13.5
& 24.7 & 10.4
& 17.4 & 9.4
& 25.3 & 11.8
& 9.1 & 6.5
\\

OmniVinci\psize{9B}
& 17.8 & 8.8
& 31.2 & 9.9
& 9.0 & 6.9
& 14.9 & 9.7
& 17.0 & 9.7
& 13.5 & 7.6
\\

MiniCPM-o-4.5\psize{9B}
& 60.6 & 30.9
& 85.2 & 51.0
& 56.9 & 29.9
& 56.6 & 26.4
& 59.4 & 32.3
& 42.7 & 14.1
\\

\midrule
\multicolumn{13}{c}{\textit{Large Open-source Omni-LLMs}} \\
\midrule

Gemma4\psize{12B}
& 50.6 & 25.2
& 69.3 & 45.3
& 46.5 & 19.1
& 55.2 & 29.2
& 54.9 & 26.0
& 28.1 & 6.0
\\

Nemotron3-Omni\psize{30B-A3B}
& 62.7 & 44.5
& 86.5 & 60.9
& 65.3 & 52.4
& 67.4 & 47.2
& 64.9 & 51.4
& 31.8 & 15.1
\\

Qwen3-Omni-Instruct\psize{30B-A3B}
& 66.6 & 43.8
& 88.0 & 63.0
& 66.3 & 53.8
& 77.4 & 38.2
& 60.8 & 48.6
& 41.7 & 17.4
\\

Qwen3-Omni-Think\psize{30B-A3B}
& 65.9 & 48.4
& 87.8 & 67.4
& 66.0 & 56.6
& 79.9 & 56.9
& 72.2 & 55.2
& 28.6 & 11.7
\\

MiMo-V2.5\psize{310B-A15B}
& 78.4 & 60.2
& 93.5 & 73.7
& 72.2 & 63.5
& 81.9 & 60.8
& 84.7 & 69.4
& 60.7 & 37.0
\\

\midrule
\multicolumn{13}{c}{\textit{Proprietary Omni-LLMs}} \\
\midrule

Gemini-2.5-Flash
& 73.7 & 56.2
& 94.0 & 76.8
& 69.4 & 60.1
& 74.3 & 54.9
& 78.1 & 63.2
& 52.6 & 28.6
\\

Gemini-2.5-Pro
& 81.0 & 63.1
& 95.1 & 80.5
& 74.7 & 66.7
& 81.9 & 50.3
& 83.0 & 69.8
& 69.5 & 47.7
\\

Gemini-3.1-Pro (Preview)
& 85.6 & 73.4
& 95.3 & 81.0
& 76.4 & 71.5
& 97.6 & 80.9
& 87.5 & 79.2
& 72.4 & 57.3
\\

Qwen3.8-Omni-Flash
& 81.7 & 63.1
& 93.2 & 76.8
& 76.7 & 64.2
& 92.7 & 66.0
& 85.4 & 72.2
& 63.0 & 39.6
\\

\bottomrule
\end{tabular}
}

% \vspace{-1mm}
\includegraphics[width=\textwidth]{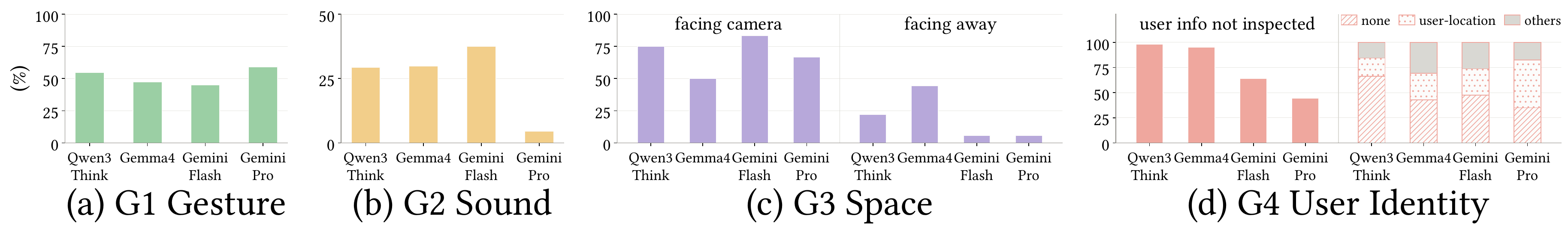}
\vspace{-7mm}
% \vspace{-6mm}
\captionof{figure}{\textbf{Representative causes of grounding failures of the base agents.} For each grounding type (G1--G4), we show a representative failure pattern.}
\label{fig:error}
% \vspace{-8mm}
\vspace{-7mm}
\end{table*}

We evaluate 16 Omni-LLMs: seven open-source models with fewer than 10B parameters, five larger open-source models, and four proprietary models. For models that do not support spatial audio input, we downmix the FOA recordings to mono. To construct \ours{} memory, we use 680 synthetic training episodes from the HM3D training split, generated using the pipeline described in \cref{sec:episode_gen}, and apply \ours{} to six models.

\subsection{Benchmarking Omni-LLMs on \name{}}
\newpara{Main results.} \cref{tab:main_results} reports the results on \name{} with base agents. Small open-source models show generally low grounding accuracy across all grounding types, with MiniCPM-o-4.5 performing best in this group. Their goal accuracy is often substantially lower than their grounding accuracy, in some cases by more than a factor of two, indicating that even when these models identify the target device, they still struggle to complete the requested goal.
Larger open-source models achieve higher overall accuracy, with a smaller gap between grounding and goal performance. Their grounding performance is strongest on G0, where identifying the target device requires little beyond speech recognition, but drops on G1--G4, which require integrating additional multimodal evidence.
Proprietary models generally outperform open-source models, yet show a similar gap between G0 and G1--G4. For these stronger models, the contrast between nearly saturated grounding and goal performance on G0 and lower performance on G1--G4 highlights multimodal grounding as the primary challenge of \name{}. Additional analysis is provided in Appendix~\ref{app:more_results}.

% \begin{table}[t]
%   \centering
%   \includegraphics[width=\linewidth]{asset/base.pdf}
%   % \includegraphics[width=\linewidth]{asset/error_ours_effect.pdf}
%   \vspace{-6mm}
%   \caption{Representative causes of grounding failures of the base agents.}
%   \label{fig:error}
%   \vspace{-4mm}
% \end{table}

\newpara{Error analysis.} \cref{fig:error} examines four representative grounding failures across Qwen3-Omni-Think, Gemma4, Gemini-2.5-Flash, and Gemini-2.5-Pro.
\textbf{G1.} In around 50\% of the episodes where the agent declares an incorrect device, the chosen device is the one nearest to the speaker's hand or body (\cref{fig:error}.a). This suggests that agents often fail to trace the pointing direction and instead fall back to selecting a device near the gesture itself.
\textbf{G2.} \cref{fig:error}.b reports, among the episodes where tbe agent declares an incorrect device, the fraction in which the agent declares the sound source prematurely, i.e., without checking the state of any candidate device, such as whether it is running or its alarm is active. Instead, the agent often relies on semantic priors about which devices typically produce a given sound: for instance, an alarm from a refrigerator is attributed to a microwave, since the agent assumes that microwaves commonly emit alarm sounds. 
\textbf{G3.} \cref{fig:error}.c shows failure rates by speaker orientation. When the speaker faces the camera, their left--right directions are reversed relative to the image, and three models fail on over 70\% of such cases, whereas error rates remain low when the speaker faces away. This suggests that models struggle to interpret directions from the speaker's egocentric view.
\textbf{G4.} The left of \cref{fig:error}.d shows, among failed episodes, the fraction in which the agent does not inspect any user-related information. Rather than querying user information to identify the device associated with the speaker's room, most of these cases end up either declaring no valid device (hatched bars) or selecting a device in the room where the user is currently located (dotted bars), as shown on the right.

% requires: \usepackage{multirow}, \usepackage{subcaption}
\begin{table*}[t]
\centering
\caption{\textbf{Results and analysis on \ours{}.} (a) Per-grounding-type results; small numbers denote the change in percentage points relative to the base agent. (b) Ablations on \ours{}. (c) Memory transfer across backbones. (d) Grounding accuracy on the real-world subset.}
\label{tab:analysis}
\vspace{-3mm}

\begin{subtable}{\linewidth}
\centering
\renewcommand{\arraystretch}{1.10}
\resizebox{\linewidth}{!}{%
\footnotesize
\begin{tabular}{
l
>{\columncolor{allshade}}C@{\hspace{2pt}}>{\columncolor{allshade}}C@{\hspace{8pt}}
C@{\hspace{2pt}}C@{\hspace{8pt}}
C@{\hspace{2pt}}C@{\hspace{8pt}}
C@{\hspace{2pt}}C@{\hspace{8pt}}
C@{\hspace{2pt}}C@{\hspace{8pt}}
C@{\hspace{2pt}}C
}
\toprule
& \multicolumn{2}{c}{\textbf{All}}
& \multicolumn{2}{c}{\textbf{G0} Speech}
& \multicolumn{2}{c}{\textbf{G1} Gesture}
& \multicolumn{2}{c}{\textbf{G2} Sound}
& \multicolumn{2}{c}{\textbf{G3} Space}
& \multicolumn{2}{c}{\textbf{G4} Identity} \\
\cmidrule(lr){2-3}\cmidrule(lr){4-5}\cmidrule(lr){6-7}\cmidrule(lr){8-9}\cmidrule(lr){10-11}\cmidrule(lr){12-13}
\textbf{Model} & Gr. & Goal & Gr. & Goal & Gr. & Goal & Gr. & Goal & Gr. & Goal & Gr. & Goal \\
\midrule

\addlinespace[2pt]
& 38.2 & 22.4 & 61.5 & 34.6 & 41.0 & 28.1 & 35.1 & 14.2 & 28.1 & 22.9 & 22.9 & 11.7 \\[-1pt]
\multirow{-2}{*}{Qwen2.5-Omni + \textbf{\ours}}
& \gain{+5.2} & \gain{+4.8}
& \gain{+6.0} & \gain{+9.1}
& \gain{+10.1} & \gain{+8.7}
& \gain{+7.0} & \gain{+0.0}
& \gain{+0.9} & \gain{+5.5}
& \gain{+2.6} & \gain{+0.8} \\

\addlinespace[2pt]
& 74.0 & 59.3 & 86.2 & 73.4 & 65.3 & 58.3 & 89.6 & 70.8 & 75.7 & 58.3 & 55.5 & 37.8 \\[-1pt]
\multirow{-2}{*}{Gemma4 + \textbf{\ours}}
& \gain{+23.4} & \gain{+34.1}
& \gain{+16.9} & \gain{+28.1}
& \gain{+18.8} & \gain{+39.2}
& \gain{+34.4} & \gain{+41.6}
& \gain{+20.8} & \gain{+32.3}
& \gain{+27.4} & \gain{+31.8} \\

\addlinespace[2pt]
& 65.2 & 49.1 & 88.8 & 66.7 & 64.9 & 55.6 & 70.1 & 53.8 & 70.8 & 57.6 & 33.9 & 16.9 \\[-1pt]
\multirow{-2}{*}{Nemotron + \textbf{\ours}}
& \gain{+2.5} & \gain{+4.6}
& \gain{+2.3} & \gain{+5.8}
& \loss{-0.4} & \gain{+3.2}
& \gain{+2.7} & \gain{+6.6}
& \gain{+5.9} & \gain{+6.2}
& \gain{+2.1} & \gain{+1.8} \\

\addlinespace[2pt]
& 70.2 & 53.9 & 87.5 & 69.8 & 70.1 & 58.3 & 86.5 & 68.8 & 77.1 & 62.2 & 35.4 & 17.2 \\[-1pt]
\multirow{-2}{*}{Qwen3-Think + \textbf{\ours}}
& \gain{+4.3} & \gain{+5.5}
& \loss{-0.3} & \gain{+2.4}
& \gain{+4.1} & \gain{+1.7}
& \gain{+6.6} & \gain{+11.9}
& \gain{+4.9} & \gain{+7.0}
& \gain{+6.8} & \gain{+5.5} \\

\addlinespace[2pt]
& 78.2 & 62.3 & 91.9 & 76.6 & 72.2 & 62.2 & 81.2 & 63.5 & 77.8 & 65.6 & 67.2 & 44.5 \\[-1pt]
\multirow{-2}{*}{Gemini-2.5-Flash + \textbf{\ours}}
& \gain{+4.5} & \gain{+6.1}
& \loss{-2.1} & \loss{-0.2}
& \gain{+2.8} & \gain{+2.1}
& \gain{+6.9} & \gain{+8.6}
& \loss{-0.3} & \gain{+2.4}
& \gain{+14.6} & \gain{+15.9} \\

\addlinespace[2pt]
& 85.0 & 69.2 & 96.6 & 79.4 & 74.3 & 66.3 & 86.8 & 68.4 & 87.8 & 73.6 & 78.1 & 58.3 \\[-1pt]
\multirow{-2}{*}{Gemini-2.5-Pro + \textbf{\ours}}
& \gain{+4.0} & \gain{+6.1}
& \gain{+1.5} & \loss{-1.1}
& \loss{-0.4} & \loss{-0.4}
& \gain{+4.9} & \gain{+18.1}
& \gain{+4.8} & \gain{+3.8}
& \gain{+8.6} & \gain{+10.6} \\

\bottomrule
\end{tabular}
}
\vspace{0.1mm}
\caption{Results across grounding types (G0--G4).}
\label{tab:prome_results}
\end{subtable}

\vspace{-1mm}

% ================= (b)(c)(d) Ablation / transfer =================
\footnotesize
\setlength{\tabcolsep}{2pt}
\begin{subtable}[c]{0.33\linewidth}
  \centering
  \begin{tabular}{lcccc}
    \toprule
    & \multicolumn{2}{c}{\scriptsize Gemini-2.5-Flash} & \multicolumn{2}{c}{\scriptsize Gemini-2.5-Pro} \\
    \cmidrule(lr){2-3} \cmidrule(lr){4-5}
    Method & Gr. & Goal & Gr. & Goal \\
    \midrule
    Base         & 73.7 & 56.2 & 81.0 & 63.1 \\
    +Tool           & \underline{75.2} & \underline{58.9} & \underline{83.9} & \underline{65.9} \\
    +\textbf{\ours} & \textbf{78.2} & \textbf{62.3} & \textbf{85.0} & \textbf{69.2} \\
    \bottomrule
  \end{tabular}
  \vspace{0.1mm}
  \caption{Ablation of \ours{}.}
  \label{tab:overall}
\end{subtable}\hfill
\begin{subtable}[c]{0.33\linewidth}
  \centering
  \begin{tabular}{lcccc}
    \toprule
    & \multicolumn{2}{c}{\scriptsize Gemma4} & \multicolumn{2}{c}{\scriptsize Nemotron3-Omni} \\
    \cmidrule(lr){2-3} \cmidrule(lr){4-5}
    Method & Gr. & Goal & Gr. & Goal \\
    \midrule
    Base         & 50.6 & 25.2 & 62.7 & 44.5 \\
    +Trans.\ Mem    & \underline{71.7} & \underline{51.7} & \underline{64.9} & \underline{48.7} \\
    +\textbf{\ours} & \textbf{74.0} & \textbf{59.3} & \textbf{65.2} & \textbf{49.1} \\
    \bottomrule
  \end{tabular}
  \vspace{0.1mm}
  \caption{Memory transfer.}
  \label{tab:mem_transfer}
\end{subtable}\hfill
\begin{subtable}[c]{0.31\linewidth}
  \centering
  \includegraphics[width=\linewidth]{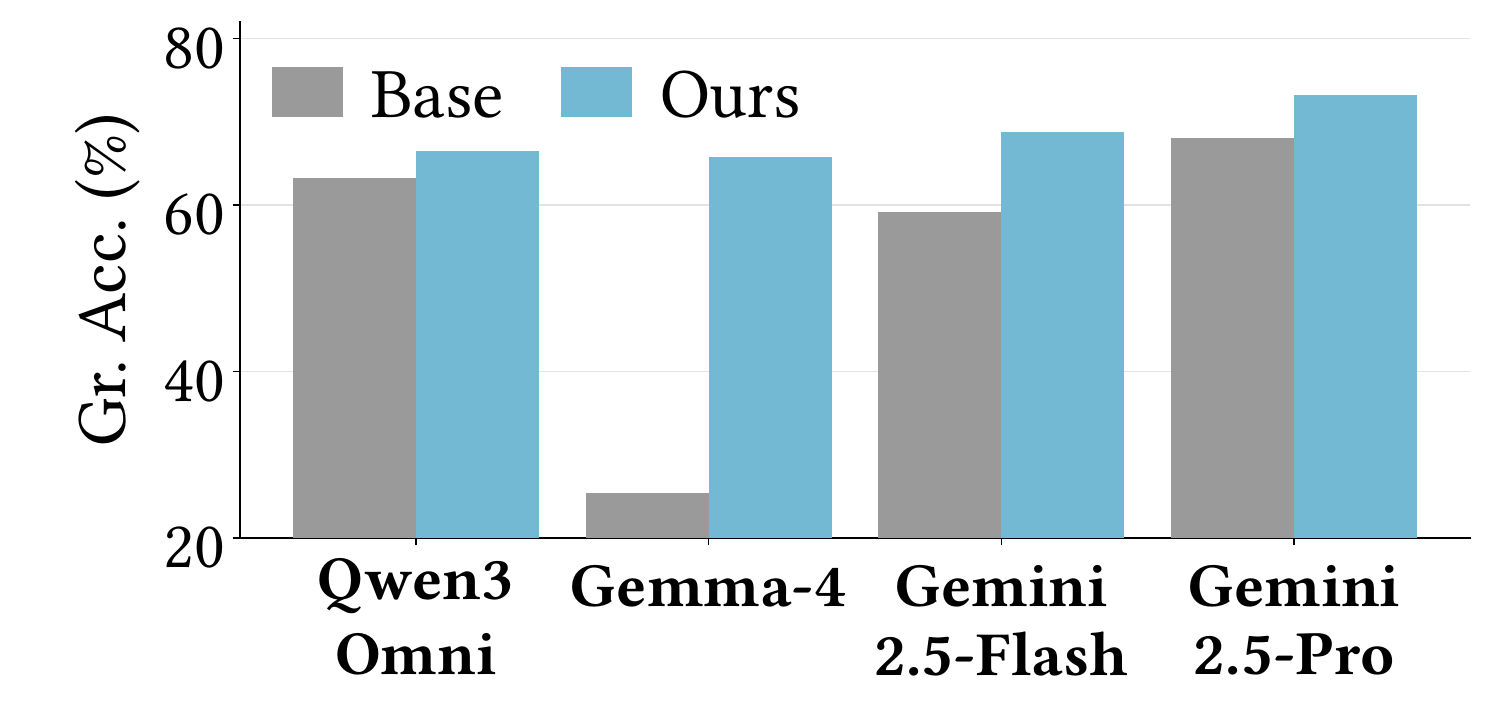}
  \vspace{0.1mm}
  \caption{Real-world transfer.}
  \label{fig:realworld}
  
\end{subtable}
\vspace{-10mm}
\end{table*}

\subsection{Results on \ours{}}
As shown in \cref{tab:prome_results}, \ours{} improves overall grounding and goal accuracy across six Omni-LLMs. Notably, Gemma4, the weakest larger open-source model, gains 23.4/34.1 percentage points in grounding/goal accuracy, respectively, surpassing the other open-source base agents. \ours{} also improves the strong Gemini models by 4.5/6.1 points on Gemini-2.5-Flash and 4.0/6.1 points on Gemini-2.5-Pro.
Across grounding types, as expected, gains are limited on G0 but larger on G1--G4.

\label{sec:ablation}
\newpara{Ablation studies on \ours{}.}
\cref{tab:overall} isolates the effects of the perception tools and procedural memory on Gemini-2.5-Flash and Gemini-2.5-Pro. Perception tools alone provide modest gains on both models, while adding procedural memory yields further improvements.

\newpara{Memory transfer.} \cref{tab:mem_transfer} applies procedural memory constructed with Qwen3-Omni-Think to Gemma4 and Nemotron3-Omni. Although each model performs best with its own memory, the transferred memory still improves over the base agent, showing that procedural memory transfers across models.

\newpara{Real-world generalization.} \cref{fig:realworld} reports grounding accuracy on the real-world subset. Although the procedural memory is learned entirely from synthetic data, \ours{} consistently improves grounding accuracy, suggesting that our synthetic episodes capture challenges that transfer to real-world interactions.
\section{Conclusion}
% We introduce \name{}, a benchmark for smart-home agents in which the information needed to resolve a user's request is distributed across modalities rather than stated in text. The benchmark spans five grounding types (speech, gesture, sound, space, and identity), four query types, and both feasible and infeasible requests. Evaluating 14 audio-visual LLMs shows that while most models resolve speech-grounded requests reliably, accuracy drops sharply once the target must be inferred from gesture, sound, spatial, or identity cues, with gesture- and identity-grounded requests remaining the most challenging even for proprietary models. We propose \ours{}, a lightweight, model-agnostic agentic system that pairs specialized perception tools with a state-conditioned procedural memory learned from synthetic experience. \ours{} consistently improves both grounding and goal accuracy. Future work includes extending \name{} to additional grounding types and multi-turn interactions, and allowing the procedural memory of \ours{} to keep evolving from deployment-time experience rather than being frozen after training.

We introduce OmniSmartHome, a multimodal smart-home agent benchmark in which the information required to resolve a user request is distributed across modalities rather than explicitly stated in language. OmniSmartHome pairs each spoken request with visual and spatial-audio observations, covers both synthetic and real-world episodes, and spans five grounding types, four query types, and both feasible and infeasible requests. Our evaluation of 16 Omni-LLMs shows that current models reliably handle requests explicitly stated in speech but struggle when the target device is inferred from multimodal contextual cues. To narrow this gap, we propose PROME, a simple, model-agnostic agent baseline that combines specialized perception tools with a procedural memory guiding tool use. PROME consistently improves both grounding and goal accuracy. We envision OmniSmartHome as a step toward multimodal smart-home assistants that understand requests as naturally as people make them and, ultimately, act proactively on what they see and hear.

\subsection*{AI use statement}

% (This section is \textbf{required} and does not count toward the page limit.)

In this work, we used generative AI tools as part of our benchmark evaluation procedure, such as the use of LLM-based judges, and to generate synthetic datasets.
We did not use generative AI tools to develop theoretical models or conceptual frameworks, formulate mathematical claims, provide critical ingredients for proving mathematical claims, assist in writing proofs, propose or refine hypotheses, design or provide feedback on research methodology or experiments, implement methods, assist with translation, clean or reformat datasets, support qualitative or thematic data analysis, or interpret results. Additionally, we used generative AI tools to improve the readability of the manuscript, draft parts of the paper, and assist in the development of a web-based interface used for human verification and human performance evaluation. We reviewed all AI-assisted work. In particular, AI-generated code and text were manually reviewed, verified, and revised by the authors. Human verification was also conducted for relevant evaluation procedures and artifacts. We take full responsibility for the final content of this work, including all text, claims, code, and artifacts produced with the aid of generative AI.
% See the ICLR 2027 AI Policy for Authors for more details. This statement should
% not be more than 1 page.

% \subsection*{Ethics statement}

% If authors feel that their paper submission raises questions regarding the Code of Ethics, they are encouraged to include a paragraph of Ethics Statement (at
% the end of the main text before references) to address potential concerns where
% appropriate. Topics include, but are not limited to, studies that involve human
% subjects, practices to data set releases, potentially harmful insights,
% methodologies and applications, potential conflicts of interest and sponsorship,
% discrimination/bias/fairness concerns, privacy and security issues, legal
% compliance, and research integrity issues (e.g., IRB, documentation, research
% ethics). This statement should not be more than 1 page.

\subsection*{Reproducibility statement}

Details on the models used in our experiments, including their providers and access methods, are provided in Appendix \ref{app:api_usage}. LLM prompts are in Appendix \ref{app:prompt}. We also release the relevant implementation code in the supplementary material to facilitate reproducibility.

% It is important that the work published in ICLR is reproducible. Authors are
% strongly encouraged to include a paragraph-long Reproducibility Statement at the
% end of the main text (before references) to discuss the efforts that have been
% made to ensure reproducibility. This paragraph should not itself describe
% details needed for reproducing the results, but rather reference the parts of
% the main paper, appendix, and supplemental materials that will help with
% reproducibility. For example, for novel models or algorithms, a link to an
% anonymous downloadable source code can be submitted as supplementary materials;
% for theoretical results, clear explanations of any assumptions and a complete
% proof of the claims can be included in the appendix; for any datasets used in
% the experiments, a complete description of the data processing steps can be
% provided in the supplementary materials. Each of the above are examples of
% things that can be referenced in the reproducibility statement.



\newpage
\bibliography{shorstrings,iclr2027_conference}
\bibliographystyle{iclr2027_conference}

\newpage
\appendix

\renewcommand{\theHsection}{appendix.\Alph{section}}
\renewcommand{\theHsubsection}{appendix.\Alph{section}.\arabic{subsection}}

\counterwithin{figure}{section}
\counterwithin{table}{section}

\renewcommand{\thefigure}{\thesection.\arabic{figure}}
\renewcommand{\thetable}{\thesection.\arabic{table}}

\makeatletter
\providecommand{\authcount}[1]{}
\makeatother

\begin{center}
    \Large \textbf{Appendix} \\
\end{center}

% Table of contents
\startcontents[supp]
\hypersetup{linkcolor=black}
% \printcontents[supp]{l}{1}{\setcounter{tocdepth}{2}}
\printcontents[supp]{l}{1}{
\setcounter{tocdepth}{2}
\setlength{\baselineskip}{1\baselineskip}
}

% \vspace{10mm}
\clearpage

% \section{Appendix}
\section{Details on \name{}}
\label{sec:benchmark_detail}

\subsection{Audio-Visual Scene Synthesis}
\label{sec:audio_pipeline}

\newpara{Speech and device sound sources.}
All spoken utterances are synthesized with MOSS-TTSD~\cite{zhang2026moss},
which natively outputs 24\,kHz waveforms. Device sounds are drawn from a
curated bank of 42 recordings covering every device--state pair referenced
by the benchmark episodes (e.g., washer spin cycles, microwave beeps, fire
alarms). The original recordings vary in format (44.1, 48, and 96\,kHz;
mono and stereo); since SoundSpaces models each emitter as a point source,
we downmix stereo files to mono and resample all sources to 24\,kHz prior
to spatialization.

\newpara{Spatial rendering.}
For each synthetic episode, we render room impulse responses (RIRs) with
SoundSpaces~2.0~\cite{chen22soundspaces2} on the corresponding HM3D
scene~\cite{ramakrishnan2021habitat}, setting the rendering sample rate to
24\,kHz and using a four-channel first-order ambisonics (FOA) receiver at
the agent position. Each mono source is convolved with its FOA RIR at the
matched rate, and the resulting episode audio is released as 24\,kHz
four-channel FOA WAV files. 

\newpara{Real-world recordings.}
The 272 real-world episodes are captured with a Zoom H3-VR ambisonic
microphone at 48\,kHz and downsampled to 24\,kHz FOA to match the
synthetic distribution.

\subsection{Details on Benchmark Design}
\begingroup
\footnotesize
\setlength{\tabcolsep}{4pt}
\renewcommand{\arraystretch}{1.3}

\begin{xltabular}{\linewidth}{
    >{\hsize=0.8\hsize\raggedright\arraybackslash}X
    >{\hsize=0.85\hsize\raggedright\arraybackslash}X
    >{\hsize=1.0\hsize\raggedright\arraybackslash}X
    >{\hsize=1.1\hsize\raggedright\arraybackslash}X
    >{\hsize=1.25\hsize\raggedright\arraybackslash}X
}

\caption{\textbf{Benchmark construction by grounding type.}
Conditions used to construct easy, medium, and hard instances,
along with infeasibility conditions for each grounding type.}
\label{tab:grounding_difficulty} \\
\toprule
\textbf{Type} & \textbf{Easy} & \textbf{Medium}
& \textbf{Hard} & \textbf{Infeasible} \\
\midrule
\endfirsthead

\multicolumn{5}{c}{\footnotesize Table \thetable\ (continued)} \\
\toprule
\textbf{Type} & \textbf{Easy} & \textbf{Medium}
& \textbf{Hard} & \textbf{Infeasible} \\
\midrule
\endhead

\midrule
\multicolumn{5}{r}{\footnotesize Table continues on the following page} \\
\endfoot

\bottomrule
\endlastfoot

\textbf{G0:} Speech
& Fluent utterance explicitly naming the device and room.
& Speech disfluencies, such as hesitations and repetitions, are added.
& Background noise is added to the disfluent utterance.
& Named target absent; required actuator unavailable;
physical limits; conflicting temporal constraints. \\

\addlinespace
\textbf{G1:} Gesture
& One device and one person.
& Additional candidate devices are introduced.
& Multiple candidate devices and additional non-speaking people.
& No device or requested capability matching the pointed target;
physical limits; conflicting temporal constraints. \\

\addlinespace
\textbf{G2:} Sound
& One candidate device capable of producing sound.
& Multiple candidate devices, with only one producing sound.
& Sounds from other devices are also present.
& Referenced sound absent;
physical limits; conflicting temporal constraints. \\

\addlinespace
\textbf{G3-V:} Visual Space
& One device and one person.
& Additional candidate devices are introduced.
& Non-speaking people are added, requiring direction interpretation
from the speaker's perspective.
& No device with the requested capability in the referenced direction;
physical limits; conflicting temporal constraints. \\

\addlinespace
\textbf{G3-A:} Auditory Space
& One candidate device capable of producing sound and one person.
& Additional candidate devices are introduced.
& Non-speaking people are added, requiring joint resolution of
the sound source, direction, and speaker.
& Referenced sound absent;
physical limits; conflicting temporal constraints. \\

\addlinespace
\textbf{G4:} Identity
& One person referring to ``my room''.
& Non-speaking people are added, requiring speaker identification
to resolve ``my room''.
& The speaker points to another person and refers to
``his room'' or ``her room''.
& Target device or requested capability absent from the person's room;
required actuator unavailable; physical limits;
conflicting temporal constraints. \\

\end{xltabular}
\endgroup
We define easy, medium, hard, and infeasible settings within each
grounding type by varying candidate devices, distractors, grounding
cues, and task feasibility.
Table~\ref{tab:grounding_difficulty} summarizes these settings.

\newpara{G0: Speech.}
Easy uses fluent, explicit references; medium adds disfluencies;
hard adds background noise. Infeasible cases involve missing
targets or actuators, physical limits, or temporal conflicts.

\newpara{G1: Gesture.}
Easy has one device and person; medium adds candidate devices;
hard adds non-speakers. Infeasible cases involve missing pointed
targets or capabilities, physical limits, or temporal conflicts.

\newpara{G2: Sound.}
Easy has one sound-capable device; medium adds candidate devices but
only one audible source; hard adds another device emitting a different
type of sound, so the agent must disambiguate under a noisy acoustic scene.
Infeasible cases involve absent referenced sounds, physical limits,
or temporal conflicts.

\newpara{G3-V: Visual Space.}
Easy has one device and person; medium adds candidate devices;
hard adds non-speakers to test speaker-relative directions.
Infeasible cases involve no suitable device in the referenced
direction, physical limits, or temporal conflicts.

\newpara{G3-A: Auditory Space.}
Easy has one sound-capable device and person; medium adds candidates;
hard adds non-speakers to test speaker-relative sound localization.
Infeasible cases involve absent referenced sounds, physical limits,
or temporal conflicts.

\newpara{G4: User Identity.}
Easy uses ``my room'' with one person; medium adds non-speakers;
hard uses pointing with ``his/her room''.
Infeasible cases involve missing devices or capabilities in that
person's room, unavailable actuators, physical limits, or temporal conflicts.

\subsection{Examples}
\label{app:examples}
\begin{figure}[p]
  \centering
  \setlength{\tabcolsep}{2pt}
  \begin{tabular}{@{}c*{4}{p{0.22\linewidth}}@{}}
    & \multicolumn{1}{c}{\scriptsize\textbf{Q1} State inquiry}
    & \multicolumn{1}{c}{\scriptsize\textbf{Q2} Explicit device control}
    & \multicolumn{1}{c}{\scriptsize\textbf{Q3} Workflow scheduling}
    & \multicolumn{1}{c}{\scriptsize\textbf{Q4} Implicit intent} \\[2pt]
    \exrow{G0}{Speech} &
      \excell{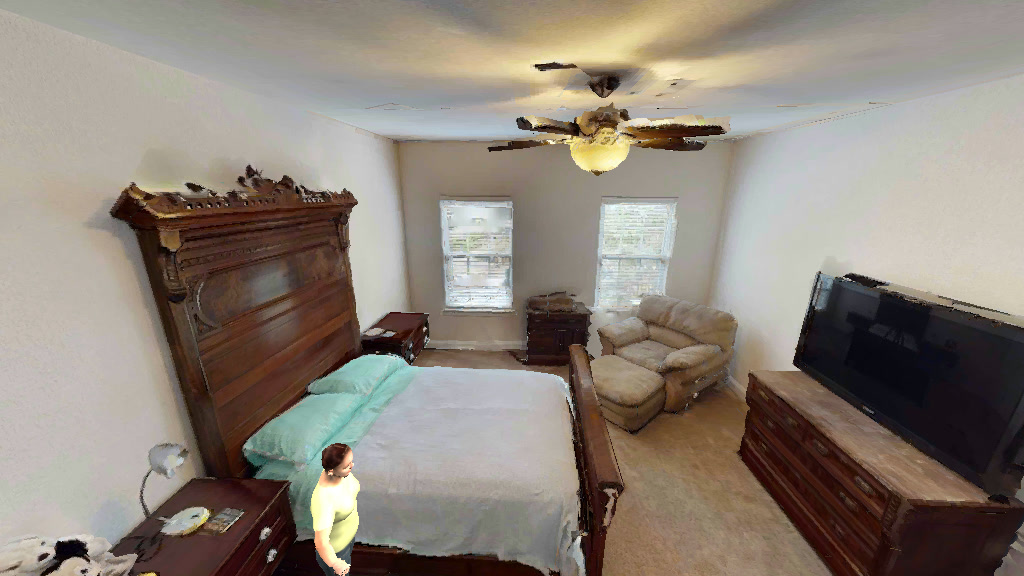}{Hey, can you tell me what the remaining time is on the tv in the living room right now?}{TV} &
      \excell{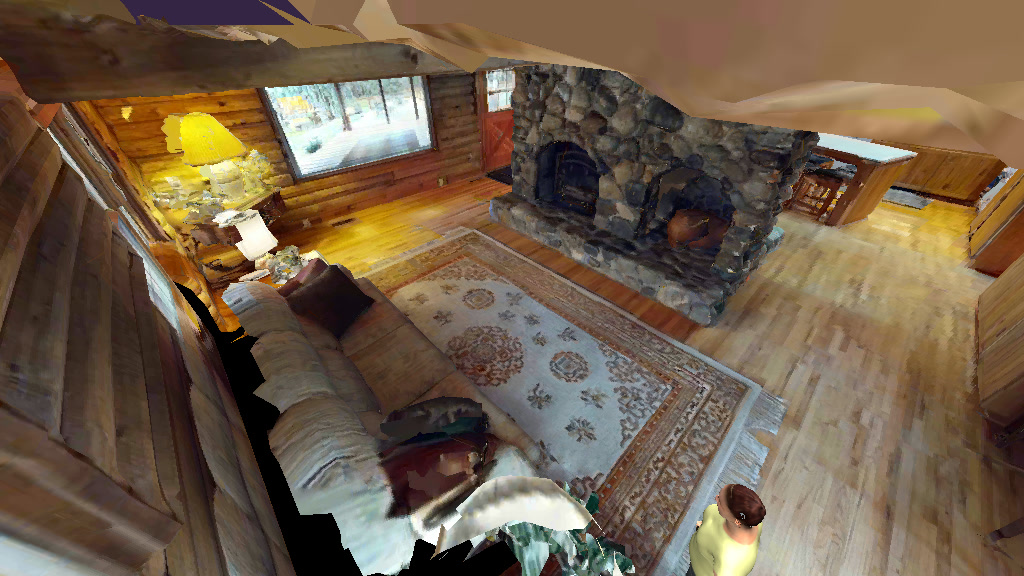}{Hey, turn on the dimmable light 1 in the living room and set its brightness to 100 percent.}{Dimmable light} &
      \excell{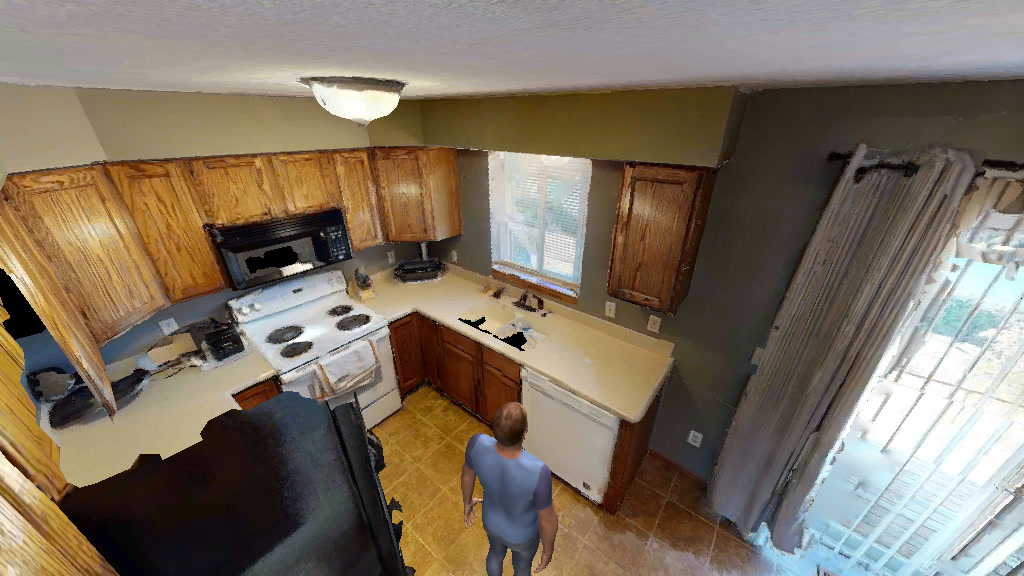}{Hey, schedule a pause for the dishwasher in the kitchen in 5 minutes, do not pause it now.}{Dishwasher} &
      \excell{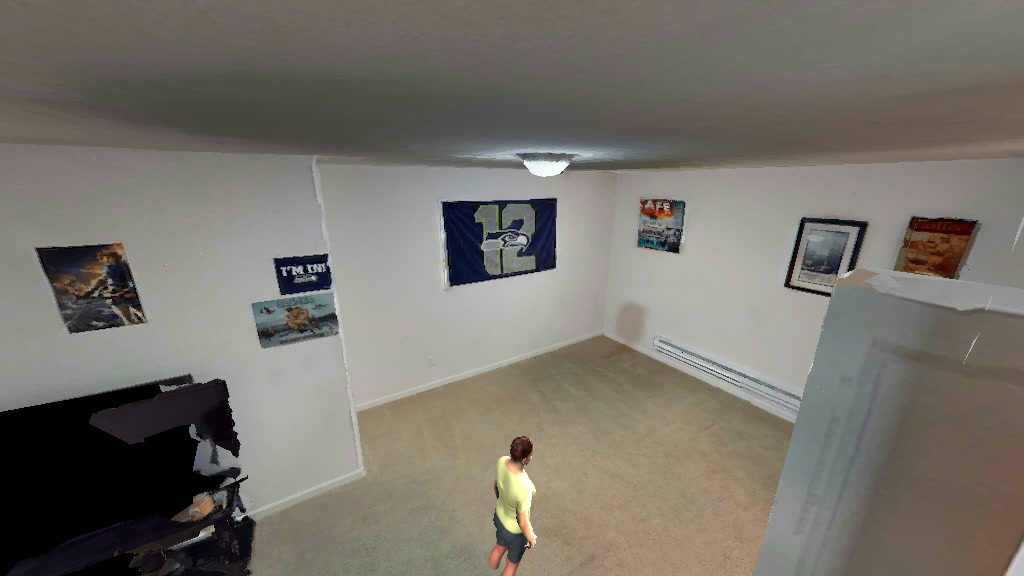}{Ugh, I am a bit too warm in the living room. My face feels flushed. I am sweating a little and getting sluggish from the heat.}{Air conditioner} \\[8pt]
    \exrow{G1}{Gesture} &
      \excell{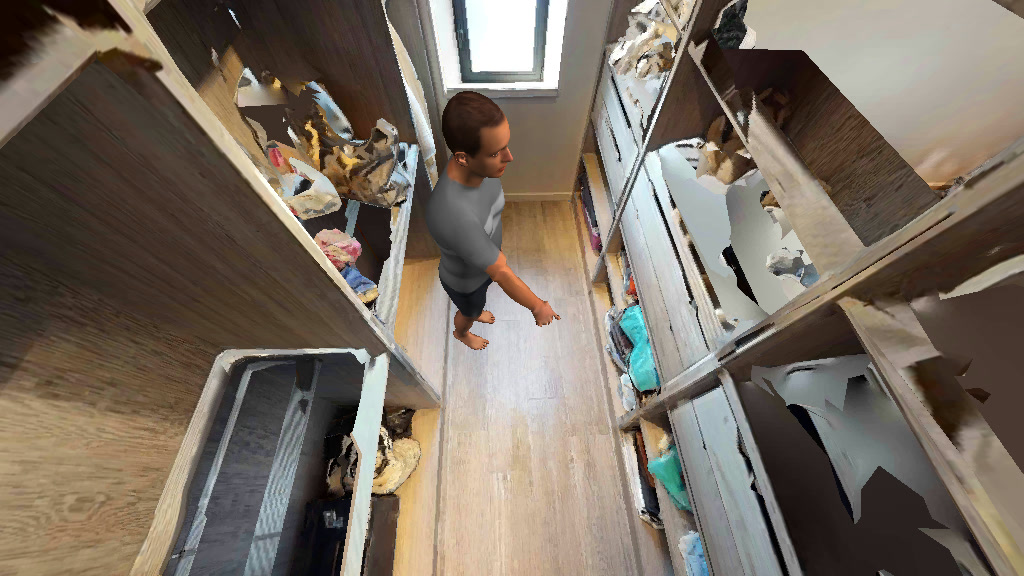}{Which operating phases does this support?}{Robot vacuum} &
      \excell{syn_g1_q2.jpg}{Play this.}{TV} &
      \excell{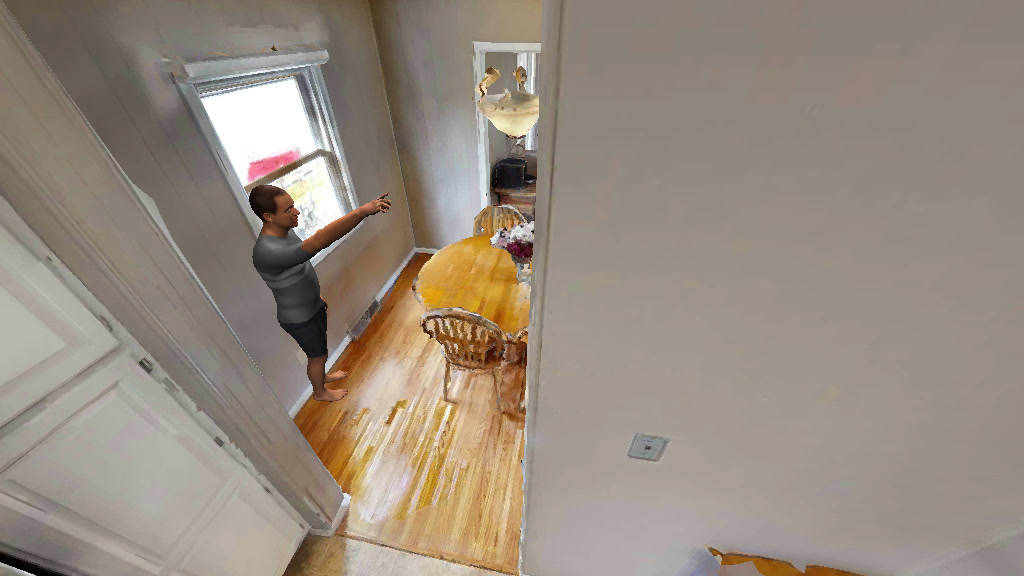}{Turn this on in five minutes.}{On/off light} &
       \\[8pt]
    \exrow{G2}{Sound} &
      \excell{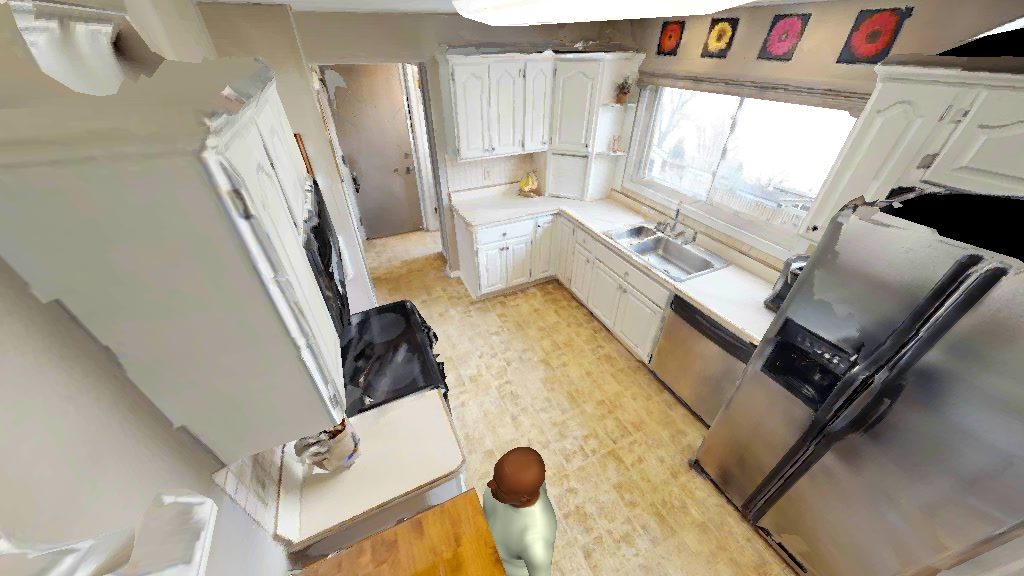}{I hear that machine noise. What is the product name of the device making that machine noise?}{Dishwasher} &
      \excell{syn_g2_q2.jpg}{Hey, stop that whirring sound right now.}{Fan} &
      \excell{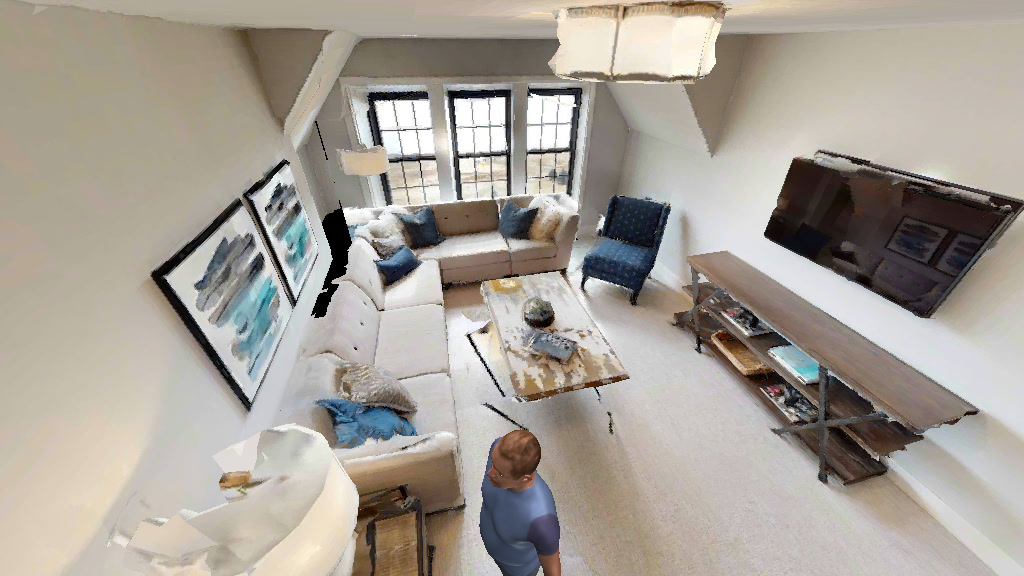}{Schedule setting the volume of that upbeat music to 70 percent in 12 minutes and do not do it now}{TV} &
       \\[8pt]
    \exrow{G3}{Space} &
      \excell{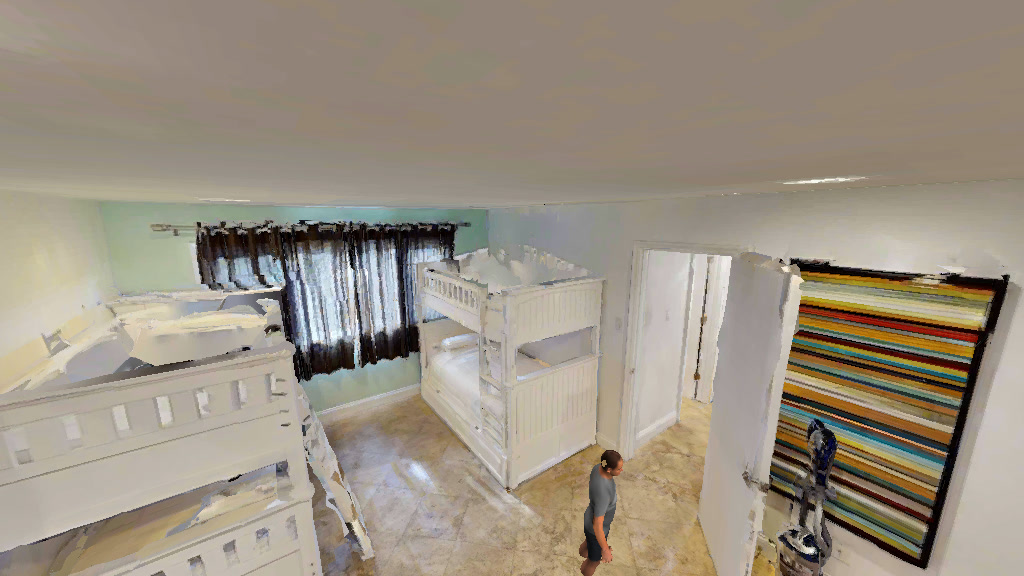}{Um, can you tell me what operational states are available for the device making that sound in front of me?}{Robot vacuum} &
      \excell{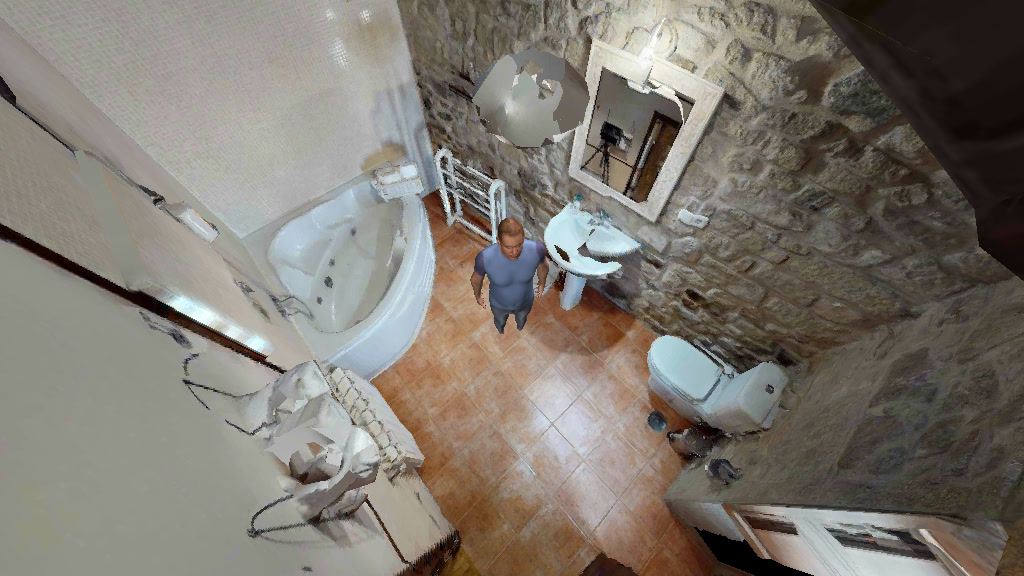}{Turn on the device above my head now.}{On/off light} &
      \excell{syn_g3_q3.jpg}{Schedule the thing making that sound on my right to stop running in 13 minutes and do not stop it now.}{Dishwasher} &
       \\[8pt]
    \exrow{G4}{Identity} &
      \excell{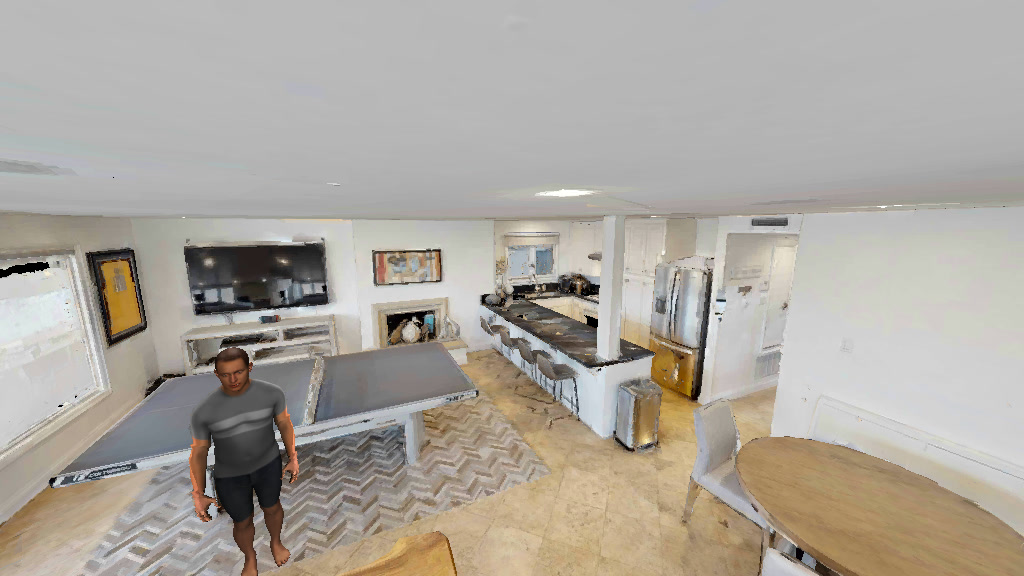}{Is the coffee machine in my room running?}{Coffee machine in Grandpa's room} &
      \excell{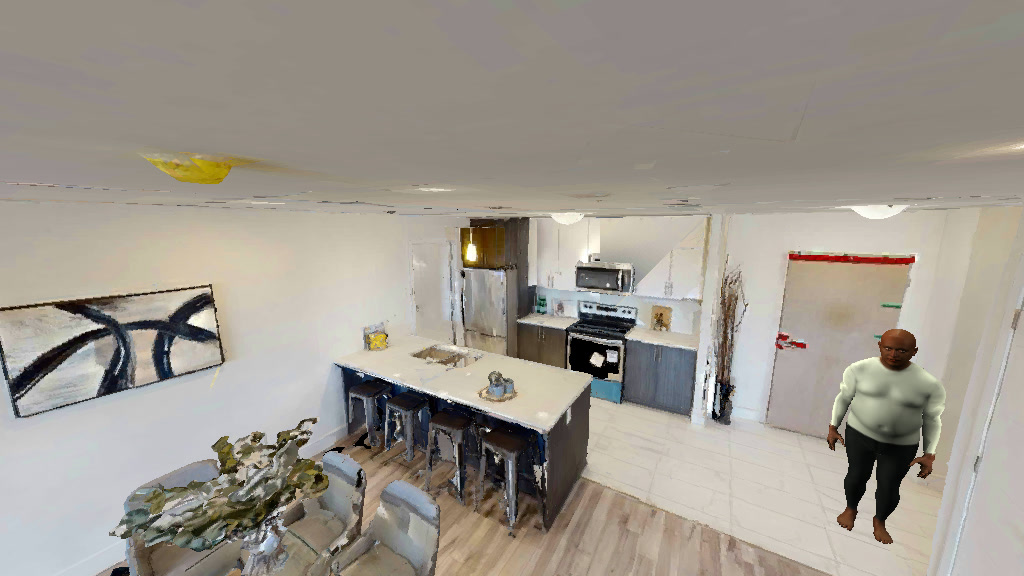}{Turn the dishwasher in my room on and start it running.}{Dishwasher in Dad's room} &
      \excell{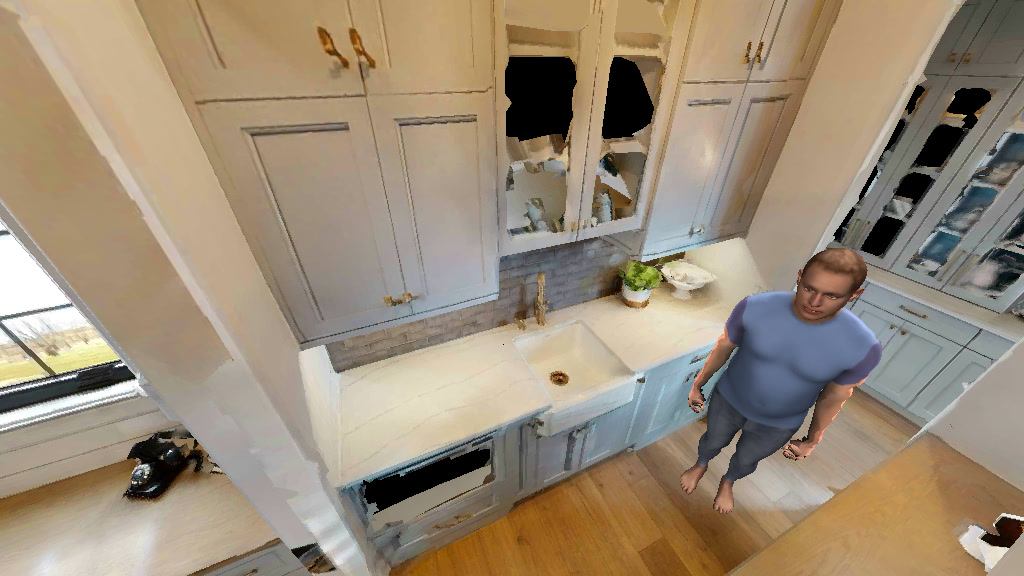}{After dinner, exactly 12 minutes after dishwasher in the bathroom finishes cleaning, power off the microwave in my room.}{Microwave in Dad's room} &
      \excell{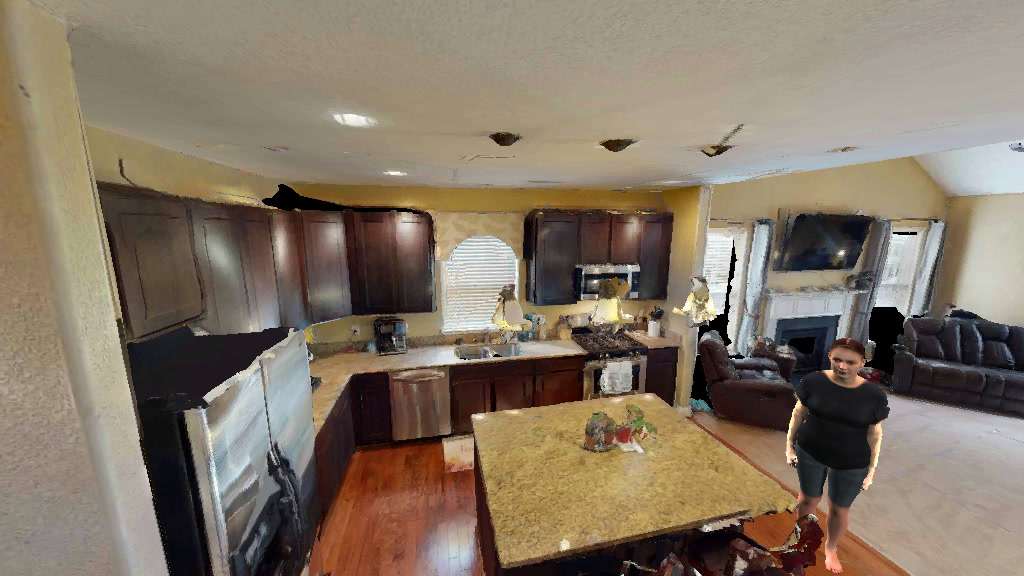}{My room is too dark.}{On/off light in Grandma's room} \\[8pt]
  \end{tabular}
  \caption{\textbf{Synthetic examples}, one per grounding type (rows) and query type (columns).
  Each cell shows the rendered observation, the spoken request (delivered to the agent only as
  spatial audio), and the ground-truth target device. Q4 is instantiated only for G0 and G4.}
  \label{fig:syn_examples}
\end{figure}

\newpara{Synthetic Data}
\Cref{fig:syn_examples} shows one synthetic episode for every valid combination of grounding type
(G0--G4) and query type (Q1--Q4). Each episode consists of a single rendered view of the room,
paired with a spatial audio recording (binaural and first-order Ambisonics) that contains the
user's spoken request and, where relevant, the sound emitted by a device; the transcript printed
under each image is \emph{not} given to the agent.
The grounding type determines which part of this observation carries the missing information.
In G0 the utterance names the device and its room outright, so correct speech recognition suffices.
In G1 the utterance is purely deictic (``Play this.''), and the target is the device that the
speaker's outstretched arm points at.
In G2 the device is identified only by the sound it makes (``stop that whirring sound''), so the
agent must recognize the sound and localize its source among the devices in view.
In G3 the device is described relative to the speaker's own body (``above my head'', ``on my
right''), which requires estimating where the speaker stands and which way they face; the
direction is either combined with a device sound or stands alone.
In G4 the request refers to ``my room'': the agent has to recognize who is speaking from their
face and voice, and then look up which room belongs to that household member, typically a room
other than the one being observed.
Implicit-intent queries (Q4) state a discomfort rather than a command (``My room is too dark.''),
and the target is whichever device in the resolved room can relieve it.

\begin{figure}[t]
  \centering
  \setlength{\tabcolsep}{2pt}
  \begin{tabular}{@{}c*{3}{p{0.31\linewidth}}@{}}
    & \multicolumn{1}{c}{\scriptsize G1 Gesture} & \multicolumn{1}{c}{\scriptsize G2 Sound}
    & \multicolumn{1}{c}{\scriptsize G3 Space} \\[1pt]
    \rotatebox{90}{\scriptsize\hspace{0.8em}Observation} &
      \includegraphics[width=\linewidth]{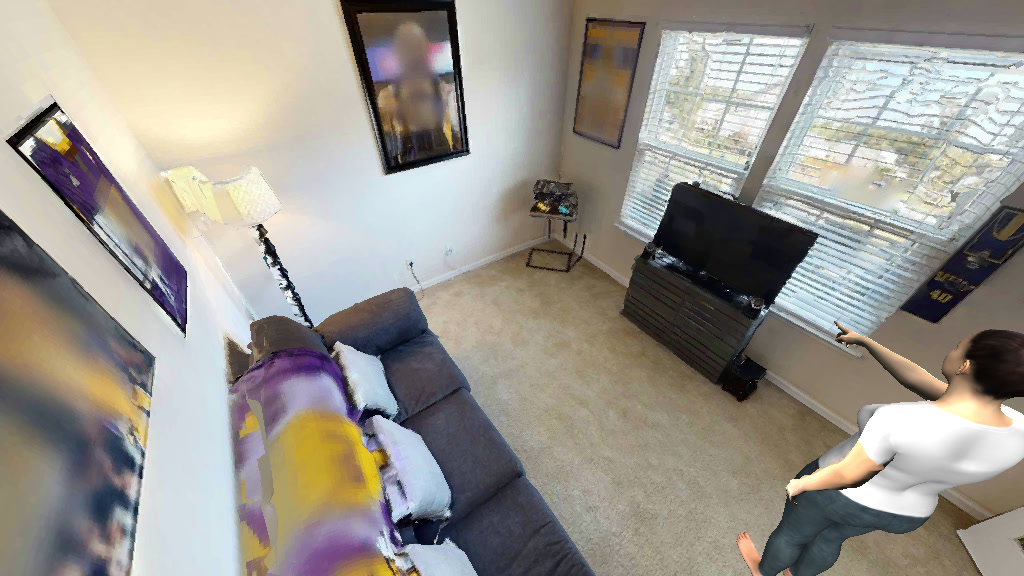} &
      \includegraphics[width=\linewidth]{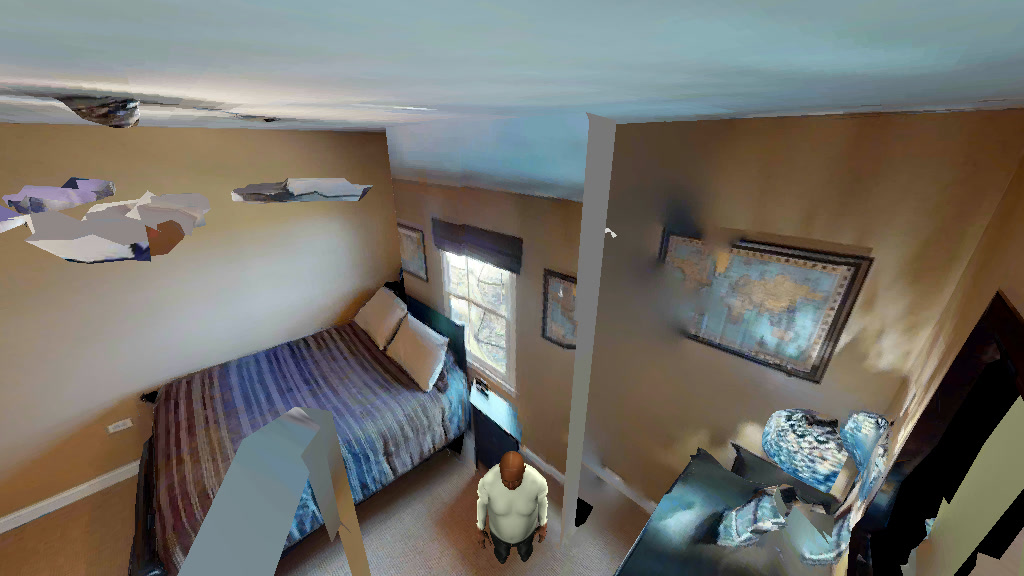} &
      \includegraphics[width=\linewidth]{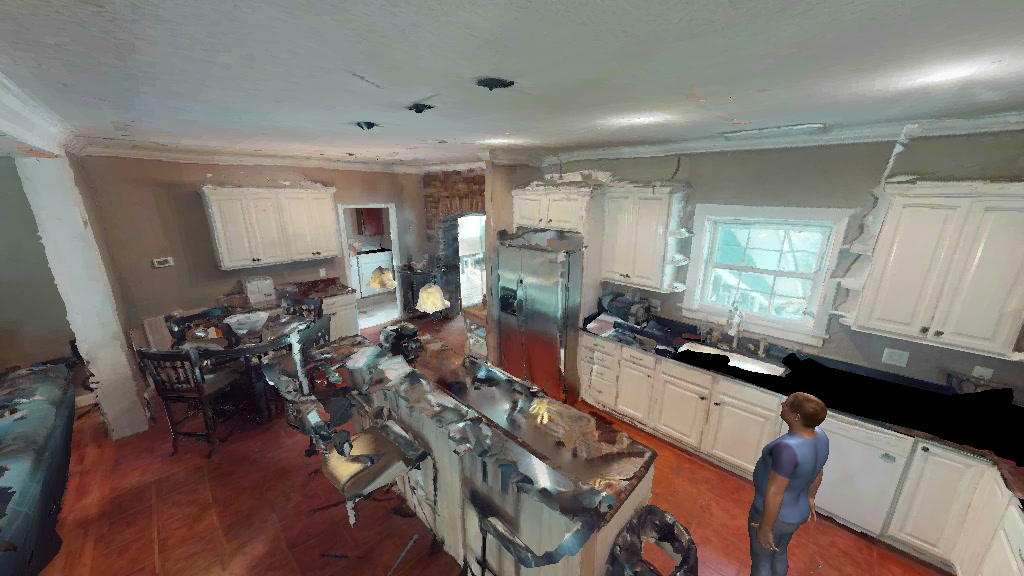} \\[2pt]
    \rotatebox{90}{\scriptsize\hspace{1.2em}Reference} &
      \includegraphics[width=\linewidth]{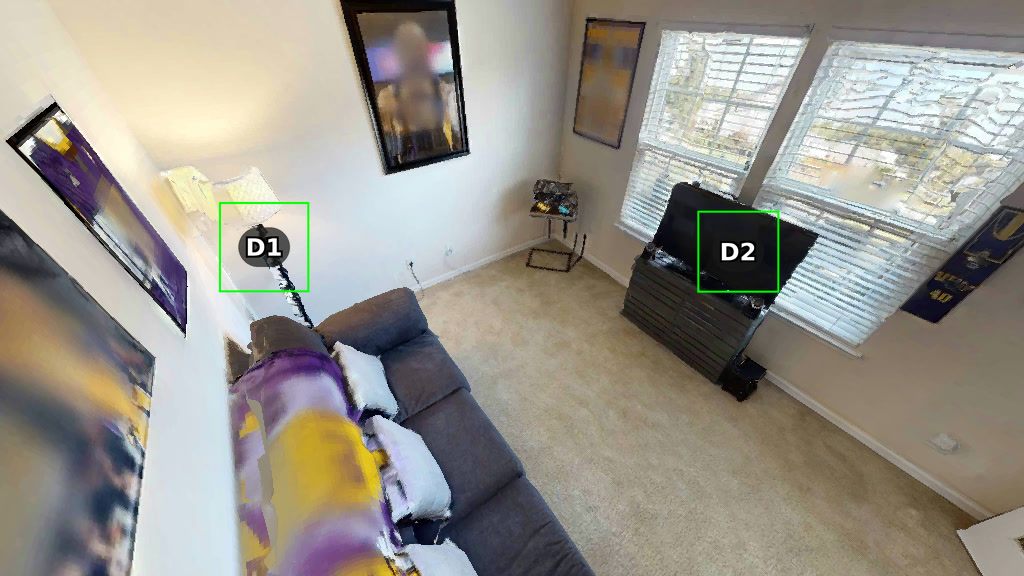} &
      \includegraphics[width=\linewidth]{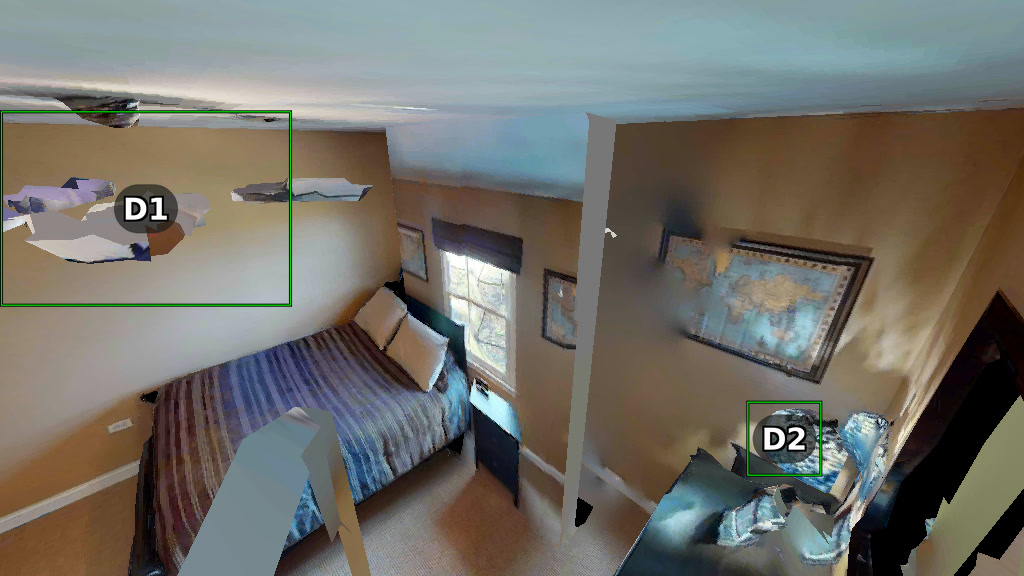} &
      \includegraphics[width=\linewidth]{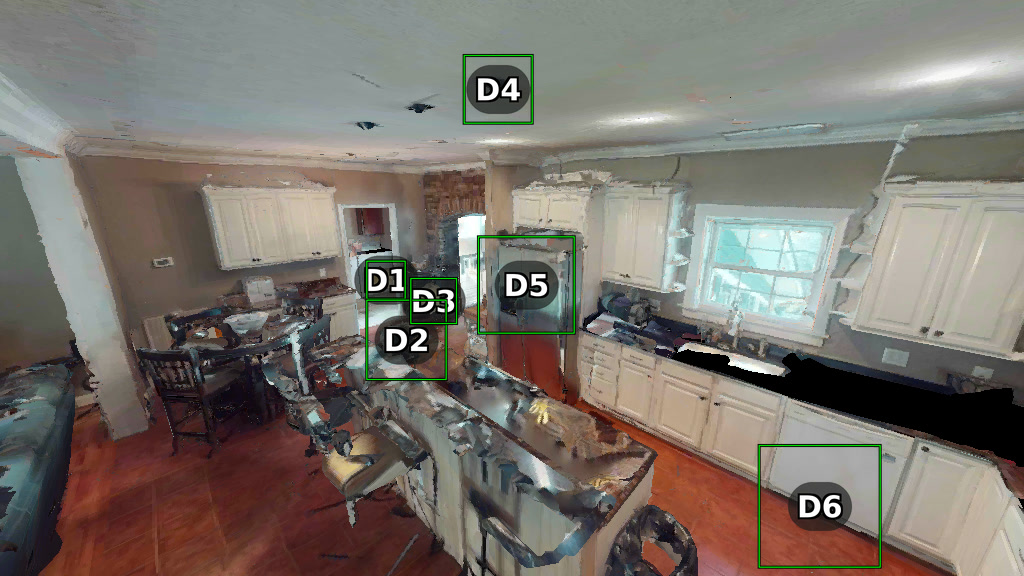} \\[3pt]
    % \rotatebox{90}{\scriptsize\hspace{1.6em}Faces} &
    %   \multicolumn{3}{c}{\includegraphics[width=0.94\linewidth]{asset/examples/syn_g4_q1_gallery.jpg}} \\
  \end{tabular}
  \caption{\textbf{Reference images.} Top: the observation the agent receives. Bottom: the
  reference image of the same room, rendered without people, in which each device carries the
  identifier used in the device list. }
  \label{fig:ref_examples}
\end{figure}

Alongside the raw observation, the simulator exposes a \emph{reference image} of the room
(\cref{fig:ref_examples}). It is rendered from the same camera without any person, and every
controllable device is outlined with a box and an identifier (D1, D2, \dots) that matches the
device list returned for that room. The reference image tells the agent \emph{where each device
is}, but not \emph{which one the user means}: deciding that the pointing arm, the beeping sound or
the speaker's left-hand side coincides with a particular box is exactly the grounding problem the
benchmark measures..

\begin{figure}[t]
  \centering
  \setlength{\tabcolsep}{2pt}
  \begin{tabular}{@{}*{5}{p{0.186\linewidth}}@{}}
      \excellreal{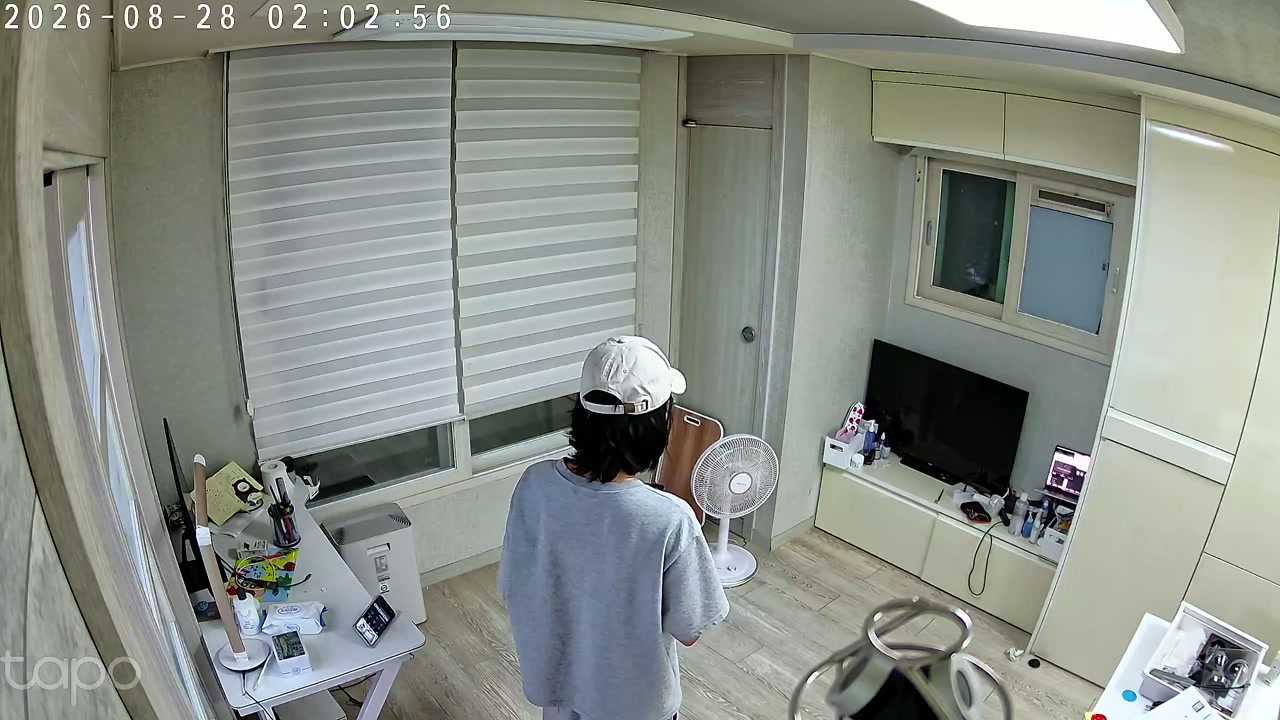}{G0 Speech $\cdot$ Q2}{Hey, switch the channel of the TV in here to SBS.}{TV} &
      \excellreal{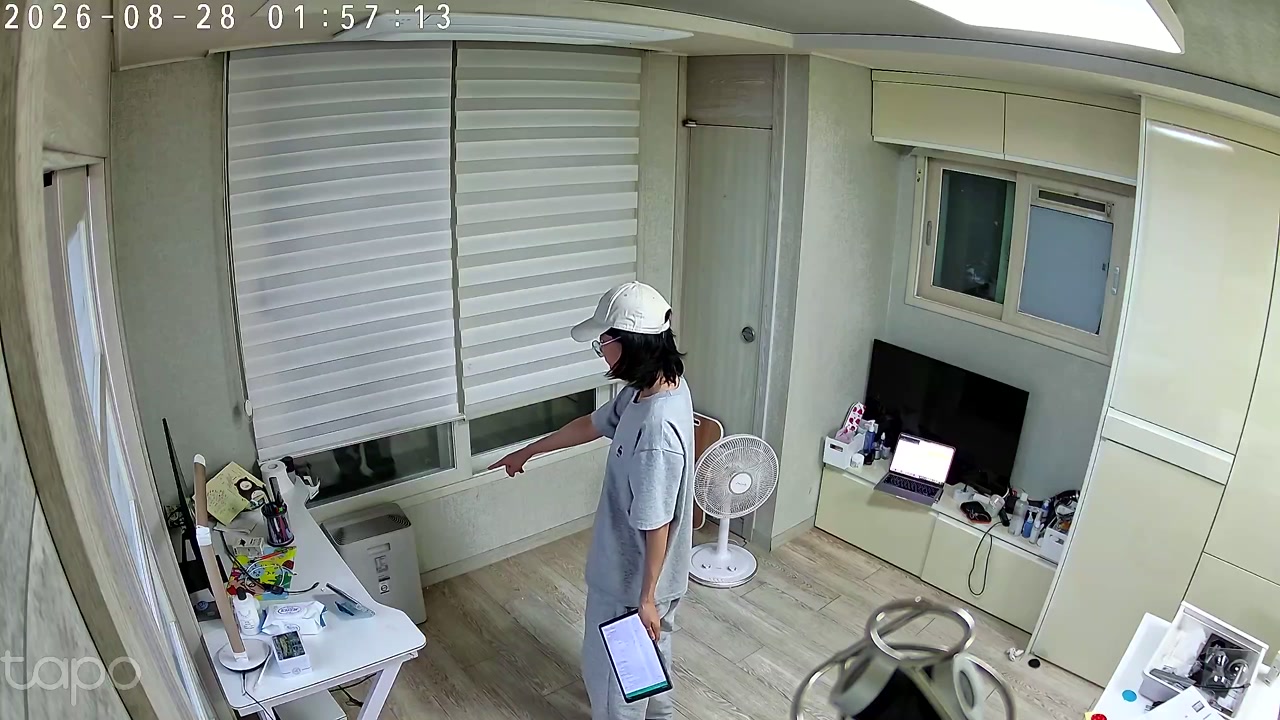}{G1 Gesture $\cdot$ Q3}{In 10 minutes, power this on.}{Air purifier} &
      \excellreal{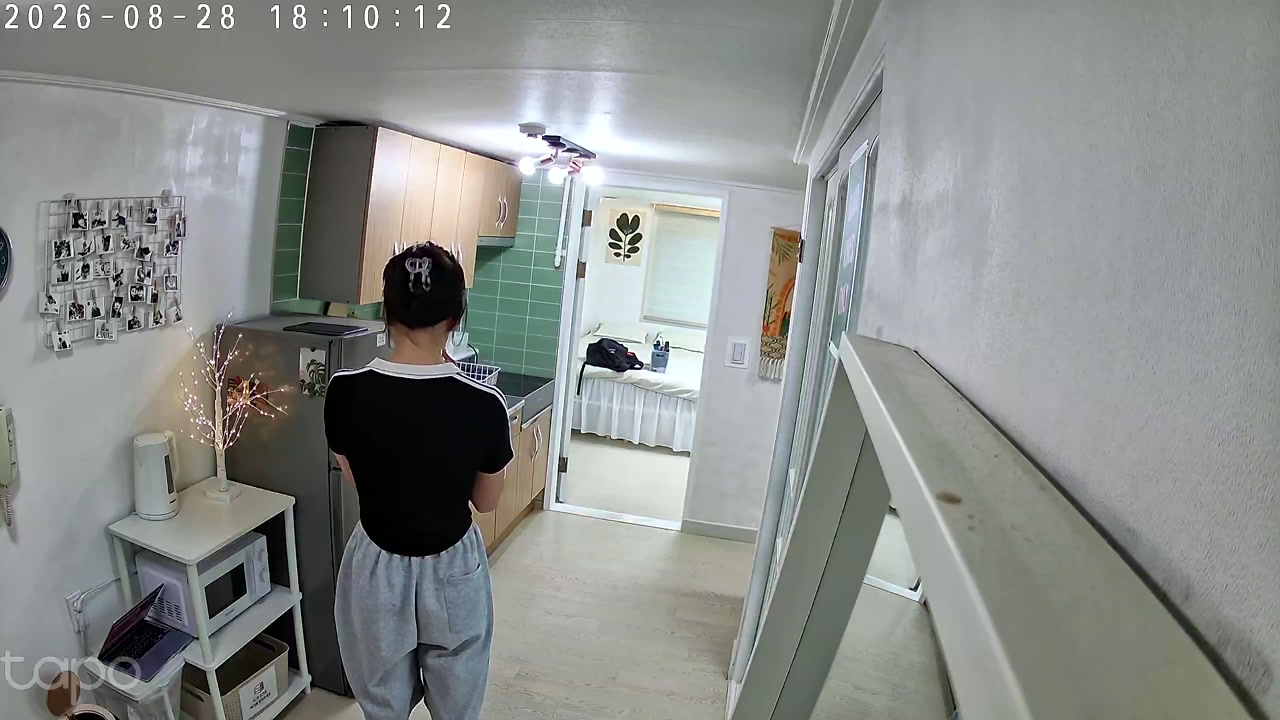}{G2 Sound $\cdot$ Q1}{What's the product name of the device making that beeping sound?}{Microwave} &
      \excellreal{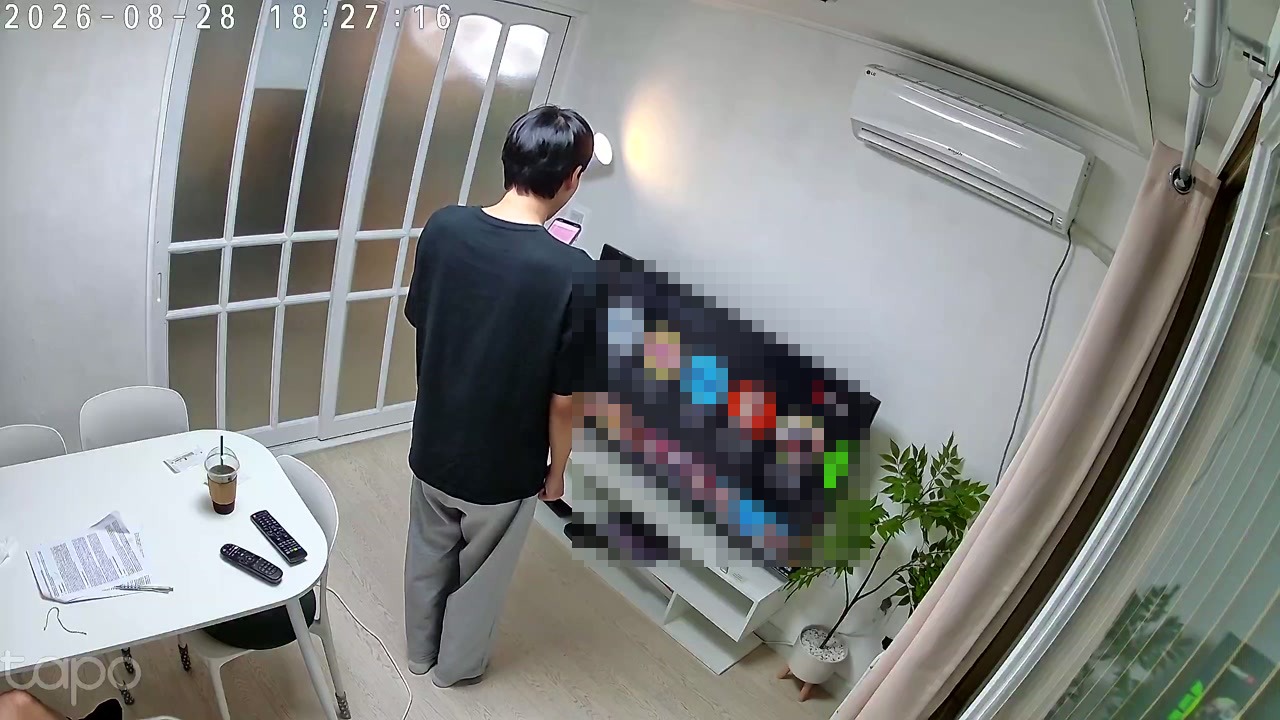}{G3 Space $\cdot$ Q1}{What brand makes the device in front of me?}{TV} &
      \excellreal{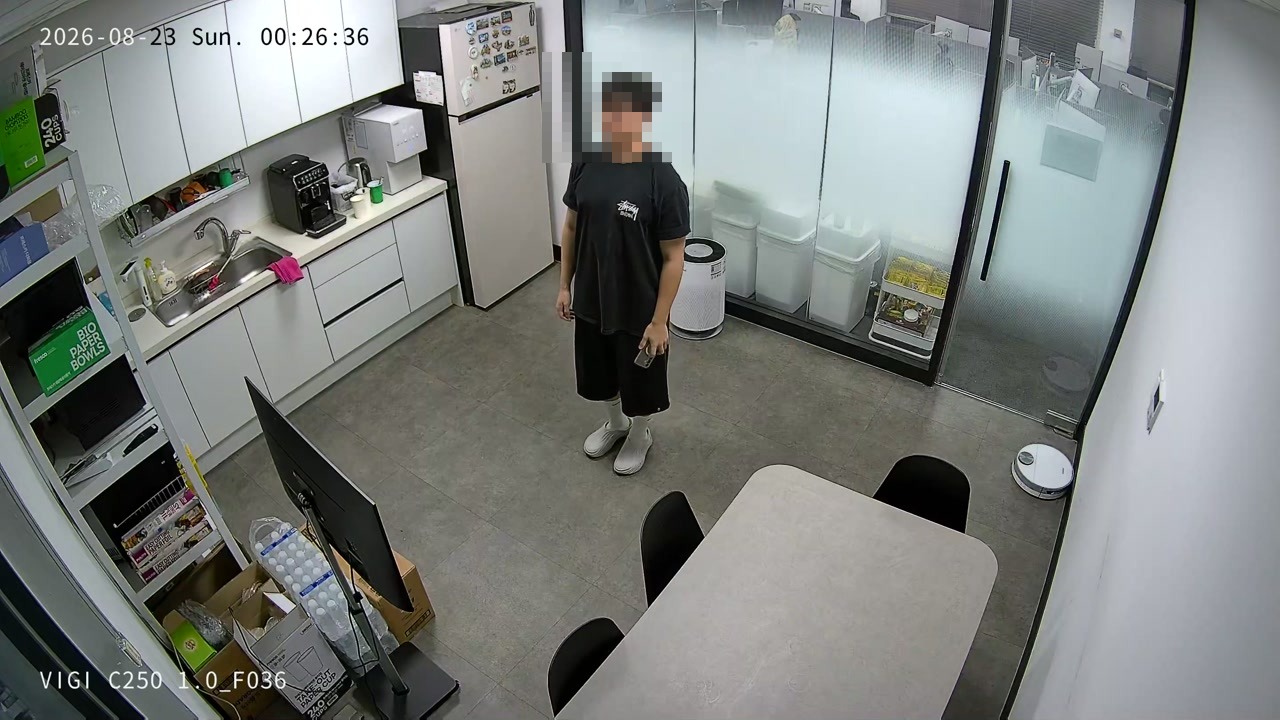}{G4 Identity $\cdot$ Q4}{It is muggy in my room, my skin feels sticky.}{Dehumidifier in Son's room} \\[7pt]
      \includegraphics[width=\linewidth]{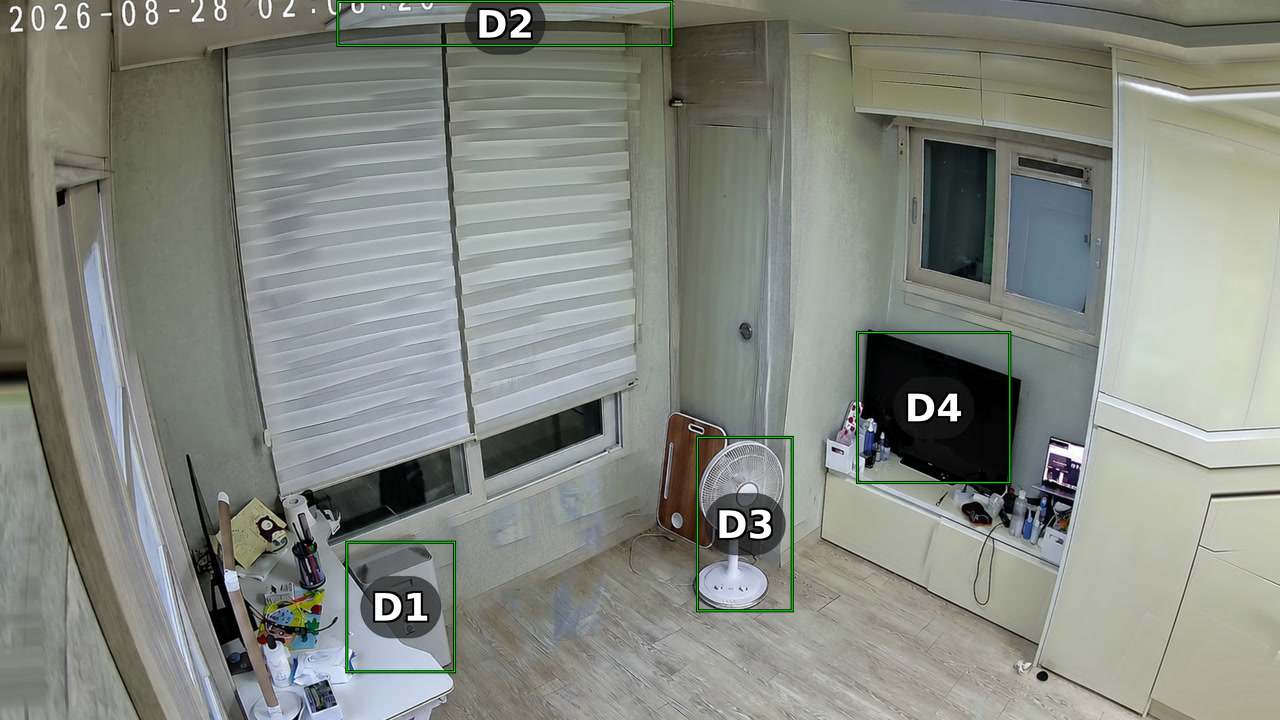} &
      \includegraphics[width=\linewidth]{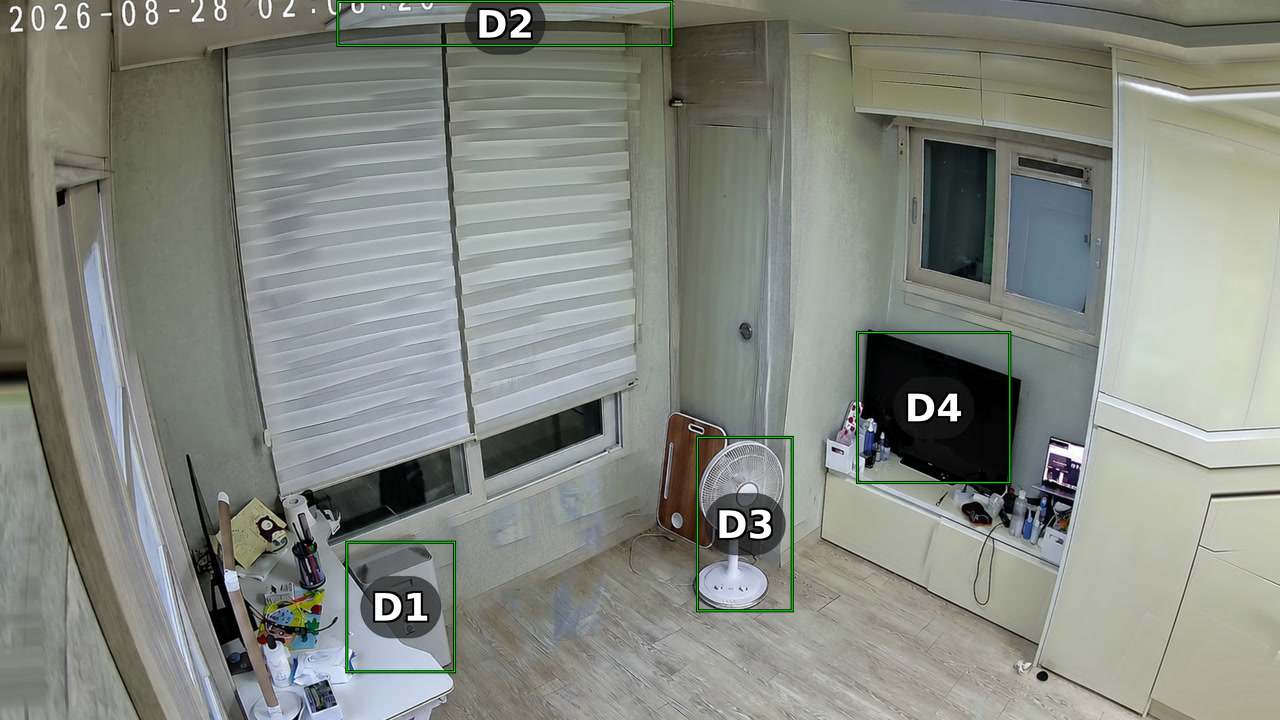} &
      \includegraphics[width=\linewidth]{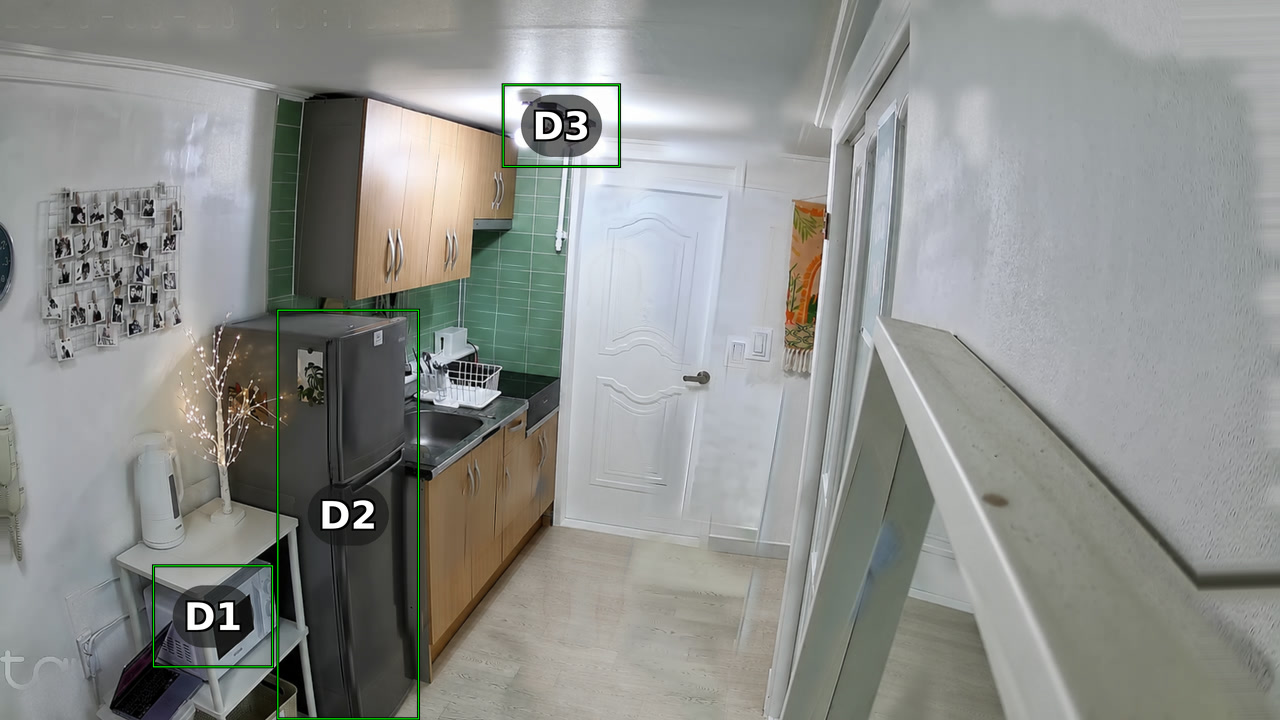} &
      \includegraphics[width=\linewidth]{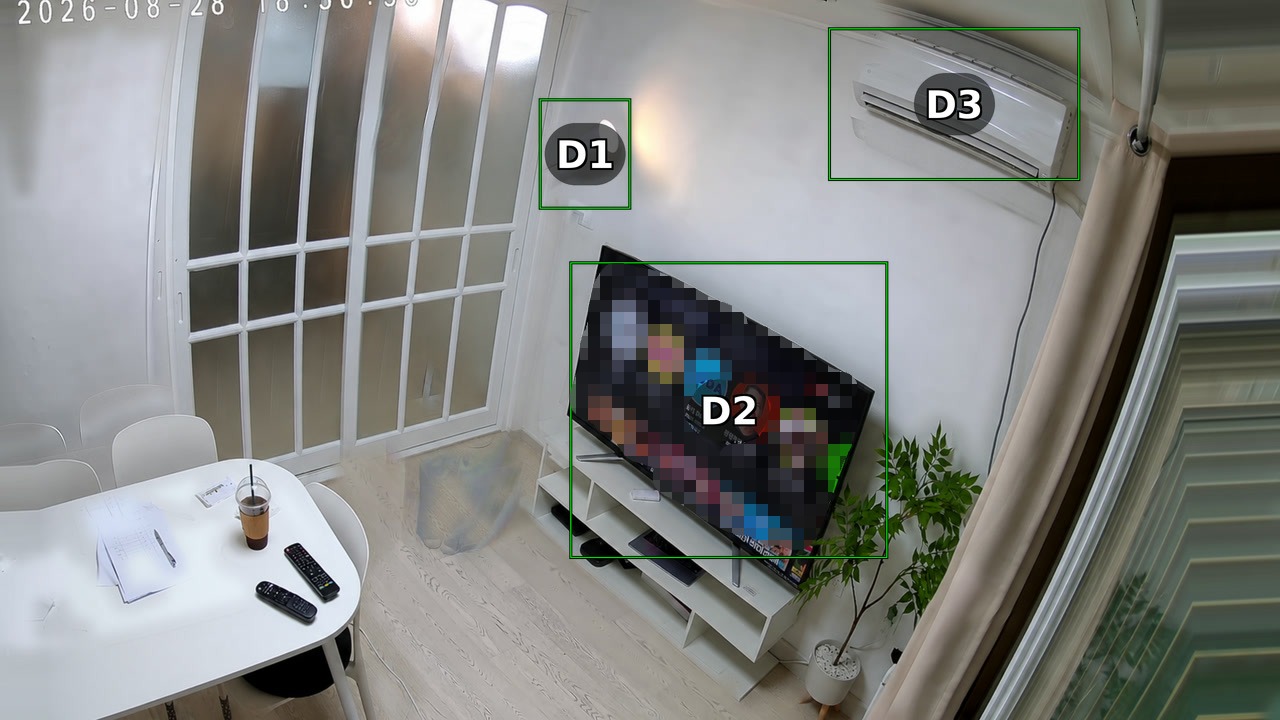} &
      \includegraphics[width=\linewidth]{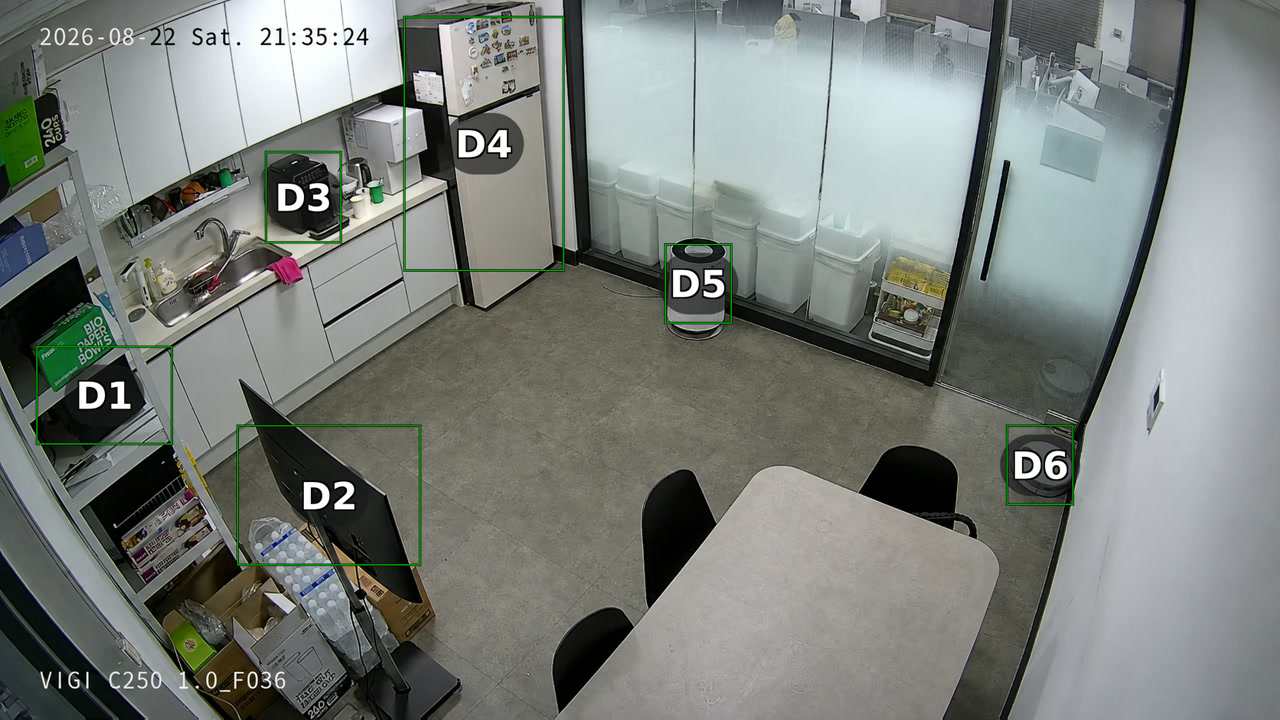} \\
  \end{tabular}
  \caption{\textbf{Real-world examples}, one per grounding type. Top: a frame of the recorded clip
  with the spoken request and the target device. Bottom: the reference image of the same room with
  device identifiers. The G4 request is infeasible: the target is the device that would have been
  needed, which the speaker's room does not contain.}
  \label{fig:real_examples}
\end{figure}

\newpara{Real-World Data}
\Cref{fig:real_examples} shows one recorded episode per grounding type. Each real-world episode is a
short video clip from a fixed wide-angle camera, synchronized with a four-channel recording from a
first-order Ambisonics microphone placed at the camera; participants act out the request in
furnished living rooms, kitchens and a laundry room that contain real appliances. The episodes
follow the same taxonomy as the synthetic set, so the same cues must be read from real footage:
the pointing arm in G1 occupies a small part of the wide-angle frame and is held only briefly, the beeping
microwave in G2 competes with room reverberation and the speaker's own voice, and ``in front of
me'' in G3 depends on the participant's body orientation rather than on the camera's.
The G4 example is an \emph{infeasible} episode: the speaker is recognized as the son, but no device
in his room can lower the humidity, so the correct behaviour is to explain that the request cannot
be carried out instead of acting on a device in the visible room. As for synthetic episodes, a
reference image with device identifiers is provided for every room (bottom row).

% \section{Benchmark Validity Check}
% \subsection{Grounding Difficulty Level}

% \subsection{oracle tool input}
% if we put oracle tool (that each gx requires for grouning ouput in the prompt itself, does the model's acc improve?

\subsection{Detail on Home Devices and Matter Cluster}

\newpara{Device list.}
\begin{table}[t]
\centering
\caption{\textbf{Device list.} Device categories and types included in our benchmark.}
\label{tab:device_list}
\setlength{\tabcolsep}{10pt}
\small
\begin{tabular}{ll}
\toprule
\textbf{Category} & \textbf{Devices} \\
\midrule
Lighting
& On/off light, dimmable light \\
Climate and air quality
& Air conditioner, fan, air purifier, dehumidifier \\
Media
& TV, audio player \\
Kitchen
& Refrigerator, freezer, microwave, coffee machine, dishwasher \\
Laundry and cleaning
& Laundry washer, laundry dryer, robot vacuum cleaner \\
Monitoring
& Electrical sensor \\
\bottomrule
\end{tabular}
\end{table}
Our benchmark covers a diverse set of smart-home devices across six categories, as summarized in \cref{tab:device_list}. The device set includes common lighting, climate, media, kitchen, cleaning, and monitoring appliances, enabling evaluation across a range of everyday smart-home scenarios.

\newpara{Matter cluster.}
We model device states and operations using Matter clusters. Each cluster exposes a set of attributes describing the current device state and, when applicable, commands that can modify the state. \cref{tab:matter_clusters} summarizes the clusters, attributes, and commands supported in our benchmark.
\clearpage
\begingroup
\footnotesize
\renewcommand{\arraystretch}{1.3}

\begin{xltabular}{\linewidth}{
    >{\hsize=0.6\hsize\raggedright\arraybackslash}X 
    >{\hsize=1.5\hsize\raggedright\arraybackslash}X 
    >{\hsize=0.9\hsize\raggedright\arraybackslash}X }

    \caption{\textbf{Matter clusters.} Attributes and commands supported by the Matter clusters used in our benchmark.}
    \label{tab:matter_clusters} \\
    \toprule
    \textbf{Cluster} & \textbf{Attributes} & \textbf{Commands} \\
    \midrule
    \endfirsthead

    \multicolumn{3}{c}{\footnotesize Table \thetable\ continues on the following page} \\
    \toprule
    \textbf{Cluster} & \textbf{Attributes} & \textbf{Commands} \\
    \midrule
    \endhead

    \midrule
    \multicolumn{3}{r}{\footnotesize (continued)} \\
    \endfoot

    \bottomrule
    \endlastfoot

Basic Information
& VendorName, VendorID, ProductName, ProductID
& -- \\

\addlinespace
Descriptor
& DeviceTypeList, ServerList, ClientList, PartsList, TagList
& -- \\

\addlinespace
Identify
& IdentifyTime, IdentifyType, FeatureMap, ClusterRevision
& Identify, TriggerEffect \\

\addlinespace
On/Off
& OnOff, GlobalSceneControl, OnTime, OffWaitTime, StartUpOnOff
& On, Off, Toggle \\

\addlinespace
Level Control
& CurrentLevel, RemainingTime, MinLevel, MaxLevel, CurrentFrequency, MinFrequency, MaxFrequency, Options, OnOffTransitionTime, OnLevel, OnTransitionTime, OffTransitionTime, DefaultMoveRate, StartUpCurrentLevel
& MoveToLevel, Move, Step, Stop, MoveToLevelWithOnOff, MoveWithOnOff, StepWithOnOff, StopWithOnOff, MoveToClosestFrequency \\

\addlinespace
Thermostat
& LocalTemperature, OccupiedCoolingSetpoint, OccupiedHeatingSetpoint, SystemMode, ControlSequenceOfOperation
& SetpointRaiseLower \\

\addlinespace
Fan Control
& FanMode, FanModeSequence, PercentSetting, PercentCurrent
& Step \\

\addlinespace
Temperature Measurement
& MeasuredValue, MinMeasuredValue, MaxMeasuredValue
& -- \\

\addlinespace
Temperature Control
& TemperatureSetpoint, MinTemperature, MaxTemperature
& SetTemperature \\

\addlinespace
Relative Humidity Measurement
& MeasuredValue, MinMeasuredValue, MaxMeasuredValue, Tolerance
& -- \\

\addlinespace
Media Playback
& CurrentState
& Play, Pause, Stop, StartOver, Previous, Next, Rewind, FastForward \\

\addlinespace
Channel
& ChannelList, Lineup, CurrentChannel
& ChangeChannel, ChangeChannelByNumber, SkipChannel \\

\addlinespace
Keypad Input
& FeatureMap, ClusterRevision
& SendKey \\

\addlinespace
Operational State
& PhaseList, CurrentPhase, CountdownTime, OperationalStateList, OperationalState, OperationalError
& Start, Stop, Pause, Resume, OperationalCommandResponse \\

\addlinespace
Microwave Mode
& SupportedModes, CurrentMode
& ChangeToMode \\

\addlinespace
Coffee Machine Mode
& SupportedModes, CurrentMode
& ChangeToMode \\

\addlinespace
Dishwasher Mode
& SupportedModes, CurrentMode
& ChangeToMode \\

\addlinespace
Laundry Washer Mode
& SupportedModes, CurrentMode
& ChangeToMode \\

\addlinespace
RTCC Mode
& SupportedModes, CurrentMode
& ChangeToMode \\

\addlinespace
Laundry Washer Controls
& SpinSpeeds, SpinSpeedCurrent, NumberOfRinses, SupportedRinses
& -- \\

\addlinespace
Laundry Dryer Controls
& SupportedDrynessLevels, SelectedDrynessLevel
& -- \\

\addlinespace
RVC Run Mode
& SupportedModes, CurrentMode, FeatureMap, ClusterRevision
& ChangeToMode \\

\addlinespace
RVC Clean Mode
& SupportedModes, CurrentMode, FeatureMap, ClusterRevision
& ChangeToMode \\

\addlinespace
RVC Operational State
& PhaseList, CurrentPhase, CountdownTime, OperationalStateList, OperationalState, OperationalError, ClusterRevision
& Pause, Resume, GoHome \\

\addlinespace
Alarm
& AlarmActive, AlarmCode
& Trigger, Silence \\

\addlinespace
Dishwasher Alarm
& Mask, Latch, State, Supported
& ModifyEnabledAlarms, Reset \\

\addlinespace
Electrical Power Measurement
& PowerMode, NumberOfMeasurementTypes, Accuracy, ActivePower, FeatureMap, ClusterRevision
& -- \\

\addlinespace
Electrical Energy Measurement
& Accuracy, FeatureMap, ClusterRevision
& -- \\

\addlinespace
Power Topology
& FeatureMap, ClusterRevision
& -- \\

\end{xltabular}
\endgroup

\section{Details on Tools}
\label{app:tool}
\subsection{Original Tools} 
\begin{table*}[t]
\centering
\caption{\textbf{Smart-home tools.}
\textsuperscript{$\dagger$} denotes tools additionally introduced for OmniSmartHome.}

\label{tab:smarthome_tools}
\vspace{-10pt}
\setlength{\tabcolsep}{1pt}

\resizebox{\textwidth}{!}{%
\small
\begin{tabular}{
>{\raggedright\arraybackslash}p{5cm}
>{\raggedright\arraybackslash}p{3cm}
>{\raggedright\arraybackslash}p{7.2cm}
}
\toprule
\textbf{Tool}
& \textbf{Args.}
& \textbf{Output} \\
\midrule

get\_rooms
& --
& Room IDs and names
\\

get\_room\_devices
& (room\_id)
& Devices in the specified room
\\

get\_room\_states
& (room\_id)
& Environmental states of the specified room
\\

get\_device\_structure
& (device\_id)
& Device endpoints, clusters, attributes, and commands
\\

get\_all\_attributes
& (device\_id)
& All attribute values of the specified device
\\

get\_attribute
& \makecell[tl]{
(device\_id,\\
endpoint\_id,\\
cluster\_id,\\
attribute\_id)
}
& Requested attribute value
\\

get\_cluster\_doc
& \makecell[tl]{
(query,\\
top\_k)
}
& Relevant Matter cluster documentation
\\

get\_environment\_control\_rules
& (state)
& Rules governing the specified environmental state
\\

execute\_command
& \makecell[tl]{
(device\_id,\\
endpoint\_id,\\
cluster\_id,\\
command\_id,\\
command\_args)
}
& Command execution result
\\

write\_attribute
& \makecell[tl]{
(device\_id,\\
endpoint\_id,\\
cluster\_id,\\
attribute\_id,\\
attribute\_value)
}
& Attribute update result
\\

get\_current\_time
& --
& Current simulation time
\\

schedule\_workflow
& \makecell[tl]{
(start\_time,\\
steps)
}
& Created workflow ID and scheduling result
\\

get\_workflow\_status
& (workflow\_id)
& Workflow status
\\

get\_workflow\_list
& --
& Registered workflows
\\

cancel\_workflow
& (workflow\_id)
& Cancellation result
\\

finish
& (answer)
& Final answer and episode termination
\\

get\_users\textsuperscript{$\dagger$}
& --
& Household user profiles and enrolled identity references
\\

declare\_target\_device\textsuperscript{$\dagger$}
& (device\_id)
& Recorded target-device declaration; ``none'' indicates no matching device
\\

\bottomrule
\end{tabular}
}

\end{table*}
\cref{tab:smarthome_tools} summarizes the smart-home tools available to the agent for querying environment states~\citep{seo2026simuhome}, inspecting Matter device structures, executing device operations, and managing scheduled workflows. We additionally introduce \texttt{get\_users} and \texttt{declare\_target\_device} for OmniSmartHome. The former provides household user information and enrolled identity references required for identity grounding, while the latter records the device selected by the agent as the grounded target.

\subsection{Perception Tools} 
\begin{table*}[t]
\centering
\caption{
\textbf{Perception tools used by the agent.}
Each tool processes the current episode observation, optionally together with enrolled household information, and returns structured perceptual evidence.
}
\label{tab:perception_tools}
\vspace{-2mm}
\setlength{\tabcolsep}{1pt}
\small

\resizebox{\textwidth}{!}{%
\begin{tabular}{l@{\hspace{8pt}}l@{\hspace{8pt}}l@{\hspace{8pt}}l@{\hspace{8pt}}l}
\toprule
\textbf{Tool}
& \textbf{Model}
& \textbf{Args.}
& \textbf{Perceptual input}
& \textbf{Output} \\
\midrule

asr
& Whisper large-v3
& -
& Episode audio
& Speech transcript.
\\

estimate\_doa
& OpenVocabularySELD
& -
& Episode audio
& \begin{tabular}[t]{@{}l@{}}
Sound-event labels and coarse\\ horizontal directions:\\
left, middle, or right.
\end{tabular}
\\

people\_detection
& Keypoint R-CNN
& -
& Episode image/video
& \begin{tabular}[t]{@{}l@{}}
Person labels and\\
bounding boxes.
\end{tabular}
\\

estimate\_pointing
& RTMW
& people\_label
& Episode image/video
& \begin{tabular}[t]{@{}l@{}}
Hand bounding box, visibility,\\
and an annotated image.
\end{tabular}
\\

estimate\_head\_pos
& DirectMHP
& people\_label
& Episode image/video
& \begin{tabular}[t]{@{}l@{}}
Facing directions:\\ front\_toward,
their\_right\_toward,\\ and their\_left\_toward.
\end{tabular}
\\

match\_face
& DINOv2
& people\_label
& \begin{tabular}[t]{@{}l@{}}
Episode image/video,\\
enrolled household\\
face images
\end{tabular}
& \begin{tabular}[t]{@{}l@{}}
Selected person label and\\
similarity scores indexed by user\_id.
\end{tabular}
\\

verify\_speaker
& WavLM
& candidate\_user\_id
& \begin{tabular}[t]{@{}l@{}}
Episode audio,\\
enrolled household\\
voices
\end{tabular}
& \begin{tabular}[t]{@{}l@{}}
Candidate user ID and speaker\\
similarity score.
\end{tabular}
\\

\bottomrule
\end{tabular}%
}

\end{table*}
\cref{tab:perception_tools} lists the perception tools of \ours{} used to gather information distributed across different modalities in the current episode. 

For audio, we use Whisper large-v3~\citep{radford2023robust} for speech transcription, OpenVocabularySELD~\citep{ovseld} for sound-event localization, and WavLM~\citep{chen2022wavlm} for speaker verification against enrolled household voices. 

For visual perception, we use Keypoint R-CNN~\citep{he2017mask} for person detection, RTMW~\citep{jiang2024rtmw} for pointing estimation, DirectMHP~\citep{zhou2023directmhp} for head-pose estimation, and DINOv2~\citep{oquab2023dinov2} for face matching against enrolled household face images. Together, these tools help the agent gather multimodal information.

% \section{Agentic Framework}

\section{Additional Results on OmniSmartHome}
\label{app:more_results}

\subsection{Baseline Models}
\label{app:api_usage}
We evaluate a diverse set of open-source and proprietary Omni-LLMs spanning different model scales and architectures.
\textbf{Small open-source models:} Video-LLaMA2~\citep{cheng2024videollama}, Qwen2.5-Omni~\citep{xu2025qwen25omnitechnicalreport}, Spatial-Omni~\citep{zhu2026spatial}, video-SALMONN~2+~\citep{tang2025video}, video-SALMONN-o1~\citep{sun2025videosalmonno}, OmniVinci~\citep{ye2026omnivinci}, and MiniCPM-o-4.5~\citep{cui2026minicpm}.
\textbf{Large open-source models:} Gemma~4~\citep{team2026gemma}, Nemotron-3-Nano-Omni~\citep{deshmukh2026nemotron}, Qwen3-Omni-Instruct/Thinking~\citep{xu2025qwen3}, and MiMo-V2.5~\citep{mimov25}.
\textbf{Proprietary models:} Gemini~2.5~Flash/Pro~\citep{comanici2025gemini}, Gemini~3.1~Pro~(Preview)~\citep{gemini3pro}, and Qwen3.8-Omni-Flash~\citep{qwen38omniflash}.
Among these, only Spatial-Omni and Qwen3.8-Omni-Flash natively accept spatial (FOA) audio; for all other models, we downmix the FOA recordings to mono before inference.
Proprietary models are accessed through their official APIs (Vertex AI for Gemini, Alibaba Cloud for Qwen3.8-Omni-Flash). Among the open-source models, MiMo-V2.5 is accessed through OpenRouter due to their scale, and the rest are run locally with their released checkpoints.

\subsection{Additional Results by Query Type}

\Cref{tab:main_results_qtype} reports accuracy by query type. For large open-source Omni-LLMs and proprietary models, grounding accuracy remains relatively stable across query types, with the exception of intent queries (Q4). We attribute this drop to the additional reasoning Q4 demands, namely interpreting the user's intent and selecting an appropriate device, rather than to multimodal perception itself; for the remaining types, resolving the multimodal reference is comparably difficult regardless of the query type. Goal accuracy, in contrast, varies substantially across query types. Consistent with the findings of \cite{seo2026simuhome}, models achieve particularly low goal accuracy on queries that require scheduling (Q3) and implicit intent inference (Q4), indicating that these capabilities remain challenging for current LLM-based agents. PROME also yields consistent gains in both grounding and goal accuracy across all query types.
% =========================================================
% Required packages / macros
% =========================================================

% Required:
% \usepackage{booktabs}
% \usepackage{array}
% \usepackage{graphicx}
% \usepackage[table]{xcolor}   % \rowcolor 쓰려면 table 옵션 필요
% \usepackage{amssymb}

% Metric column
\definecolor{gainblue}{RGB}{90,120,160}
\definecolor{lossred}{RGB}{160,115,120}
\definecolor{neutralgray}{RGB}{110,110,110}
\definecolor{modelhighlight}{RGB}{235,240,250}  % 이미 정의돼 있으면 이 줄 제거
\definecolor{allshade}{RGB}{245,245,245}

% =========================================================
% Score macros
%
%   \scoreup{75.2}{+4.3}     -> 75.2_{+4.3} (blue)   (subscript style, 미사용)
%   \scoredown{68.1}{-1.2}   -> 68.1_{-1.2} (red)
%   \scoreflat{70.0}{0.0}    -> 70.0_{0.0}  (gray)
%   \ms{74.1}{1.4}           -> 74.1_{±1.4} (mean ± std)
%
%   \gain{+4.3}              -> (+4.3) small blue    (Gain 행에서 사용)
%   \loss{-1.2}              -> (-1.2) small red
%   \flat{0.0}               -> (0.0)  small gray
% =========================================================

% =========================================================
% Main table (query types Q1--Q4)
% =========================================================

\begin{table*}[t]
\centering

\caption{
\textbf{Results on OmniSmartHome across query types (Q1--Q4).}
For each query type, we report goal accuracy (Goal) and
grounding accuracy (Gr.), averaged over grounding types.
For our agentic system, the (Gain) row reports absolute changes
in percentage points relative to the corresponding vanilla backbone.
}

\label{tab:main_results_qtype}
\vspace{-3mm}
\renewcommand{\arraystretch}{1.10}

\resizebox{\textwidth}{!}{%
\footnotesize
\begin{tabular}{
l
C@{\hspace{2pt}}C@{\hspace{8pt}}
C@{\hspace{2pt}}C@{\hspace{8pt}}
C@{\hspace{2pt}}C@{\hspace{8pt}}
C@{\hspace{2pt}}C
}

\toprule

&
\multicolumn{2}{c}{\textbf{Q1} State Inquiry}
&
\multicolumn{2}{c}{\textbf{Q2} Control }
&
\multicolumn{2}{c}{\textbf{Q3} Schedule}
&
\multicolumn{2}{c}{\textbf{Q4} Intent}
\\

\cmidrule(lr){2-3}
\cmidrule(lr){4-5}
\cmidrule(lr){6-7}
\cmidrule(lr){8-9}

\textbf{Model}
&
Gr. & Goal
&
Gr. & Goal
&
Gr. & Goal
&
Gr. & Goal
\\

% =========================================================
% Open-source AVLLMs (<7B)
% =========================================================

\midrule
\multicolumn{9}{c}{\textit{Open-source Omni-LLMs (<10B)}} \\
\midrule

Video-LLaMA2\psize{7B}
& 6.0 & 9.2
& 5.2 & 0.6
& 5.8 & 6.7
& 6.8 & 0.5
\\
% \rowcolor{modelhighlight}
Qwen2.5-Omni\psize{7B}
& 31.2 & 32.9
& 36.2 & 8.8
& 30.4 & 13.3
& 35.9 & 12.0
\\
Spatial-Omni\psize{7B}
& 18.3 & 20.4   % q1 (N=480)
& 23.1 & 2.5    % q3 (N=480)
& 19.8 & 15.0   % q4 (N=480)
& 16.7 & 2.6    % q2 (N=192)
\\
video-SALMONN2+\psize{7B}
& 32.3 & 24.6
& 24.8 & 6.0
& 23.1 & 14.0
& 29.2 & 5.2
\\

video-SALMONN-o1\psize{7B}
& 24.4 & 16.0
& 27.1 & 5.6
& 23.8 & 12.9
& 20.3 & 1.0
\\

OmniVinci\psize{9B}
& 16.9 & 17.5
& 17.7 & 1.2
& 16.7 & 10.2
& 22.9 & 2.1
\\

MiniCPM-o-4.5\psize{9B}
& 61.3 & 41.7
& 63.7 & 32.1
& 60.6 & 20.6
& 51.0 & 27.1
\\

% =========================================================
% Open-source AVLLMs
% =========================================================

\midrule
\multicolumn{9}{c}{\textit{Large Open-source Omni-LLMs}} \\
\midrule

% \rowcolor{modelhighlight}
Gemma4 \psize{12B}
& 50.4 & 22.3
& 49.4 & 24.6
& 56.0 & 27.5
& 40.1 & 28.1
\\

% \rowcolor{modelhighlight}
Nemotron3-Omni\psize{30B-A3B}
& 63.7 & 62.1
& 67.7 & 41.9
& 60.0 & 36.2
& 54.2 & 28.1
\\
Qwen3-Omni-Instruct\psize{30B-A3B}
& 72.1 & 66.2
& 71.9 & 42.7
& 59.6 & 26.9
& 57.3 & 32.3
\\

% \rowcolor{modelhighlight}
Qwen3-Omni-Think\psize{30B-A3B}
& 68.8 & 65.4
& 72.7 & 49.6
& 59.0 & 34.6
& 58.9 & 37.5
\\

MiMo-V2.5\psize{310B-A15B}
& 74.6 & 73.3
& 81.7 & 64.6
& 82.7 & 46.9
& 69.3 & 50.0
\\

% =========================================================
% Closed-source AVLLMs
% =========================================================

\midrule
\multicolumn{9}{c}{\textit{Proprietary Omni-LLMs}} \\
\midrule

% \rowcolor{modelhighlight}
Gemini-2.5-Flash
& 72.5 & 69.0
& 77.9 & 59.2
& 77.9 & 46.5
& 55.2 & 41.7
\\
% \rowcolor{modelhighlight}
Gemini-2.5-Pro
& 80.4 & 82.1
& 83.5 & 59.8
& 84.6 & 50.0
& 67.2 & 56.8
\\

Gemini-3.1-Pro (Preview)
& 83.5 & 84.2
& 89.4 & 76.2
& 91.9 & 64.8
& 65.6 & 60.9
\\
Qwen3.8-Omni-Flash
& 81.2 & 81.0
& 85.4 & 63.1
& 86.0 & 50.6
& 63.0 & 49.5

\\
\midrule
\multicolumn{9}{c}{\textit{Agent Baseline}} \\
\midrule
% \rowcolor{modelhighlight}
& 39.4 & 31.9
& 41.2 & 15.8
& 34.6 & 21.0
& 37.0 & 18.8
\\[-1pt]
% \rowcolor{modelhighlight}
\multirow{-2}{*}{Qwen2.5-Omni + \textbf{\ours}}
% \quad Gain 
& \gain{+8.1} & \loss{-1.0}
& \gain{+5.0} & \gain{+7.1}
& \gain{+4.2} & \gain{+7.7}
& \gain{+1.0} & \gain{+6.8}
\\

% \rowcolor{modelhighlight}
& 72.5 & 66.9
& 80.0 & 66.0
& 75.4 & 48.8
& 59.4 & 49.5
\\[-1pt]
% \rowcolor{modelhighlight}
% \quad Gain 
\multirow{-2}{*}{Gemma4  + \textbf{\ours}}
& \gain{+22.1} & \gain{+44.6}
& \gain{+30.6} & \gain{+41.5}
& \gain{+19.4} & \gain{+21.2}
& \gain{+19.3} & \gain{+21.4}
\\

% \rowcolor{modelhighlight}
& 66.7 & 63.7
& 70.8 & 52.9
& 61.7 & 37.9
& 56.2 & 31.2
\\[-1pt]
% \rowcolor{modelhighlight}
% \quad Gain 
\multirow{-2}{*}{Nemotron3-Omni + \textbf{\ours}}
& \gain{+2.9} & \gain{+1.7}
& \gain{+3.1} & \gain{+11.0}
& \gain{+1.7} & \gain{+1.7}
& \gain{+2.1} & \gain{+3.1}
\\

% \rowcolor{modelhighlight}
& 72.9 & 69.6
& 76.9 & 58.5
& 65.6 & 40.8
& 57.8 & 35.4
\\[-1pt]
% \rowcolor{modelhighlight}
% \quad Gain 
\multirow{-2}{*}{Qwen3-Omni-Think + \textbf{\ours}}
& \gain{+4.2} & \gain{+4.2}
& \gain{+4.2} & \gain{+9.0}
& \gain{+6.7} & \gain{+6.2}
& \loss{-1.0} & \loss{-2.1}
\\

% \rowcolor{modelhighlight}
& 80.6 & 77.1
& 81.5 & 66.5
& 80.0 & 49.4
& 59.9 & 46.9
\\[-1pt]
% \rowcolor{modelhighlight}
% \quad Gain 
\multirow{-2}{*}{Gemini-2.5-Flash + \textbf{\ours}}
& \gain{+8.1} & \gain{+8.1}
& \gain{+3.5} & \gain{+7.3}
& \gain{+2.1} & \gain{+2.9}
& \gain{+4.7} & \gain{+5.2}
\\

% \rowcolor{modelhighlight}
& 85.0 & 85.0
& 87.1 & 70.8
& 86.5 & 52.9
& 76.6 & 66.1
\\[-1pt]
% \rowcolor{modelhighlight}
% \quad Gain 
\multirow{-2}{*}{Gemini-2.5-Pro + \textbf{\ours}}
& \gain{+4.6} & \gain{+2.9}
& \gain{+3.5} & \gain{+11.0}
& \gain{+1.9} & \gain{+2.9}
& \gain{+9.4} & \gain{+9.4}
\\

\bottomrule
\end{tabular}
}
\end{table*}

% =========================================================
% Conditional goal accuracy table
% Requires: C column type, \psize, modelhighlight (main table과 공유)
% =========================================================

\begin{table}[t]
\centering

\caption{
\textbf{Goal accuracy conditioned on the grounding outcome.}
$P(L\mid G)$ is the goal accuracy when the declared target device is
correct, and $P(L\mid \neg G)$ when it is incorrect.
}

\label{tab:cond_goal}
\vspace{-3mm}
\renewcommand{\arraystretch}{1.10}

\footnotesize
\begin{tabular}{
l
@{\hspace{10pt}}
>{\centering\arraybackslash}p{1.5cm}@{\hspace{4pt}}
>{\centering\arraybackslash}p{1.5cm}
}

\toprule

\textbf{Model}
& $P(L\mid G)$ & $P(L\mid \neg G)$
\\

% =========================================================
% Open-source AVLLMs (<7B)
% =========================================================

\midrule
\multicolumn{3}{c}{\textit{Open-source Omni-LLMs (<10B)}} \\
\midrule
Video-LLaMA2\psize{7B}
& 43.5 & 2.7
\\

Qwen2.5-Omni\psize{7B}
& 33.8 & 9.6
\\

video-SALMONN2+\psize{7B}
& 35.5 & 5.8
\\

video-SALMONN-o1\psize{7B}
& 24.8 & 3.7
\\

OmniVinci\psize{9B}
& 32.4 & 3.7
\\

MiniCPM-o-4.5\psize{9B}
& 49.1 & 3.0
\\

% =========================================================
% Open-source AVLLMs
% =========================================================

\midrule
\multicolumn{3}{c}{\textit{Open-source Omni-LLMs}} \\
\midrule

Gemma4\psize{12B}
& 49.0 & 0.9
\\

Nemotron3-Omni\psize{30B-A3B}
& 68.5 & 4.3
\\

Qwen3-Omni-Instruct\psize{30B-A3B}
& 64.9 & 1.7
\\

Qwen3-Omni-Think\psize{30B-A3B}
& 72.8 & 1.3
\\

% =========================================================
% Proprietary AVLLMs
% =========================================================
MiMo-V2.5
& 74.3 & 4.0
\\
\midrule
\multicolumn{3}{c}{\textit{Proprietary Omni-LLMs}} \\
\midrule

Gemini-2.5-Flash
& 74.6 & 4.9
\\

Gemini-2.5-Pro
& 75.9 & 8.4
\\

Gemini-3.1-Pro (Preview)
& \textbf{83.8} & 11.9
\\
Qwen3.8-Omni-Flash
& 75.5 & 7.7
\\

\bottomrule
\end{tabular}
\vspace{-4mm}
\end{table}
\paragraph{Grounding is a prerequisite for goal completion.}
Table~\ref{tab:cond_goal} decomposes goal accuracy by whether the agent
declared the correct target device. Two patterns emerge. First, when
grounding fails, goal accuracy collapses: $P(L\mid\neg G)$ stays below
12\% for every model and below 5\% for most, so a wrong referent is
almost never recovered downstream. Second, correct grounding is
necessary but not sufficient, and how much of it converts into goal
success depends strongly on model capacity. Small open-source models
complete the goal in only 32--49\% of the episodes they ground correctly,
as they often fail at the subsequent multi-step device operation and
scheduling even with the right target in hand. Larger open-source models
reach 65--73\%, and proprietary models 75--84\%. Hence, for weaker
models both grounding and post-grounding execution are bottlenecks,
whereas for stronger models the remaining errors are increasingly
dominated by grounding itself.

\subsection{Additional Results by Difficulty Level}
\label{sec:grounding_difficulty}

\begin{figure}[t]
\centering
\includegraphics[width=\linewidth]{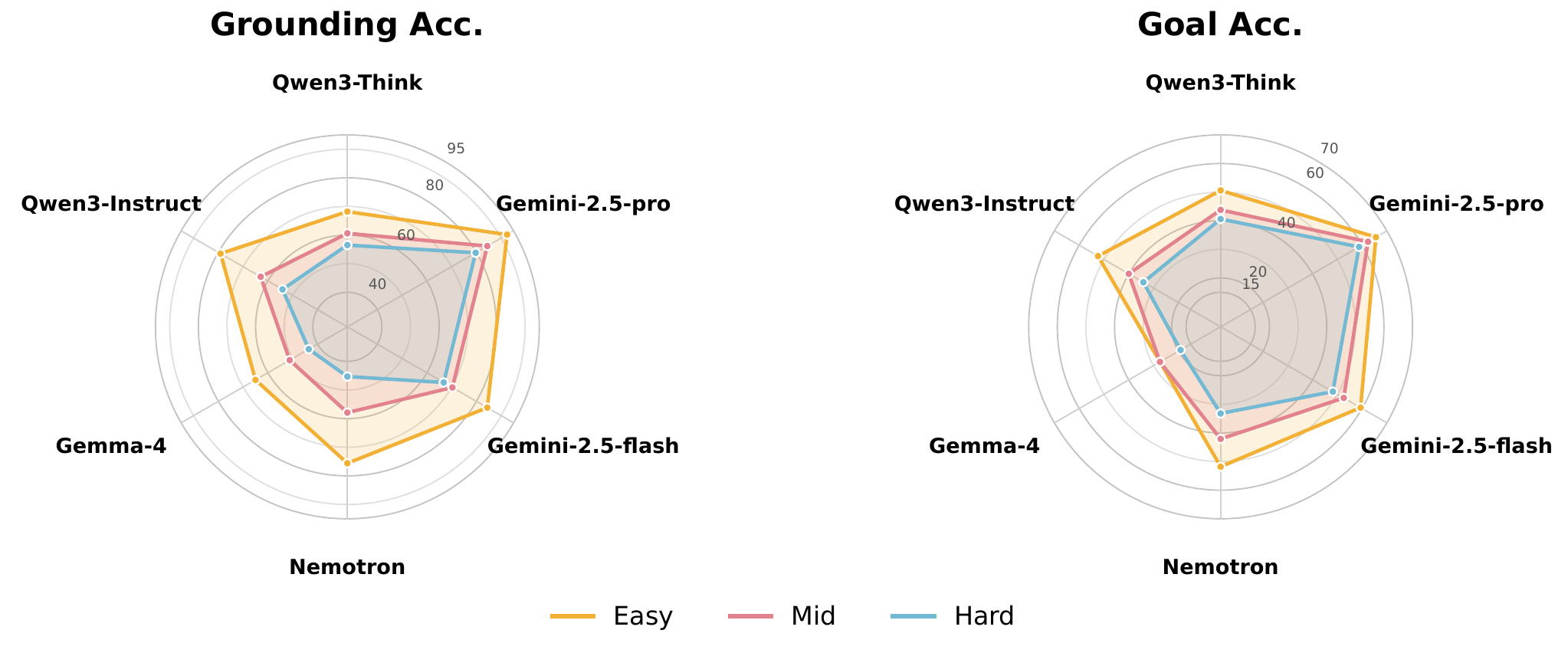}
\caption{
\textbf{Model performance across grounding difficulty levels for feasible queries.} The \textcolor{easycolor}{\textit{easy}}, \textcolor{midcolor}{\textit{mid}}, and \textcolor{hardcolor}{\textit{hard}} levels are defined according to the type and number of distractors. Both grounding accuracy and goal accuracy consistently decrease as grounding difficulty increases.
}
\label{fig:difficulty}
\end{figure}
Our synthetic dataset, \textsc{Feasible}, is categorized into three levels of \emph{grounding difficulty}: \textit{easy}, \textit{mid}, and \textit{hard}. These difficulty levels are defined based on the composition of distractors in each instance, including their types and quantities. As the distractors become more numerous or more difficult to distinguish from the target evidence, the grounding task becomes correspondingly more challenging.

Figure~\ref{fig:difficulty} presents model performance across the three difficulty levels. We observe a clear and consistent ordering for both \emph{grounding accuracy} and \emph{goal accuracy}: \textcolor{easycolor}{\textit{easy}} $>$ \textcolor{midcolor}{\textit{mid}} $>$ \textcolor{hardcolor}{\textit{hard}}. This pattern holds broadly across models, showing that increasing grounding difficulty not only makes it harder to identify the correct evidence, but also leads to lower downstream task success. The results further validate that our difficulty stratification captures meaningful differences in the level of grounding challenge induced by distractors.

\subsection{Modality Ablation}
\label{sec:modality_ablation}

\begin{wrapfigure}{r}{0.36\linewidth}
\vspace{-12pt}
\centering
\includegraphics[width=\linewidth]{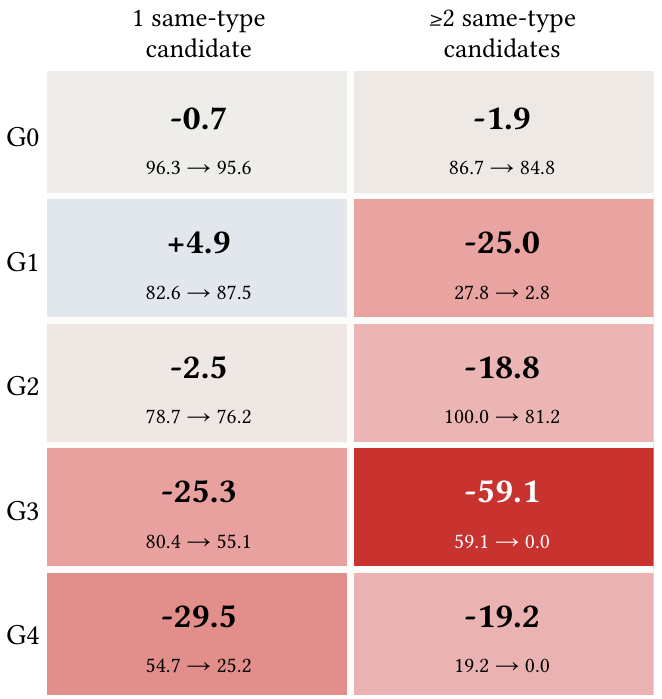}
\caption{\textbf{Modality ablation.} Grounding accuracy of Gemini-2.5-Flash with the full multimodal input and text-only input.}
\label{fig:modality_ablation}
\vspace{-25pt}
\end{wrapfigure}

To assess whether multimodal observations genuinely contribute to grounding in OmniSmartHome, we evaluate a text-only variant of the vanilla agent, removing all visual and acoustic observations and providing the user request only as a transcript. We distinguish the \textit{type} of a device, i.e., its category (e.g., dimmable light), from its \textit{instance}, i.e., a particular device of that type (e.g., dimmable\_light\_1 or dimmable\_light\_2). Fig.~\ref{fig:modality_ablation} reports the grounding accuracy of Gemini-2.5-Flash, separating cases where the target is the only instance of its type in the target room from those where multiple devices share the same type.

The results reveal a clear division of labor between language and multimodal evidence: language is often sufficient to identify the \textit{type} of the referent, whereas resolving its \textit{instance} requires audio-visual context. For G0, the spoken request itself contains all information necessary for grounding, and replacing it with its transcript therefore causes almost no degradation. For G1--G4, however, multimodal evidence becomes increasingly important, especially when multiple devices share the target type. When the target is the only instance of its type, linguistic cues can still narrow the candidate set. In contrast, when two or more same-type candidates are present, language alone cannot distinguish among them, forcing grounding to rely on audio-visual evidence and resulting in sharp drops (G1: 27.8$\rightarrow$2.8, G2: 100.0$\rightarrow$81.2, G3: 59.1$\rightarrow$0.0). G4 degrades substantially in both settings because grounding requires the agent to first identify the speaker, then infer the corresponding room, and finally locate the device; without multimodal observations, this reasoning chain already breaks at its first step, irrespective of the number of candidate devices.

\subsection{Error Analysis}
\label{app:error}

\cref{fig:error} reports one representative failure mechanism per grounding type for Qwen3-Omni-Think, Gemma4-12B, Gemini-2.5-Flash, and Gemini-2.5-Pro, in the base agent setting. All four panels are computed from recorded episode traces (tool calls, observations, and the declared device) together with the benchmark ground truth; no additional model calls are made. An episode counts as a \emph{grounding failure} if the declared device is not the referent or no device is declared, and as a \emph{wrong pick} if a device other than the referent is declared. Episodes that ended in a protocol or infrastructure failure (malformed output, timeout, context overflow) are excluded throughout.

\paragraph{G1.}
Among feasible synthetic episodes, we examine wrong picks. Candidates are the in-frame device bounding boxes returned by \texttt{get\_room\_devices}. The bar shows the share of wrong picks in which the declared device is the candidate closest, in image space, to the user's hand or body (which coincide in most rooms).

\paragraph{G2.}
Among feasible synthetic and real-world episodes, we again examine wrong picks. The bar shows the share of wrong picks in which the agent declared the device \emph{without reading the state of any device} beforehand, i.e., no state-reading tool (\texttt{get\_attribute}, \texttt{get\_device\_structure}, \texttt{get\_all\_attributes}) was called on any device before the declaration. Such declarations rely on the room layout and the sound alone and never verify that the chosen device is in the described state. 

\paragraph{G3.}
We consider feasible synthetic episodes whose referent is described as ``on my left'' or ``on my right.'' The user's orientation is measured as the angle between their body-forward vector and the vector from the user to the camera, taken from the scene layout. Episodes with an angle below $80^\circ$ are \emph{facing camera} ($n{=}12$), where the user's left appears on the image's right and left/right must be mirrored; those above $100^\circ$ are \emph{facing away} ($n{=}18$), where image and user sides coincide. Episodes in between are excluded. The bars show grounding failure rates within each group.

% \begin{figure}[t]
%     \centering
%     \includegraphics[width=0.4\linewidth]{zolaman_lr.jpeg}
%     \caption{\textbf{Illustration of the facing-camera setting}. The speaker faces the camera and left--right directions are reversed relative to the image.}
%     \label{fig:facing-camera}
% \end{figure}

\paragraph{G4.}
Among feasible synthetic and real-world episodes, we examine grounding failures. The bar shows the share of failures in which the agent never called a speaker- or face-identification tool (\texttt{get\_users}, \texttt{match\_face}, \texttt{verify\_speaker}, \texttt{people\_detection}) before declaring, i.e., it committed to a device without attempting to identify the speaker.

% \newpara{Spatial vs.\ mono audio.}
% To isolate the contribution of spatial hearing, we evaluate Qwen3.8-Omni-Flash on the full benchmark under two audio inputs: 4-channel FOA and mono audio alone.
% Removing the spatial channels lowers overall accuracy,
% and the drop concentrates in the space-grounded group G3 (Fig.~\ref{fig:ablation_foa}).
% Performance on the other groups remains relatively stable, as cases such as C0 and C1 can often be resolved using non-spatial visual or audio cues, without relying on spatial audio.

% \begin{figure}[t]
%   \centering
%   \includegraphics[width=0.7\linewidth]{asset/ablation_foa.pdf}
%   \caption{\textbf{Effect of spatial audio.} Grounding and goal accuracy with 4-channel FOA and mono audio on Qwen3.8-Omni-Flash. }
%   \label{fig:ablation_foa}
% \end{figure}

% \subsection{Error Analysis}

\section{\ours{}}
\label{app:ours}

\subsection{Details of \ours{}}
\label{app:details_of_method}
\ours{} adds a small procedural memory $\mathcal{M}=\{\pi_\sigma\}_{\sigma\in\Sigma}$, a bank of grounding procedures indexed by the grounding schema registry $\Sigma$.

\newpara{Grounding schema.}
The grounding schema registry $\Sigma$ is mined from the synthetic training episodes using only the audio-visual observations of each episode; target labels, grounding-type labels, and the home state are never shown.
Mining proceeds in two passes with an Omni-LLM (Qwen3-Omni-Think~\citep{xu2025qwen3}).
In the first pass, the model describes the reusable cues of each episode in one sentence, and is forbidden to name any device, room, person, tool, or tool sequence.
In the second pass, episodes are visited in random order while the table is grown incrementally: for each episode, the model either assigns it to an existing grounding schema or adds a new one, judging sameness by the kind of evidence relation required to identify the target rather than by wording or by the downstream request.
This yields $|\Sigma|=8$ grounding schemas (\cref{tab:schema_registry}). Since they are induced automatically, they do not correspond one-to-one to our predefined grounding types.

\begin{table}[t]
\centering
\footnotesize
\setlength{\tabcolsep}{5pt}
\renewcommand{\arraystretch}{1.15}
\caption{\textbf{The grounding schema registry.} Eight grounding schemas were induced from the synthetic training episodes by the table-construction prompt (Appendix~\ref{app:ground_prompt}) and frozen; at inference, the selection prompt picks exactly one per episode. Each $\sigma$ names the evidence relation that identifies the target device.}
\begin{tabular}{@{}clp{8.6cm}@{}}
\toprule
\textbf{ID} & \textbf{Evidence relation} & \textbf{Grounding Schema} \\
\midrule
$\sigma_1$ & pointing gesture & Identifying a specific device in a room image by a pointing gesture when the user refers to it as `this' or `that' to request information about it. \\
$\sigma_2$ & position rel.\ to user & Identifying a specific device in a room image by its position relative to the user's body (e.g., above head, to the left) when the user refers to it with such spatial descriptors. \\
$\sigma_3$ & audible sound & Identifying a specific device in a room image by its audible sound when the user refers to it as the source of that sound to request information about it. \\
$\sigma_4$ & device + room named & Identifying a specific device in a room image by the user explicitly naming the device and room to request an action on it. \\
$\sigma_5$ & household member & Identifying a specific household member from a room image to resolve pronoun references (e.g., `his' or `her') for room-based requests. \\
$\sigma_6$ & room named & Identifying a specific device in a room image by the user explicitly naming the room to request an action on a device within that room. \\
$\sigma_7$ & sound + position & Identifying a specific device in a room image by both the audible sound it is producing and its position relative to the user's body to request information about it. \\
$\sigma_8$ & room state & Identifying a specific device in a room image by the user describing a room state (e.g., air quality, temperature) to determine which device can modify that state. \\
\bottomrule
\end{tabular}
\label{tab:schema_registry}
\end{table}

At test time, $\Sigma$ is frozen. Given the request and the initial observation, the agent retrieves $\sigma^\ast\in\Sigma$ once per episode by choosing among the eight entries.

\newpara{State.}
Progress within an episode is summarized by a discrete agent state $s_t=(p_t,c_t,r_t)\in\mathcal{S}$ with $|\mathcal{S}|=5\times3\times4=60$:
\begin{itemize}
\item $p_t\in\{\texttt{prelist},\texttt{unique},\texttt{multi},\texttt{absent},\texttt{unknown}\}$:
      candidate plausibility---whether the room has been listed yet and, if so, whether one, several, or no listed devices match the requested type;
\item $c_t\in\{0,1,2{+}\}$: how many candidates have had their state read through tools such as \texttt{get\_attribute};
\item $r_t\in\{\texttt{none},\texttt{spatial},\texttt{identity},\texttt{both}\}$:
      which class of perception tool has been called.
\end{itemize}
The state is updated deterministically after each ReAct step: \textsc{List} tools such as \texttt{get\_room\_devices} or \texttt{get\_users} set $p_t$, \textsc{Read} tools increment $c_t$, and perception tools set $r_t$ (action classes are defined below).

\newpara{Action and guidance.}
The action set is
$\mathcal{A}=\{\textsc{List},\textsc{Read},\textsc{Perc}{:}\langle\text{tool}\rangle,\textsc{Declare}\}$,
and every tool the agent may call before declaring maps to exactly one entry:
\begin{itemize}
\item \textsc{List}: room, home, and household-roster queries
      (e.g., \texttt{get\_room\_devices}, \texttt{get\_users});
\item \textsc{Read}: attribute and structure queries on a device
      (e.g., \texttt{get\_attribute});
\item \textsc{Perc}{:}$\langle\text{tool}\rangle$: one of the seven perception tools in \cref{tab:perception_tools}
      (e.g., \texttt{people\_detection}, \texttt{estimate\_doa});
\item \textsc{Declare}: declaring the target device.
\end{itemize}
Tools used after device declaration, such as command execution and scheduling, are outside $\mathcal{A}$, and no guidance is injected once the target has been declared.
The selected action $a_t$ is verbalized as a \textsc{[Procedural Guidance]} inserted before the agent's next reasoning step. The guidance first states the current state (``Now: the devices are listed but the candidates are not narrowed yet; no candidate state read yet; no perception tool used yet.'') and then suggests the evidence to gather next (``read the candidates' state'' or ``call \texttt{estimate\_doa} to acquire the missing evidence'').
The guidance names only the type of evidence to gather, never a candidate device.

\newpara{Initial policy.}
The initial logits $\theta^{0}$ combine a cost prior shared by all grounding schemas with a utility bonus $\Delta$ for perception tools:
\begin{equation}
\theta^{0}_{\sigma,s,a}=-\gamma\,\mathrm{cost}(a,s)+\Delta_{\sigma,s,a},
\qquad \gamma=3,
\label{eq:prior_logit}
\end{equation}
where $\Delta_{\sigma,s,a}=0$ for every non-perception action.
\begin{center}\small
\begin{tabular}{@{}l p{0.72\linewidth}@{}}
\toprule
Action & Cost \\
\midrule
\textsc{List} & $0$ before the room is listed; $10$ after \\
\textsc{Read} & $0$ while unresolved; $0.5$ after one verifying read of a unique candidate \\
\textsc{Perc}{:}$\langle\text{tool}\rangle$ & $1$ \\
\textsc{Declare} & $0$ once a unique candidate is verified; \newline
                   $0.5$ if unverified, or if $2{+}$ candidates were read without a match; \newline
                   $1$ otherwise \\
\bottomrule
\end{tabular}
\end{center}
The cost prior encodes a general-purpose procedure: list the room, check the current states of candidate devices, and declare, calling a perception tool only when the ambiguity cannot be resolved from the Omni-LLM's own perception of the initial observation.
Since each perception tool queries an external expert model and thus adds latency, the cost prior generally discourages such calls.

The utility bonus $\Delta$ estimates, per $(\sigma,s)$ and perception tool $a$, whether invoking the tool there actually helps.
We roll out the memory-free agent on the training episodes twice: a \emph{vanilla} rollout $\tau_{\text{ref}}^{-}$ without perception tools and a \emph{perception-only} rollout $\tau_{\text{ref}}^{+}$ with them, i.e., the reference rollouts of \cref{eq:reinforce}.
Whenever the perception-only rollout of episode $e$ invokes tool $a$ at $(\sigma_e,s)$, we record the same-episode gain
$d=G(\tau_{\text{ref}}^{+})-\hat r^{-}(e)$,
where $G\in\{0,1\}$ is the grounding reward of \cref{eq:reinforce} and $\hat r^{-}(e)$ is the vanilla success rate on that episode.
With $\bar d$ the mean gain over the collected pairs,
\begin{equation*}
\Delta_{\sigma,s,a}=
\begin{cases}
\mathrm{clip}(\lambda\bar d,-\kappa,\kappa) & \text{if } \#\text{pairs}\ge n_{\min}
        \text{ and } |\bar d|/\mathrm{SE}(d)\ge z_{\min},\\
0 & \text{if } \#\text{pairs}\ge n_{\min} \text{ but not significant},\\
-\kappa & \text{if } \#\text{pairs}<n_{\min},
\end{cases}
\end{equation*}
with $\lambda=15$, $\kappa=3$, $n_{\min}=10$, $z_{\min}=1$.
Tools that are never used in training are excluded from the support of $\pi_\sigma$.

\newpara{Policy update.}
We refine the initial logits $\theta^{0}$ into $\theta$ with the single batched update of \cref{eq:reinforce}.
For each training episode $e$, the group $\mathcal{G}_e=\{\tau,\tau_{\text{ref}}^{+},\tau_{\text{ref}}^{-}\}$ consists of a new memory-guided rollout $\tau$, whose actions are sampled from the initial policy $\pi^{0}_{\sigma_e}$ induced by $\theta^{0}$, together with the two reference rollouts already collected for $\Delta$.
In practice, $G(\tau)$ combines a binary grounding term with small shaping terms:
\begin{equation*}
G(\tau)=\mathbb{1}[\text{correct}]
 +0.1\cdot\mathbb{1}[\text{valid declaration}]
 +w_{\text{eff}}\,\mathbb{1}[\text{correct}]\,
   \frac{\min_{\tau'\in\mathcal{G}_e}|\tau'|}{|\tau|}
 -0.04\cdot\#\text{perception calls},
\end{equation*}
with $w_{\text{eff}}=0.01$ and $|\tau|$ the number of steps.
The binary term dominates; the remaining terms favor valid, short, and perception-frugal trajectories.
The advantage $A(\tau)=R(\tau)-b_e$ uses the group mean $b_e=\frac{1}{|\mathcal{G}_e|}\sum_{\tau'\in\mathcal{G}_e}R(\tau')$ as the baseline, as in \cref{eq:reinforce}.

In practice, \cref{eq:reinforce} is instantiated with three additions that stabilize a single update from one batch.
(i) \emph{Per-entry normalization}: the gradient of each logit is normalized by its visit count, so that $\theta_{\sigma,s,a}$ is shifted by the mean advantage $\bar A_{\sigma,s,a}$ of the rollouts that executed $a$ at $(\sigma,s)$; for a tabular softmax policy this corresponds to the natural-gradient form of \cref{eq:reinforce}.
(ii) \emph{Clipping}: the shift is clipped to $[-c,c]$ to bound the deviation from $\pi^{0}_\sigma$.
(iii) \emph{Significance gating}: as for $\Delta$, the shift is applied only to entries executed by at least $n'_{\min}=8$ rollouts and whose mean advantage is statistically distinguishable from $0$.

\newpara{Inference.}
The resulting $\mathcal{M}$ is frozen for evaluation. After each ReAct step, the agent state is updated to $s_t$ and the next action is chosen from $\pi_{\sigma^\ast}[s_t]$. During training rollouts we sample $a_t\sim\pi^{0}_{\sigma_e}[s_t]$ to explore alternative evidence-gathering orders; at evaluation we decode greedily, $a_t=\arg\max_{a}\pi_{\sigma^\ast}[s_t](a)$.

\subsection{Real-world Transfer}
\begin{figure}[t]
\centering
\includegraphics[width=0.4\linewidth]{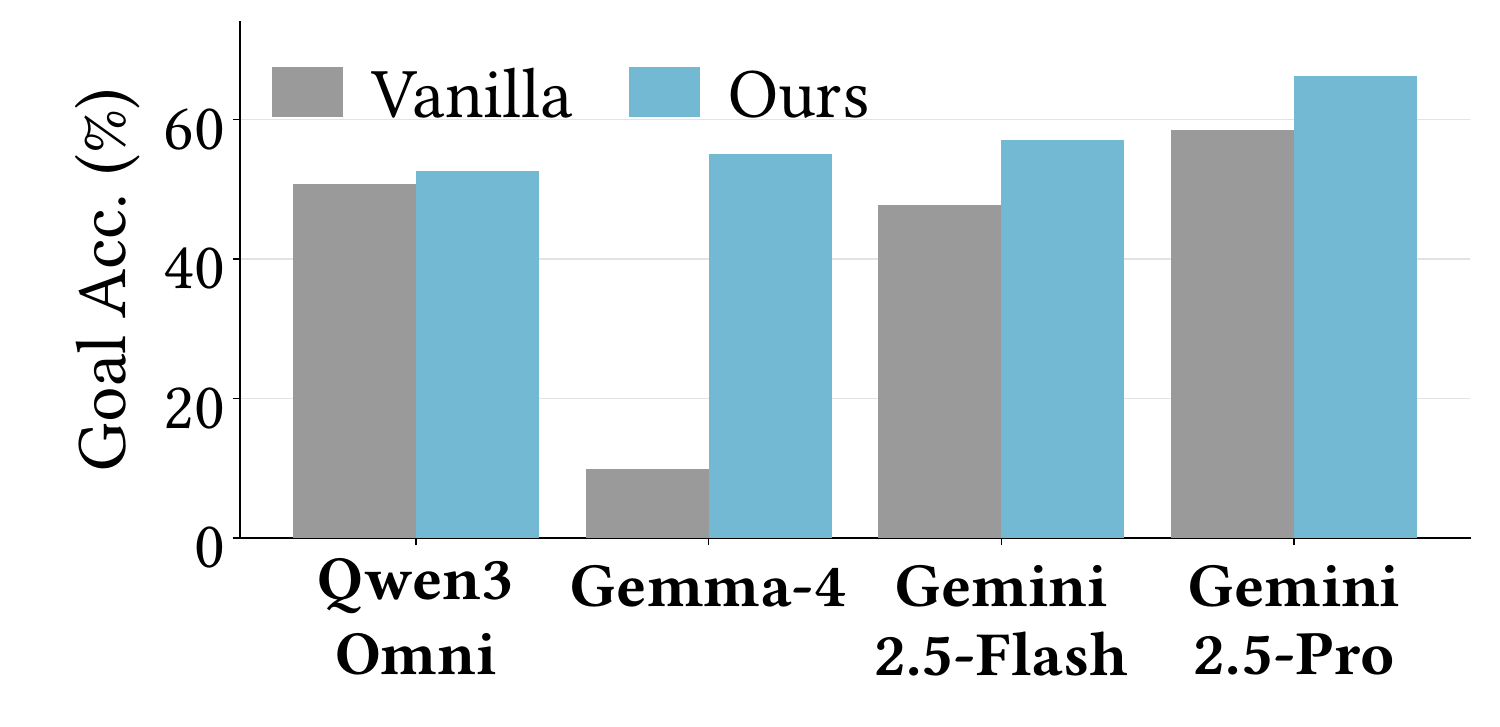}
\vspace{-3mm}
\caption{\textbf{Real-world transfer performance.} In addition to grounding accuracy, we report goal accuracy for \ours{} in the real-world setting. Despite using memory constructed solely from synthetic training data, \ours{} improves both grounding and goal accuracy in real-world environments.}
\label{fig:realworld_goal}
\end{figure}

In \cref{sec:ablation}, we reported only grounding accuracy for the real-world transfer experiment of \ours{}. We additionally report goal accuracy in \cref{fig:realworld_goal}.
Consistent with the grounding results, the memory constructed solely from synthetic training data also improves goal accuracy in real-world environments, indicating that the acquired knowledge transfers effectively beyond the synthetic domain.

\subsection{Oracle Tool Calling}
\label{sec:oracle_tool}

\begin{table}
\centering
\footnotesize
\setlength{\tabcolsep}{6pt}
\renewcommand{\arraystretch}{1.05}
\caption{\textbf{Effect of perception quality} on Qwen3-Omni-Think
(synthetic split). \emph{Predicted}: off-the-shelf perception tools;
\emph{Oracle}: tools return ground-truth attributes from the simulator.
Best in \textbf{bold}, second best \underline{underlined}.}
\label{tab:tool_calling}
\vspace{-2mm}
\begin{tabular}{@{}lcc@{}}
\toprule
\textbf{Method} & Gr. & Goal \\
\midrule
Qwen3-Omni-Think                & 66.4 & 47.9 \\
\quad +Predicted tools          & 65.8 & 47.1 \\
\quad +Oracle tools             & \underline{69.5} & \underline{51.1} \\
\quad +\textbf{\ours}           & \textbf{70.9} & \textbf{54.1} \\
\bottomrule
\end{tabular}
\vspace{-4mm}
\end{table}
Tool calling in the main paper relies on off-the-shelf visual and audio
perception models, so part of the remaining gap could be attributed to
perception errors rather than to the agent itself. To isolate this factor,
we define \emph{oracle} perception tools that return ground-truth
attributes directly from the simulator---person locations, head pose, and
sound source positions---which is possible only because the benchmark is
synthetically generated. 

This experiment is therefore restricted to the
synthetic split; the oracle tools cannot be instantiated on real-world data. \cref{tab:tool_calling} reports the results on Qwen3-Omni-Think. Replacing
predicted perception with oracle perception improves grounding from 65.8 to
69.5 and goal accuracy from 47.1 to 51.1, confirming that perception noise
accounts for part of the gap. However, even with perfect perception, oracle
tool calling remains below \ours{} (70.9 / 54.1), which uses only the
predicted tools. The gain of \ours{} thus does not come from better
perception alone but from how perceptual evidence is organized and used;
the goal-accuracy gap (+3.0) in particular indicates that oracle
observations by themselves are insufficient for correct action selection.

\subsection{Comparison with other methods}
\begin{table*}[t]
\centering
\caption{
\textbf{Comparison with agent baselines on OmniSmartHome.}
We report grounding accuracy (Gr.) and goal accuracy (Goal)
for each grounding type.
}
\label{tab:agent_baselines}
\vspace{-2mm}
\renewcommand{\arraystretch}{1}\small
\resizebox{\textwidth}{!}{%
\footnotesize
\begin{tabular}{
l
cc
cc
cc
cc
cc
cc
}
\toprule

&
\multicolumn{2}{c}{\textbf{All}}
&
\multicolumn{2}{c}{\textbf{G0} Speech}
&
\multicolumn{2}{c}{\textbf{G1} Gesture}
&
\multicolumn{2}{c}{\textbf{G2} Sound}
&
\multicolumn{2}{c}{\textbf{G3} Space}
&
\multicolumn{2}{c}{\textbf{G4} Identity}
\\

\cmidrule(lr){2-3}
\cmidrule(lr){4-5}
\cmidrule(lr){6-7}
\cmidrule(lr){8-9}
\cmidrule(lr){10-11}
\cmidrule(lr){12-13}

\textbf{Method}
&
Gr. & Goal
&
Gr. & Goal
&
Gr. & Goal
&
Gr. & Goal
&
Gr. & Goal
&
Gr. & Goal
\\

\midrule
\multicolumn{13}{c}{\textit{Qwen2.5-Omni}} \\
\midrule

Vanilla
& 33.0 & 17.6
& 55.5 & 25.5
& 30.6 & 19.4
& 28.1 & \textbf{14.2}
& 27.1 & 17.4
& 20.3 & 10.9
\\
\midrule

AWM
& 36.0 & 16.9
& \textbf{64.8} & 31.5
& 38.9 & 16.3
& 26.4 & 12.2
& 25.3 & 11.5
& 20.3 & 10.2
\\

MemP
& 35.9 & 14.3
& 63.8 & 20.8
& 28.8 & 15.6
& 30.9 & 12.2
& 27.4 & 14.6
& \textbf{23.4} & 8.1
\\

% \rowcolor{modelhighlight}
\textbf{\ours}
& \textbf{38.2} & \textbf{22.4}
& 61.5 & \textbf{34.6}
& \textbf{41.0} & \textbf{28.1}
& \textbf{35.1} & \textbf{14.2}
& \textbf{28.1} & \textbf{22.9}
& 22.9 & \textbf{11.7}
\\

\midrule
\multicolumn{13}{c}{\textit{Qwen3-Omni-Think}} \\
\midrule

Vanilla
& 65.9 & 48.4
& 87.8 & 67.4
& 66.0 & 56.6
& 79.9 & 56.9
& 72.2 & 55.2
& 28.6 & 11.7
\\
\midrule

AWM
& 64.1 & 46.8
& \textbf{89.3} & 66.9
& 64.6 & 53.1
& 70.1 & 50.3
& 72.6 & 56.2
& 27.6 & 12.0
\\

MemP
& 64.5 & 47.5
& 86.5 & 66.7
& \textbf{70.5} & 57.6
& 72.9 & 51.4
& 70.8 & 54.9
& 27.1 & 12.5
\\

% \rowcolor{modelhighlight}
\textbf{\ours}
& \textbf{70.2} & \textbf{53.9}
& 87.5 & \textbf{69.8}
& 70.1 & \textbf{58.3}
& \textbf{86.5} & \textbf{68.8}
& \textbf{77.1} & \textbf{62.2}
& \textbf{35.4} & \textbf{17.2}
\\

\bottomrule
\end{tabular}
}

\vspace{-3mm}
\end{table*}
\cref{tab:agent_baselines} compares our approach with two representative agent memory methods, Agent Workflow Memory (AWM)~\citep{wang2025awm} and MemP~\citep{fang2026memp}. AWM induces reusable workflows from past agent trajectories and selectively provides them to guide subsequent actions, while MemP further develops procedural memory through explicit memory construction, retrieval, and update mechanisms. Across both Qwen2.5-Omni and Qwen3-Omni-Think, our method achieves the highest grounding and goal accuracy. In particular, our method shows clear improvements on G2 (Sound), G3 (Space), and G4 (Identity), where resolving the request requires grounding ambiguous references in multimodal evidence. These results indicate that our procedural memory is better suited to OmniSmartHome than directly applying agent memory methods that are not designed for smart-home settings.

\subsection{Latency Analysis}
\begin{table}[t]
\centering
\caption{\textbf{Interaction overhead of PROME.} Average number of ReAct steps for vanilla and PROME across models.}
\begin{tabular}{l cc c}
\toprule
Model & Vanilla & \ours{} & $\Delta$ \\
\midrule
Gemma-4\psize{12B}               & 6.39 & 8.54 & +2.15 \\
Nemotron3-Omni\psize{30B-A3B}    & 4.90 & 5.51 & +0.62 \\
Qwen3-Omni-Think\psize{30B-A3B}  & 4.57 & 5.18 & +0.61 \\
Gemini-2.5-Flash                 & 5.22 & 7.01 & +1.78 \\
Gemini-2.5-Pro & 6.06 & 7.95 & +1.90\\
\midrule
Average                          &  5.43 & 6.84 & +1.41 \\
\bottomrule
\end{tabular}
\label{tab:latency}
\end{table}

We analyze the additional interaction overhead introduced by \ours{} by measuring the number of ReAct steps required relative to the vanilla baseline. As shown in the \cref{tab:latency}, \ours{} incurs only a modest increase in the number of steps, requiring on average 1.41 additional ReAct steps across models. This suggests that the performance gains of \ours{} come with limited additional inference overhead.

\section{LLM Prompt}
\label{app:prompt}
\subsection{Grounding Schema Registry Construction Prompt}
\label{app:ground_prompt}
\begin{promptbox}

\textbf{System Prompt}

\medskip
You maintain a compact registry of reusable smart-home grounding
situations.

\medskip
Inspect the current request, initial image, and initial audio, then
compare its grounding situation with the existing lessons as
\textbf{semantic categories}, not by exact wording. If it overlaps an
existing grounding situation, select that lesson. Create a new lesson
\textbf{only} when the current request requires a completely different
kind of grounding ambiguity.

\medskip
Judge sameness by the \textbf{kind of evidence relation} needed to
identify the target, rather than surface details or the downstream
request. Different values of the same relation generally belong
together; a genuinely different combination of evidence relations may
be a new situation. Use the task description, image, and audio together.

\medskip
A new lesson must describe \textbf{target identification only}, in one
clear general sentence, without concrete names, tool names, or a tool
sequence. Return only the requested JSON.

\medskip
\textbf{User Prompt}

\medskip
\begin{itemize}
    \item \textbf{Existing lessons:} \texttt{\{lesson\_id: lesson\}} for
    every lesson currently in the registry (empty at the start).
    \item \textbf{Request:} \texttt{\{query\}}
    \item \textbf{Initial observation:} the room image and the recording
    of the request, attached as image and audio inputs.
\end{itemize}

\textbf{Output (JSON only):} either \texttt{\{"lesson\_id": <existing id>\}}
or \texttt{\{"new\_lesson": "<one general sentence>"\}}.

\end{promptbox}

\subsection{Q1 Feasible Judge Prompt}

\begin{promptbox}

\textbf{System Prompt}

\medskip
You are a strict evaluator for smart-home LLM agents answering a
question about \textbf{ONE device} that the user identified only by
the sound it makes or by where it sits relative to them---never by
the device name, type, or room.

\medskip
Return \textbf{EXACTLY ONE character}: \texttt{A} (pass) or
\texttt{B} (fail). No other text.

\medskip
\textbf{WHAT TO JUDGE}

\begin{itemize}
    \item The user asked for the value of a single device attribute.
    \item Judge \textbf{ONLY} whether the agent's Final Answer reports
    the correct value for that attribute.
\end{itemize}

\textbf{CORRECTNESS}

\begin{itemize}
    \item The reported value must match the Expected Value provided below.
    Equivalent phrasings, formatting, and Matter unit conversions are
    acceptable (e.g., \texttt{2300 == 23.0}~$^\circ$\texttt{C},
    an enum name == its integer code,
    \texttt{True == "on"},
    \texttt{"LG Electronics" == "LG"}).

    \item The answer must be \textbf{grounded}: the agent must have
    actually retrieved the value with some tool, not guessed it.
\end{itemize}

\textbf{DO NOT PENALIZE}

\begin{itemize}
    \item Which tool the agent used to locate the device or read the value
    (\texttt{get\_room\_devices},
    \texttt{find\_devices\_by\_cluster},
    \texttt{get\_device\_structure},
    \texttt{get\_attribute}---all acceptable).

    \item Whether the agent mentions the room name or the device name
    (the user never gave a room or a name).

    \item Extra explanation or situational context in the answer,
    as long as the correct value is clearly stated.
\end{itemize}

Output \texttt{A} only if the value is correct \textbf{AND} grounded
in an observation; otherwise output \texttt{B}.

\medskip
\textbf{User Prompt}

\medskip
\begin{itemize}
    \item \textbf{User Query:} \texttt{\{query\}}
    \item \textbf{Target attribute(s) and the expected correct value(s):}
    \texttt{\{expected\_values\}}
    \item \textbf{Agent's ReAct Steps (tool calls and observations):}
    \texttt{\{react\_steps\}}
    \item \textbf{Agent's Final Answer:}
    \texttt{\{final\_answer\}}
\end{itemize}

\textbf{EVALUATION STEPS}

\begin{enumerate}
    \item Read the expected value(s) for the asked attribute(s).
    \item Check the agent's tool observations actually contain that value
    for the target device (grounding).
    \item Check the Final Answer states that value correctly
    (allow equivalent phrasings and unit conversions).
    \item Output \texttt{B} only if the value is wrong, missing, or fabricated.
    Ignore tool choice, room-name mention, and device-name mention.
\end{enumerate}

\textbf{Your Decision:} \texttt{A or B}

\end{promptbox}
\subsection{Infeasible Judge Prompt}

\begin{promptbox}

\textbf{System Prompt}

\medskip
You are a strict evaluator for smart-home LLM agents. The person's
request in this episode was \textbf{INFEASIBLE} for one specific,
labelled reason, and the agent was expected to recognise this, say
that it cannot (fully) carry out the request, and give that reason.

\medskip
Return \textbf{EXACTLY ONE character}: \texttt{A} (pass) or
\texttt{B} (fail). No other text.

\medskip
\textbf{Pass (\texttt{A}) ONLY IF the agent's Final Answer meets ALL conditions:}

\begin{enumerate}
    \item \textbf{Refusal:}
    it clearly states that the request cannot be carried out / was not
    carried out (a partial or hedged ``I did it but \ldots'' is not a refusal).

    \item \textbf{Right Reason:}
    the reason it gives matches the labelled reason type. Naming the
    type verbatim counts, and so does a general explanation that amounts
    to the same thing (see the description and examples given for the
    labelled type). A different reason, or a refusal with no reason, fails.

    \item \textbf{No False Claim:}
    it does \textbf{NOT} claim to have performed, scheduled, or changed
    anything toward the request. Read-only exploration in the trace is fine.

    \item \textbf{Consistency:}
    any factual claim it makes agrees with the tool observations in the trace.
\end{enumerate}

Otherwise, output \texttt{B}.

\medskip
\textbf{DO NOT PENALIZE}

\begin{itemize}
    \item Which tools were used or how many steps were taken.

    \item How the answer is worded or how long it is.

    \item Whether the room or device names are mentioned.

    \item Extra suggestions or clarifying questions---as long as
    conditions 1--4 hold.
\end{itemize}

\medskip
\textbf{User Prompt}

\medskip
\begin{itemize}
    \item \textbf{Person's request:}
    \texttt{\{query\}}

    \item \textbf{Labelled reason type:}
    \texttt{\{reason\_type\}}
    \begin{itemize}
        \item \textbf{Meaning:} \texttt{\{reason\_meaning\}}
        \item \textbf{Acceptable explanations sound like:}
        \texttt{\{reason\_examples\}}
    \end{itemize}

    \item \textbf{Ground-truth details (from the episode's evaluation spec):}
    \texttt{\{goals\}}

    \item \textbf{Agent's ReAct Steps (tool calls and observations):}
    \texttt{\{react\_steps\}}

    \item \textbf{Agent's Final Answer:}
    \texttt{\{final\_answer\}}
\end{itemize}

\textbf{EVALUATION STEPS}

\begin{enumerate}
    \item Does the Final Answer state that the request cannot be /
    was not carried out?

    \item Does the reason it gives match
    ``\texttt{\{reason\_type\}}'' (verbatim, or a general explanation
    amounting to the meaning above)?

    \item Does it avoid claiming to have performed or scheduled the
    requested change?

    \item Are its factual claims consistent with the tool observations?
\end{enumerate}

Pass only if all four hold.

\medskip
\textbf{Your Decision:} \texttt{A or B}

\paragraph{Reason types.}
The judge is parameterised by the episode's labelled infeasibility type;
\texttt{\{reason\_meaning\}} and \texttt{\{reason\_examples\}} are filled from
the table below.

\begin{itemize}
    \item \texttt{non\_existent}: no device matching what the person
    referred to exists in the room: the named / described device is not
    there, no device is making the described sound, or there is no device
    in the described direction.
    \emph{Examples:} ``there is no such device in the room'',
    ``nothing here is making that sound'',
    ``I could not find any device in that direction''.

    \item \texttt{non\_actuator}: the person wants a \textbf{room} state
    changed (brighter / darker / warmer / cooler / more or less humid /
    cleaner air) but no device in that room can move that state.
    \emph{Examples:} ``there is no light in that room to make it brighter'',
    ``nothing in the room can cool it down''.

    \item \texttt{physical\_limit}: the target device exists and was
    correctly identified, but it is already at the limit of what was asked
    (e.g.\ fan speed already 100 percent, brightness already at maximum),
    so the requested change cannot go any further.
    \emph{Examples:} ``it is already at maximum'',
    ``the fan is already at 100 percent so it cannot go faster''.

    \item \texttt{temporal\_conflict}: the timing of the request is
    contradictory or impossible: the person stated two times that do not
    agree (e.g.\ ``in 25 minutes, at 11:43'' when those are different
    moments), or asked for an ordering of events that cannot hold.
    \emph{Examples:} ``the two times you gave do not match'',
    ``the schedule contradicts itself''.
\end{itemize}
\end{promptbox}

\section{Human Annotation}
\subsection{Human Dataset Verifier}
\label{sec:human_valid}
\begin{figure*}[t]
    \centering
    \includegraphics[width=\textwidth]{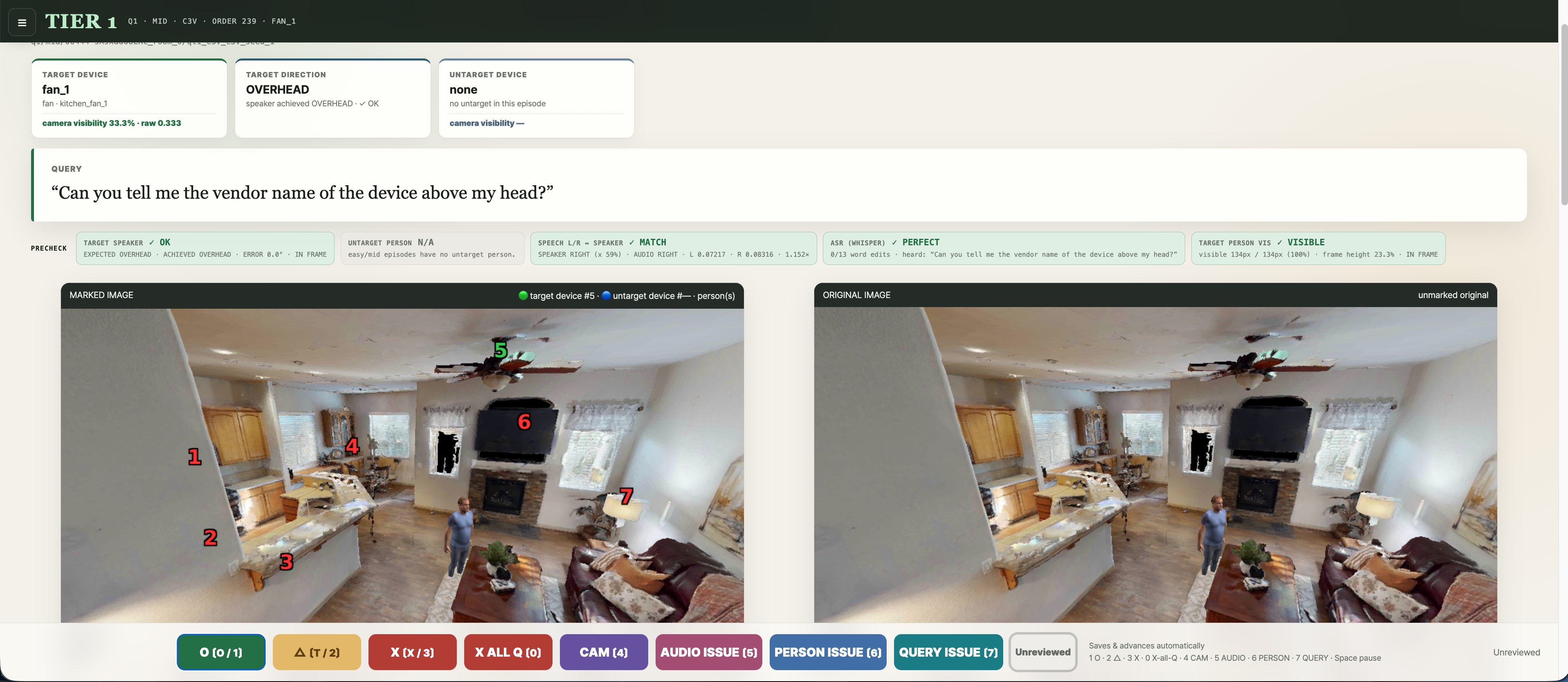}
    \caption{Human verification interface used to manually inspect and validate dataset examples.}
    \label{fig:annotator}
\end{figure*}

We manually verified the dataset to ensure the quality and correctness of the collected examples. Human annotators inspected each example using the annotation interface shown in \cref{fig:annotator}, and checked whether the provided inputs and annotations were consistent with the intended task.

\subsection{Human Performance}
\begin{figure*}[t]
    \centering
    \includegraphics[width=\textwidth]{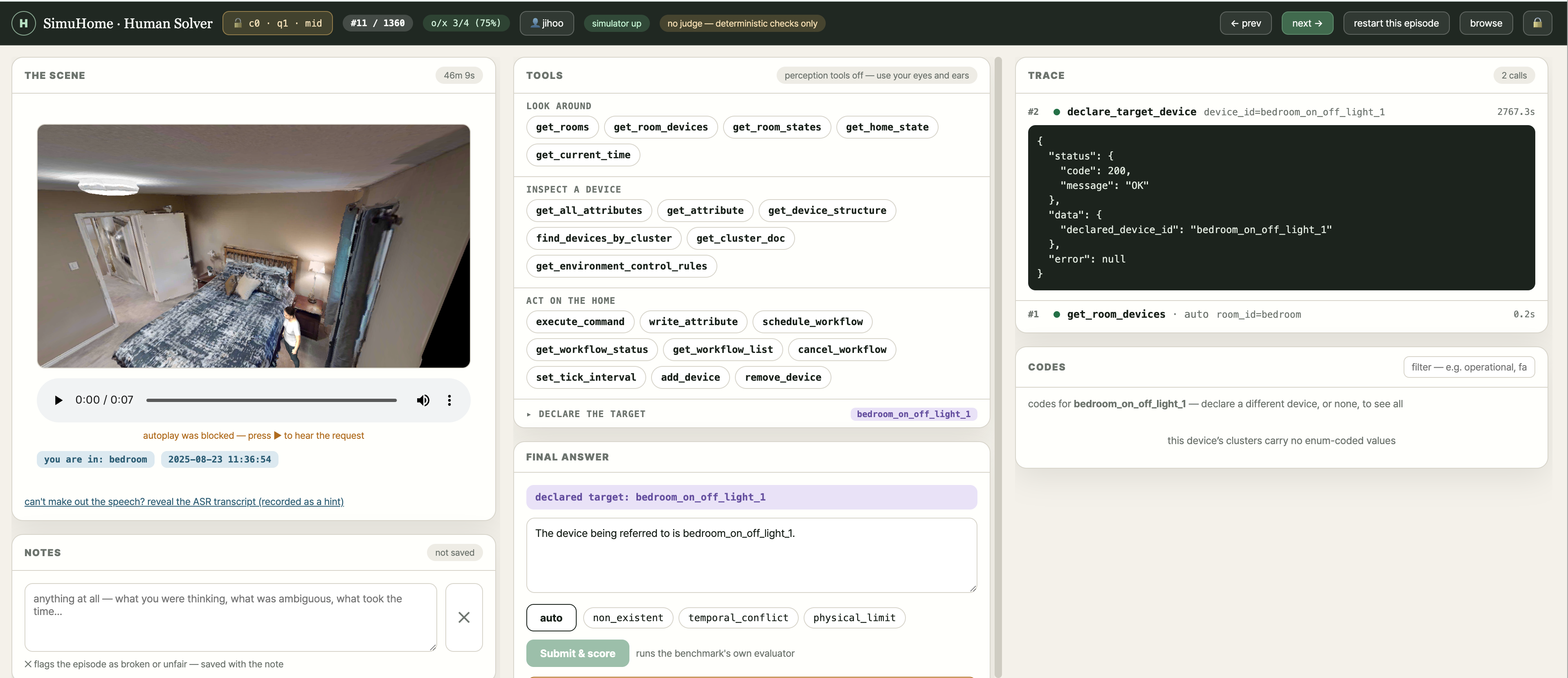}
    
    \caption{Human solver interface used to evaluate human performance on our benchmark. Participants inspect the task observations and interact with the environment to complete the given user request.}
    \label{fig:human_solver}
\end{figure*}

Evaluating human performance on agentic tasks is time-consuming, as solving each task requires multiple steps of environment exploration, reasoning, and interaction. Following prior agent benchmarks that estimate human performance on a representative subset of tasks~\cite{gou2026mindweb,zhou2024webarena,lu2024mathvista,chen2026mist}, we evaluated human performance on a balanced sample of approximately 25\% of the benchmark. We sampled 400 episodes in total and assigned 200 episodes to each of two human participants. Participants interacted with the environment through the human solver interface shown in \cref{fig:human_solver} and attempted to complete each task following the same task specification used for agent evaluation.

\section{Limitation}

Our benchmark assumes a single-turn interaction setting and thus does not capture multi-turn interactions in which users may provide additional context, clarification, or corrections. In addition, in the synthetic portion of the benchmark, visual observations are provided as static images rather than videos, which limits the evaluation of temporally evolving visual information. On the method side, \ours{} is primarily designed to improve grounding and does not explicitly address other challenging aspects of embodied or smart-home agents, such as temporal scheduling and implicit intent reasoning. Finally, although we analyze the number of steps taken by the agent, we have not yet evaluated the latency introduced by perception-tool inference or the feasibility of real-time deployment. Addressing these limitations is an important direction for future work.
 
\end{document}